%% file: egpaper_final.tex
\documentclass[10pt,twocolumn,letterpaper]{article}

\usepackage{iccv}
\usepackage{times}
\usepackage{epsfig}
\usepackage{graphicx}
\usepackage{amsmath}
\usepackage{amssymb}

\usepackage[accsupp]{axessibility}
\usepackage{dirtytalk}
\usepackage{gensymb}
\usepackage{caption}
\usepackage{subcaption}
\usepackage{enumitem}
\usepackage{multirow}
\usepackage{cuted}
\usepackage{cite}

\newcommand{\imgWidth}{0.124}

\newlength\savewidth\newcommand\shline{\noalign{\global\savewidth\arrayrulewidth
  \global\arrayrulewidth 1pt}\hline\noalign{\global\arrayrulewidth\savewidth}}
\newcommand{\tablestyle}[2]{\setlength{\tabcolsep}{#1}\renewcommand{\arraystretch}{#2}\centering\footnotesize} 

\makeatletter
\renewcommand\paragraph{\@startsection{paragraph}{4}{\z@}%
  {.5em \@plus1ex \@minus.2ex}%
  {-.5em}%
  {\normalfont\normalsize\bfseries}}
\makeatother

\usepackage[breaklinks=true,bookmarks=false]{hyperref}

\iccvfinalcopy 

\def\iccvPaperID{1808} 

\ificcvfinal\fi

\begin{document}

\title{Beyond Skin Tone: A Multidimensional Measure of Apparent Skin Color}

\author{
William Thong\thanks{Corresponding author: william.thong@sony.com}\\
Sony AI
\and
Przemyslaw Joniak\thanks{Work done while an intern at Sony AI}\\
The University of Tokyo
\and
Alice Xiang\\
Sony AI
}

\maketitle
\ificcvfinal\thispagestyle{empty}\fi

\begin{abstract}
This paper strives to measure apparent skin color in computer vision, beyond a unidimensional scale on skin tone.
In their seminal paper Gender Shades, Buolamwini and Gebru have shown how gender classification systems can be biased against women with darker skin tones.
Subsequently, fairness researchers and practitioners have adopted the Fitzpatrick skin type classification as a common measure to assess skin color bias in computer vision systems.
While effective, the Fitzpatrick scale only focuses on the skin tone ranging from light to dark.
Towards a more comprehensive measure of skin color, we introduce the hue angle ranging from red to yellow.
When applied to images, the hue dimension reveals additional biases related to skin color in both computer vision datasets and models.
We then recommend multidimensional skin color scales, relying on both skin tone and hue, for fairness assessments.
\end{abstract}

\section{Introduction}

This paper focuses on measuring apparent skin color in images for fairness benchmarking, and moves toward a multidimensional score beyond the skin tone depth.
Adverse decisions can arise in common computer vision models, as strikingly identified by Buolamwini and Gebru~\cite{buolamwini2018gender}.
This has an impact on real-life applications, as models can produce wrong skin lesion diagnostics~\cite{daneshjou2022disparities} or incorrect heart rate measurements~\cite{shcherbina2017accuracy,bent2020investigating,gottlieb2022assessment} for individuals with darker skin tones.
It is therefore critical to identify to what extent visual datasets and models are affected by changes in skin color.
To achieve this, it becomes necessary to develop comprehensive skin color scores to characterize images of individuals, otherwise a fairness evaluation based on skin color would not be feasible.

Describing the apparent skin color remains an open challenge, as the final visual color perception results from a complex physical and biological phenomenon~\cite{jablonski2004evolution}.
The skin is a multilayered structure, which varies among individuals: not every individual will have the same amount and distribution of carotene, hemoglobin, or melanin throughout the different layers~\cite{Chardon1991SkinCT,DelBino2013VariationsIS}.
Hence, modeling the intrinsic skin color through the amount of reflected, scattered, or absorbed light for every individual is a difficult and complex task in images~\cite{tsumura2003image}.
Additionally, color perception from the human visual system adds more complexity to the problem, as colors can be differently perceived depending on their context or people's cultures~\cite{gevers2012color}.
Instead, it is simpler and more appropriate to develop representative color scales to capture the variations in apparent skin color.

\begin{figure}[t]
\centering
  \hfill
  \begin{subfigure}{.0225\linewidth}
  \centering
  \includegraphics[width=0.95\linewidth]{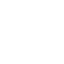}\\
  \includegraphics[width=0.95\linewidth]{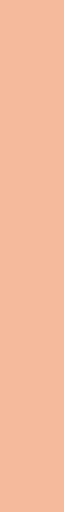}\\
  \includegraphics[width=0.95\linewidth]{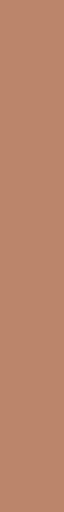}\\
  \includegraphics[width=0.95\linewidth]{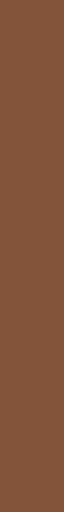}\\
  \includegraphics[width=0.95\linewidth]{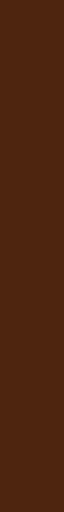}
  \end{subfigure}
  \begin{subfigure}{.18\linewidth}
  \centering
  \includegraphics[width=0.95\linewidth]{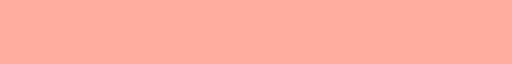}
  \includegraphics[width=0.95\linewidth]{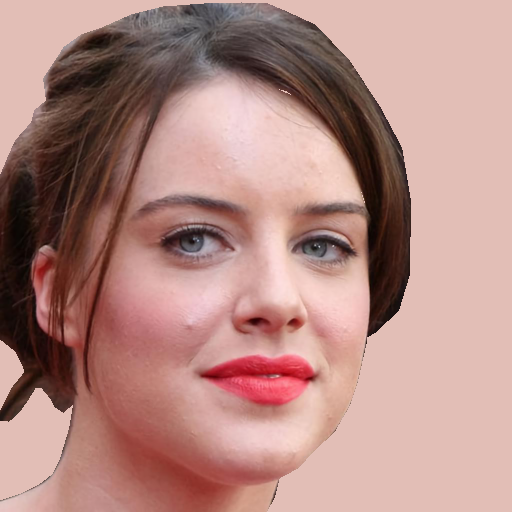}
  \includegraphics[width=0.95\linewidth]{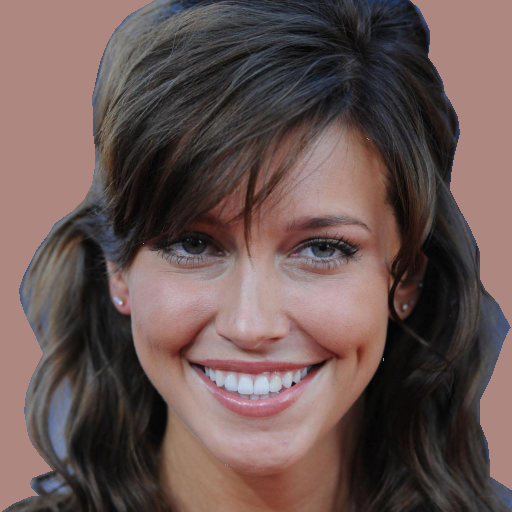}
  \includegraphics[width=0.95\linewidth]{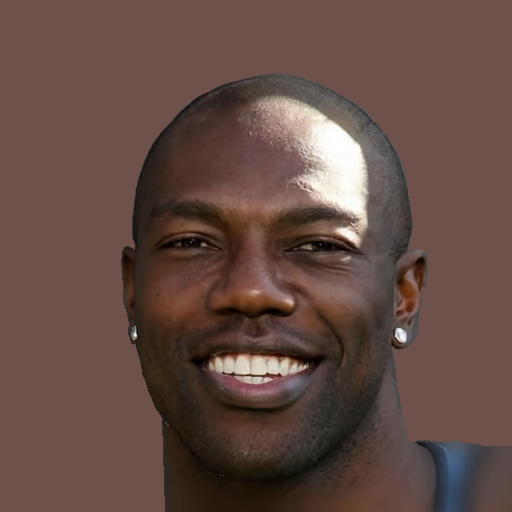}
  \includegraphics[width=0.95\linewidth]{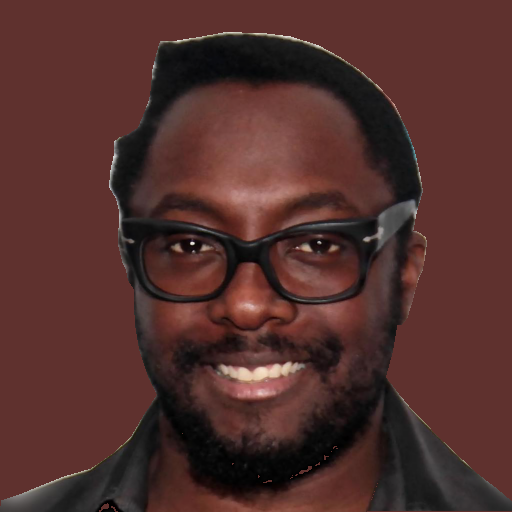}
  \end{subfigure}\hfill
  \begin{subfigure}{.18\linewidth}
  \centering
  \includegraphics[width=0.95\linewidth]{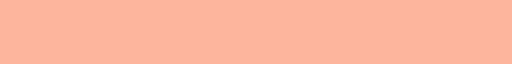}
  \includegraphics[width=0.95\linewidth]{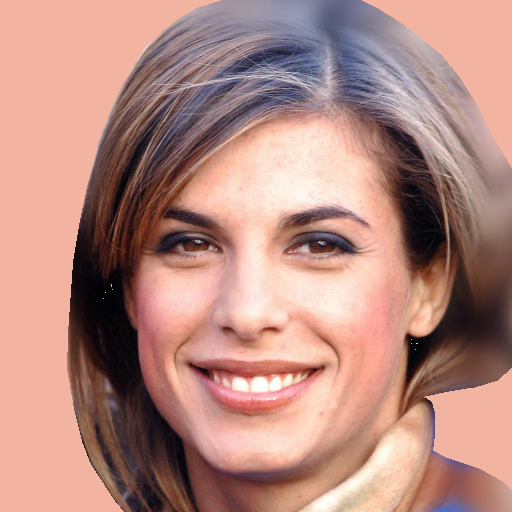}
  \includegraphics[width=0.95\linewidth]{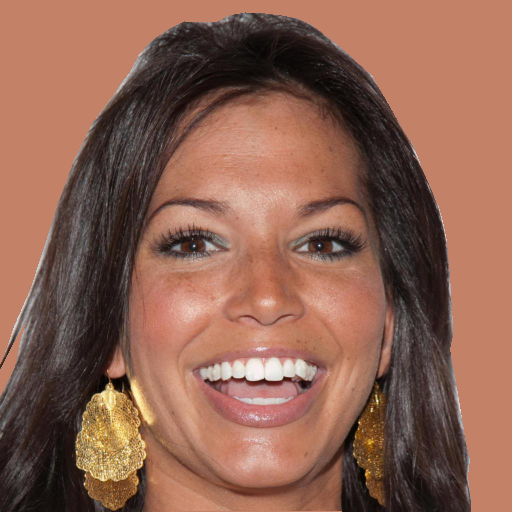}
  \includegraphics[width=0.95\linewidth]{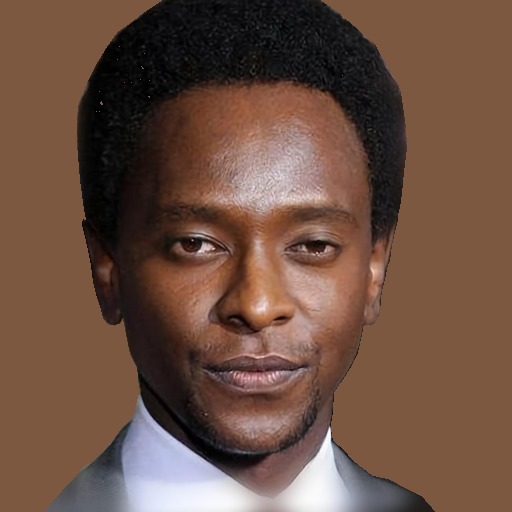}
  \includegraphics[width=0.95\linewidth]{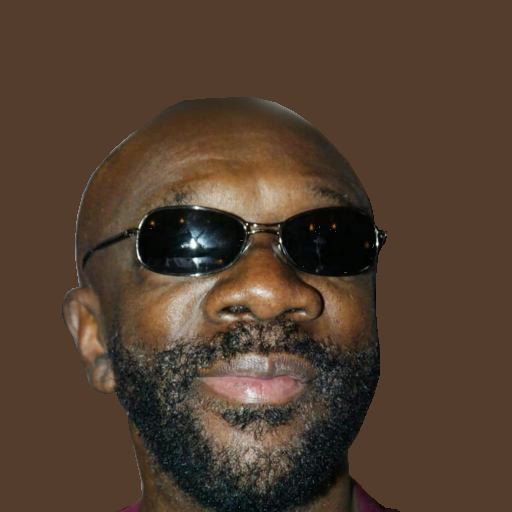}
  \end{subfigure}\hfill
  \begin{subfigure}{.18\linewidth}
  \centering
  \includegraphics[width=0.95\linewidth]{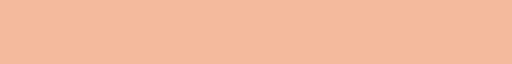}
  \includegraphics[width=0.95\linewidth]{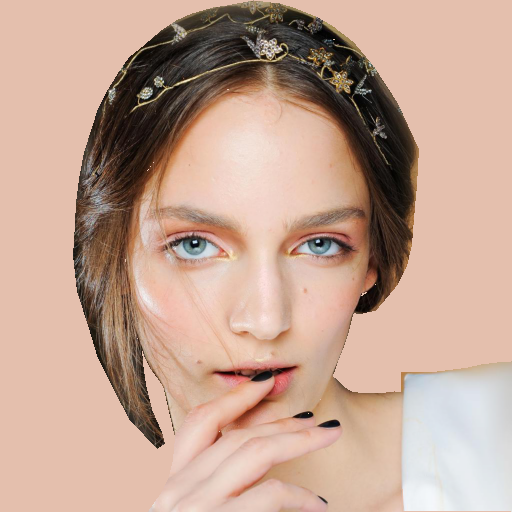}
  \includegraphics[width=0.95\linewidth]{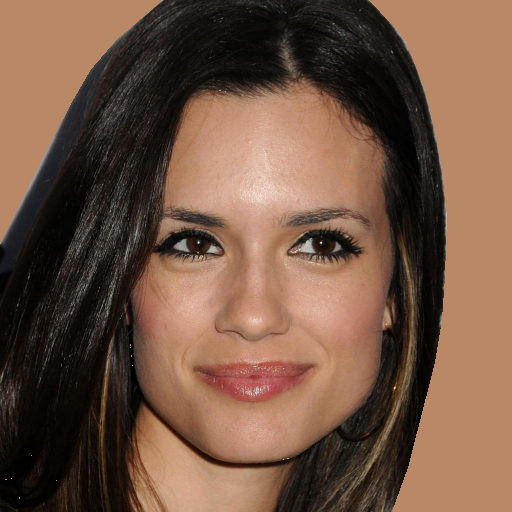}
  \includegraphics[width=0.95\linewidth]{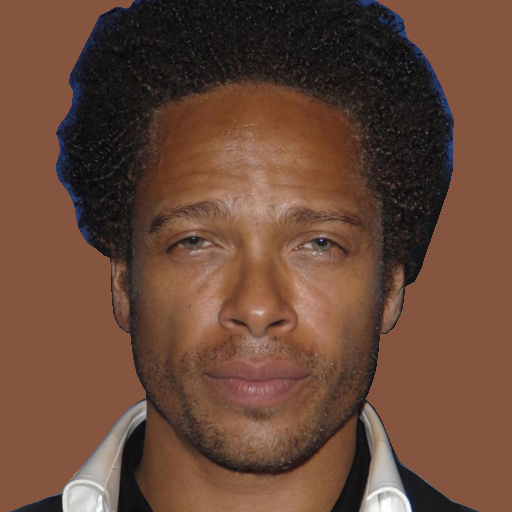}
  \includegraphics[width=0.95\linewidth]{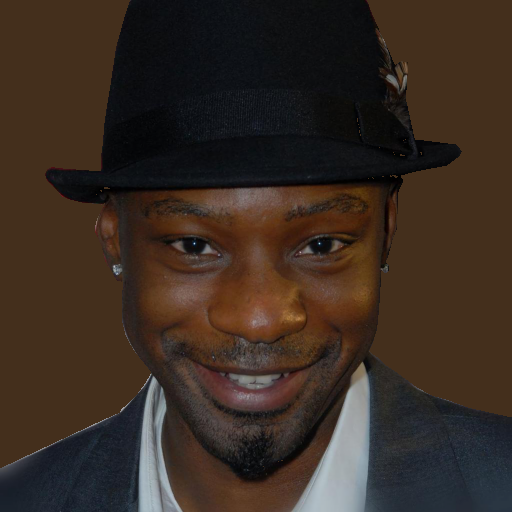}
  \end{subfigure}\hfill
  \begin{subfigure}{.18\linewidth}
  \centering
  \includegraphics[width=0.95\linewidth]{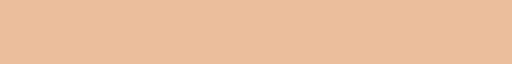}
  \includegraphics[width=0.95\linewidth]{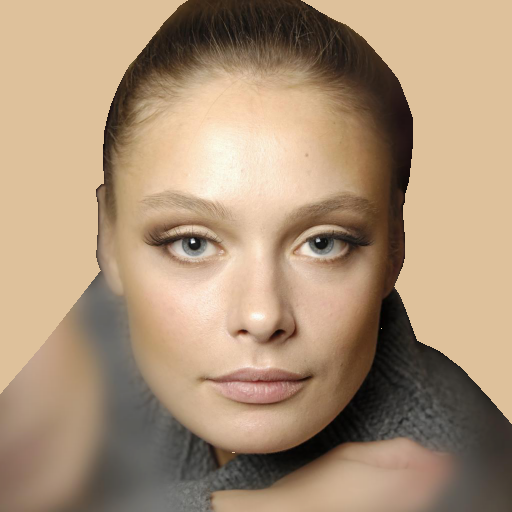}
  \includegraphics[width=0.95\linewidth]{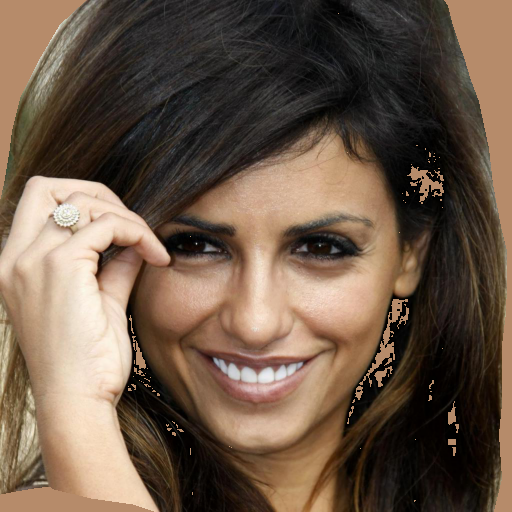}
  \includegraphics[width=0.95\linewidth]{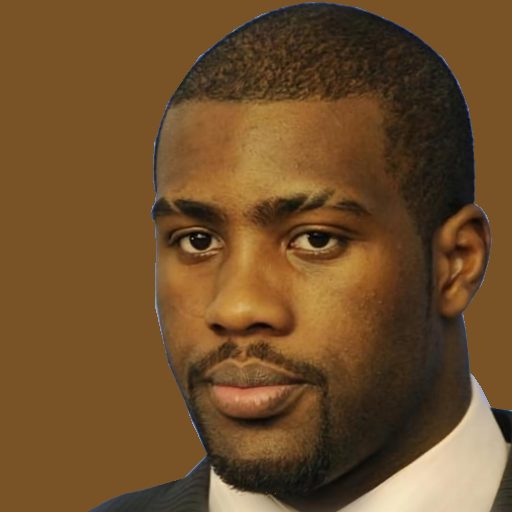}
  \includegraphics[width=0.95\linewidth]{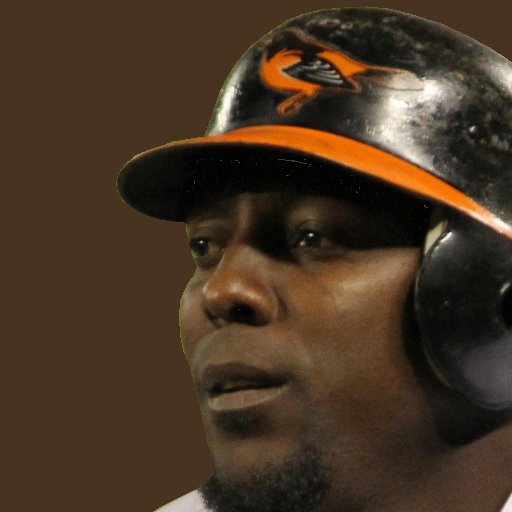}
  \end{subfigure}\hfill
  \begin{subfigure}{.18\linewidth}
  \centering
  \includegraphics[width=0.95\linewidth]{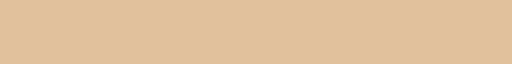}
  \includegraphics[width=0.95\linewidth]{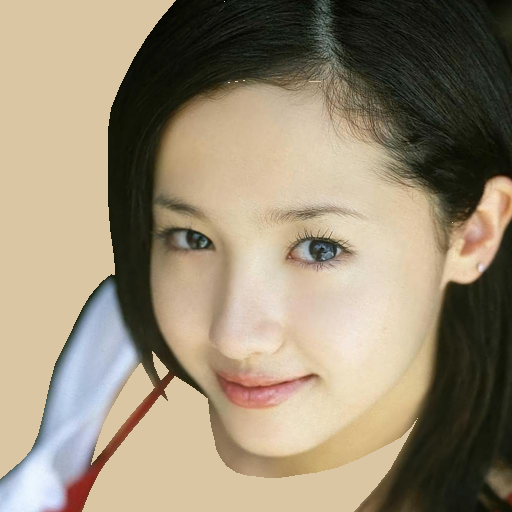}
  \includegraphics[width=0.95\linewidth]{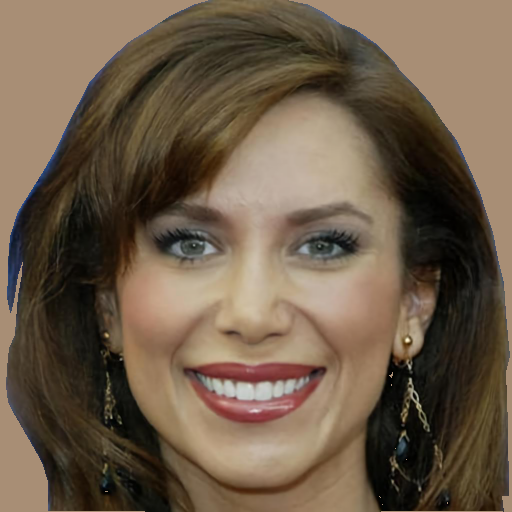}
  \includegraphics[width=0.95\linewidth]{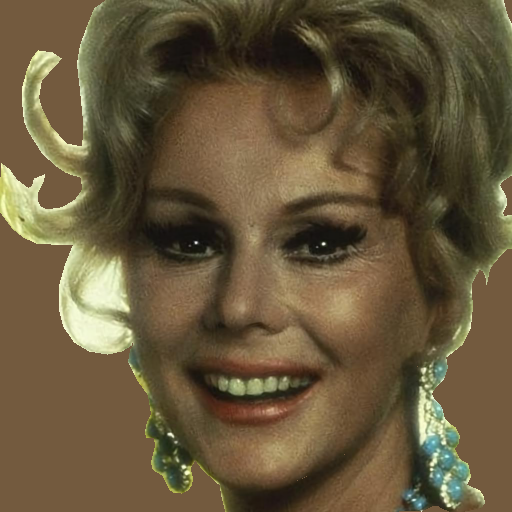}
  \includegraphics[width=0.95\linewidth]{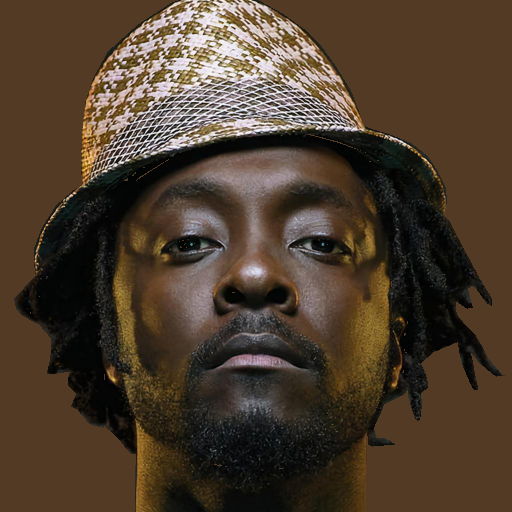}
  \end{subfigure}\hfill\null
  \caption{\textbf{Apparent skin color} in images results from a complex phenomenon, where it varies in tone from light to dark (vertical axis) and hue from red to yellow (horizontal axis).
  This figure depicts examples from CelebAMask-HQ.
  In this paper, we introduce a measure of the skin hue, besides the commonly used skin tone, to quantify apparent skin color variations.}
  \label{fig:teaser}
\end{figure}

The commonly accepted standard for skin color scale is the Fitzpatrick skin type classification~\cite{Fitzpatrick1988TheVA}, which categorizes skin color into six different types based on \textit{skin tone}, ranging from \textit{light} to \textit{dark}.
It has unsurprisingly become a useful tool for fairness analysis~\cite{buolamwini2018gender,raji2019actionable} because skin tone annotations may serve as a proxy for race or ethnicity annotations. Indeed, such sensitive attributes are subject to significant data privacy protections~\cite{gdpr}, and are often unavailable or inferred in visual datasets~\cite{karkkainen2021fairface,andrus2021we}, making the measure of skin tone an alternative fairness tool.
Yet, while practical and effective, reducing the skin color to its tone is limiting given the skin constitutive complexity.
As illustrated in Figure~\ref{fig:teaser}, apparent skin color also varies along other axis, such as the \textit{skin hue}.
For example, when aging, Asian skin becomes darker and \textit{more yellow} while Caucasian skin becomes darker and \textit{redder}~\cite{de2007development}. 
Focusing on skin tone would not capture such change in hue as it only assesses the skin lightness or darkness.
In this paper, we therefore promote a multidimensional scale to better represent apparent skin color variations among individuals in images.

Akin to previous works on fairness analysis in computer vision (\eg, \cite{buolamwini2018gender,raji2019actionable, zhao2021understanding,wilson2019predictive}), we are interested in characterizing \textit{apparent} skin color rather than \textit{true} skin color.
The \textit{apparent} skin color is the one depicted in images, and the one that a computer vision model would see, while the \textit{true} skin color characterizes the constitutive skin color without the influence of external factors such as illumination or color cast.
Assessing the \textit{true} skin color is more important for dermatology~\cite{kinyanjui2020fairness,sachdeva2009fitzpatrick} or cosmetics~\cite{DelBino2013VariationsIS,ly2020research} applications as the constitutive color leads to more specific diagnostics or treatments 
and requires an active involvement with practitioners to avoid any misusage or mistrust~\cite{groh2022towards}.
In this paper, we rather focus on the assessment of computer vision models, which are fed images in the wild, and thus only consider the \textit{apparent} skin color in images.

Our main contribution is to demonstrate the relevance and benefits of a multidimensional skin color scale for fairness assessments in computer vision.
\textit{First}, we introduce a step towards more comprehensive apparent skin color scores. Rather than classifying skin color in types, as done with the Fitzpatrick scale, we measure automatically and quantitatively skin color in a multidimensional manner in images. We propose to focus on the perceptual light $L^*$, as a measure of skin tone, and the hue angle $h^*$, as a measure of skin hue; which results in a multidimensional measure for every image.
\textit{Second}, we showcase the benefits of a multidimensional measure of skin color by
(i) quantifying to what extent common image datasets are skewed towards light-red skin color and under-represent dark-yellow skin color, and how generative models trained on these datasets reproduce a similar bias;
(ii) revealing multidimensional skin color biases in saliency-based image cropping and face verification models;
and (iii) measuring the causal effect of skin color in attribute prediction in multiple commercial and non-commercial models.
Overall, our contributions to assessing skin color in a multidimensional manner offer novel insights, previously invisible, to better understand biases in the fairness assessment of both datasets and models.

\section{Background on Skin Color Scales}\label{sec:skintone}

Scoring the skin color of individuals has wide implications in various fields.
We focus in this section on the Fitzpatrick skin type classification scale~\cite{Fitzpatrick1988TheVA}, which is widely used in practice but starts to raise ethical concerns due to its original design and its lack of representativeness~\cite{pichon2010measuring}.
We describe below its categorization methodology, as well as its current usage, limitations and impact for fairness assessment in computer vision.

\paragraph{Fitzpatrick skin type} classification scale categorizes the skin response to ultraviolet A light~\cite{Fitzpatrick1988TheVA}.
This was originally used to assess the suntanning pathways of skin,~\ie, how sensitive skin is to sun exposure, and how it evolves over time. While the classification was first proposed to categorize Caucasian skin into four different types, it has later been extended to include brown and darker skins, resulting in six different skin types~\cite{Fitzpatrick1988TheVA}.
The scale starts from type I lighter skin, which \say{always burns and never tans}, and continues to type VI darker skin, which \say{never burns}.

Since its original introduction, the Fitzpatrick skin type has been, and still is, a central role in the literature. It now goes beyond the characterization of suntanning pathways, and serves as a color classification scale for skin in fields such as cosmetics to measure the efficacy of aesthetic cosmetology (\eg,~\cite{DelBino2013VariationsIS,ly2020research}) or clinical dermatology (\eg,~\cite{kinyanjui2020fairness,sachdeva2009fitzpatrick}) for skin analysis and treatment.
In this paper, akin to the seminal work of Buolamwini and Gebru~\cite{buolamwini2018gender}, we focus on leveraging apparent skin color scores to study potential discrimination coming from data and model biases.

However, such widespread use of the Fitzpatrick skin type raises ethical questions as it was originally developed for Caucasian skin tones, implying a lack of representativeness and utility for other groups.
As highlighted by Pichon~\etal~\cite{pichon2010measuring}, the Fitzpatrick skin type has limited utility in dermatology for several ethnic groups such as Asians~\cite{choe2006difference}, Arabs~\cite{venkataram2003correlating} or African-Americans~\cite{wilkes2015fitzpatrick}, because characterizing how skin \say{burns} or \say{tans} might be restrictive.
Adding race labels does not provide better predictors for skin characteristics either, as it is still not enough to represent its variation~\cite{he2014self}.
This representativeness limitation accentuates the need for more comprehensive classification types to characterize skin color.
Concurrent to our work, the Monk Skin Tone scale\cite{Monk_2019,schumann2023consensus} proposes ten different types for an alternative to better reflect a diversity of communities.
In this paper, we go beyond the assessment of skin tone and propose the skin hue to achieve a multidimensional score for apparent skin color.

Another limitation of the Fitzpatrick skin type comes from the annotation process. While the usage of a spectrophotometer to measure the skin reflectance remains the standard to provide the ground truth score of the true or constitutive skin color~\cite{ly2020research}, it might not be available in practice.
As such skin type annotation is usually self-reported by the subject~\cite{Fitzpatrick1988TheVA}, labeled by domain experts~\cite{ware2020racial}, or labeled by non-domain experts~\cite{buolamwini2018gender}.
Unfortunately, in all the above cases, it appears that misclassification can occur in the Fitzpatrick scale, as current skin type definitions are not objective and descriptive enough, and do not capture the skin color variability among different ethnicities~\cite{eilers2013accuracy}.
Another reason for such misclassification comes from the annotator bias, where the demographics and background of an annotator influence the skin color score assignment~\cite{Campbell2020IsAP}.
These annotation limitations indicate the need for quantitative and more objective measures for skin color. In this paper, we extract skin color scores in facial images automatically, without the need of external human annotators, to obtain a representative scalar scores.

\paragraph{Social impact of skin color scales.}
In his seminal paper, Fitzpatrick~\cite{Fitzpatrick1988TheVA} considers race and ethnicity as cultural and political constructs. 
Current scales have been developed to describe skin color rather than providing a race or ethnic label.
Still, the Fitzpatrick skin type has deviated from its original usage as up to fifty percent of dermatologists use the Fitzpatrick types to describe race and ethnicity~\cite{ware2020racial}, and doing so can be harmful as such simplification ignores their complexity~\cite{hanna2020towards}.
Furthermore, the lack of representativeness in the Fitzpatrick skin type can lead to mis-representing some ethnical groups~\cite{grother2019face,phillips2011other}, as skin color can have a wide spectrum within an ethnic group or between two groups. For example, two groups can have a similar skin tone but differ on other axes.
In light of this, this paper focuses on understanding performance discrepancy related to skin color, with more comprehensive scores to better represent human skin color variation.

\paragraph{Applications in fairness assessment.}
In recent years, there has been growing awareness on the limitations of computer vision models to be biased against under-represented groups~\cite{buolamwini2018gender}.
It is thus critical to develop fairness tools that can help assess potential biases and document them in datasheets~\cite{gebru2021datasheets} and model cards~\cite{mitchell2019model}.
In the context of this paper, we are interested in developing a fairness tool to better assess and quantify biases related to skin color.
Towards this goal, collecting skin tone annotations has enabled bias identification in facial recognition~\cite{buolamwini2018gender,raji2019actionable,jaiswal2022two}, image captioning~\cite{zhao2021understanding}, person detection~\cite{wilson2019predictive}, skin image analysis in dermatology~\cite{kinyanjui2020fairness,daneshjou2022disparities}, face reconstruction~\cite{feng2022towards} and detecting deepfakes~\cite{hazirbas2021towards} among other tasks.
In this paper, we build on this line of work and propose complementary and novel tools for measuring and extracting multidimensional skin color scores, and showcase their relevance and effectiveness for revealing dataset and model biases.

While the paper focuses on bias identification and measurement, our multidimensional skin color scale could also enable how to address bias mitigation. Such related works commonly rely on data augmentation~\cite{ramaswamy2021fair}, adversarial debiasing~\cite{wang2019balanced}, contrastive learning~\cite{park2022fair}, independent classifiers~\cite{wang2020towards}, or debiasing both feature and label embeddings~\cite{thong2021feature}. Our skin color scores could inform which samples models are struggling with, and provide a solution by augmenting images or mitigating model representations in a multidimensional manner.

\section{Multidimensional Skin Color Scores}\label{sec:comprehensive}

Given the limitations of the Fitzpatrick scale in its definition or annotation process, deriving quantitative metrics enables more reliable skin color scores.
Instead of asking a subject for a self-identification of the skin color type or collecting the skin color type from an annotator, it is preferable to compute a skin color score from a point measurement.
Indeed, this mitigates the subjectivity of the (self-)annotator as well as the inter-rater reliability~\cite{Chardon1991SkinCT,ly2020research,weatherall1992skin}.
In this section, we introduce the importance of the color space for representing images and define metrics for skin color scores.

\paragraph{Colorimetry} aims to represent faithfully the human perception of colors. Towards this goal, the Commission Internationale de l'Eclairage (CIE) establishes standards regarding illuminants, tristimulus values, or color spaces~\cite{international2004colorimetry}.
While images are usually represented in the standard RGB space, it might be more relevant to represent them in a color space that better reflects the human perception when assessing skin color variation.

In this paper, we are particularly interested in the CIE $L^*a^*b^*$ (CIELAB) color space, originally introduced in 1976~\cite{international2004colorimetry}, which
correlates with the response of the human eye by covering its entire range of color perception:
the $L^*$ component corresponds to the perceptual lightness and ranges from black at value 0 to white at value 100;
the $a^*$ component describes the green-red opponent colors, with negative values corresponding to green and positive values to red;
and the $b^*$ component refers to the blue-yellow opponent colors, with negative values corresponding to blue and positive values to yellow.

\paragraph{Individual typology angle} (ITA) provides a quantitative alternative to the Fitzpatrick scale~\cite{Chardon1991SkinCT}.
The individual typology angle is commonly used to describe the skin color on spectrophotometer measurements for aesthetic cosmetology~\cite{ly2020research} or clinical dermatology~\cite{wilkes2015fitzpatrick}. In fairness analyses, it has notably been applied to natural images of faces images~\cite{merler2019diversity,karkkainen2021fairface}, and skin analysis~\cite{kinyanjui2020fairness}.
Concretely, the individual typology angle is defined in the CIELAB color space as follows:
\begin{equation}
    \textrm{ITA} = \textrm{arctan}\left(\frac{L^* - 50}{b^*}\right) \times \frac{180\degree}{\pi}
\end{equation}
where a perceptual lightness at value 50 corresponds to a maximum chroma. Only the $L^*$ and $b^*$ components are selected as they are the ones that explain best the variation of the suntanning pathways of the skin.

Despite its advantages of providing a quantitative measure, the individual typology angle is not a comprehensive skin color score. It was originally developed for the suntanning pathways of Caucasian skin~\cite{Chardon1991SkinCT}, similar to the Fitzpatrick scale~\cite{Fitzpatrick1988TheVA}. In their original paper, Chardon~\etal~\cite{Chardon1991SkinCT} even raise this limitation as they acknowledge the absence of the $a^*$ component, which should be taken into account to better represent other types of population.
Furthermore, while there exists a mapping between angles and Fitzpatrick skin color types (\eg, values above 28\degree~correspond to light skin tones, type I to III; and to dark skin tones, type IV to VI, otherwise), the low correlation with expert annotations makes it less reliable~\cite{groh2022towards}.
Given these limitations, there is a need to provide a more comprehensive assessment of skin color.

\paragraph{Hue angle} provides a colorimetric measure to describe the perceived gradation of color~\cite{international2004colorimetry}. In the CIELAB color space, the hue angle is defined as follows:
\begin{equation}
    h^* = \textrm{arctan}\left(b^*/a^*\right)
\end{equation}
where $h^*$ goes from 0 to 360\degree~as $a^*$ and $b^*$ components are unbounded. That said, we are mainly interested in angles between 0 and 90\degree~(\ie, positive values of $a^*$ and $b^*$), as the skin color is expressed through red and yellow colors~\cite{Chardon1991SkinCT}.
Initially proposed by Weatherall and Coombs~\cite{weatherall1992skin} for skin color measurements, the hue angle has proven to be suitable as an additional dimension to design proprietary skin color scales for aesthetics cosmetology~\cite{de2007development}.

While these other fields have been exploring a multidimensional measure for skin color, fairness benchmarks in computer vision have mainly focused on a unidimensional measure of skin tone.
In this paper, we propose to consider both the skin tone and the skin hue as a multidimensional measure for skin color analysis in images. To the best of our knowledge, the hue angle has not been used for fairness analysis in computer vision. 
We take inspiration from de Rigal~\etal~\cite{de2007development}, and focus on the measurement of $L^*$ for skin tone and $h^*$ for skin hue. Measuring $L^*$ instead of ITA avoids having correlated measures as both ITA and $h^*$ contain the $b^*$ component.

To extract the apparent skin color scores from images, we build on the algorithm initially proposed by Merler~\etal~\cite{merler2019diversity} for the Diversity in Faces dataset (see Section 4.6 in their paper), and generalize their method to handle any scalar scoring value, any face pose and facial variation. We provide the details of the methodology and a robustness analysis to different illuminations in Figure~\ref{fig:robustness}.
We focus in the next section on understanding the effect of skin color in image datasets on computer vision models.

\section{Fairness Benchmarking with Multidimensional Skin Color Scores}\label{sec:benchmark}

First, we quantify the skin color bias in face datasets, and in generative models trained on such datasets. This reveals a skewness towards light-red skins and an under-representation of dark-yellow skins.
Second, we break down results by skin color of saliency-based image cropping and face verification algorithms. This reveals that model bias not only exists for skin tone, but also for skin hue.
Third, we investigate the causal effect of skin color in attribute prediction. This reveals performance differences when skin color changes, as classifiers tend to predict people with lighter skin tones as more feminine, and people with redder skin hue as more smiley.
Code can be found at \href{https://github.com/SonyResearch/apparent_skincolor}{https://github.com/SonyResearch/apparent\_skincolor}.

\subsection{Skin color bias in datasets}

Skin color scores enable the assessment of potential biases in a given image dataset. By extracting them for every sample in the dataset, it is possible to estimate the distribution of every subgroup and characterize how unbalanced the distribution might be.
We focus on the CelebAMask-HQ~\cite{lee2020CelebAMask}, composed of 30,000 images available with a non-commercial research agreement, and FFHQ-Ageing~\cite{or2020lifespan}, composed of 70,000 images available with a Creative Commons BY-NC-SA 4.0 license. Both datasets provide ground truth segmentation masks of the skin, which we use to extract skin color scores. Note that both datasets are derivative datasets from CelebA~\cite{liu2015faceattributes} and FFHQ~\cite{karras2019style}, respectively, with images crawled from social media platforms.

\begin{figure}
\centering
  \begin{subfigure}{.5\linewidth}
  \centering
  \includegraphics[width=\linewidth]{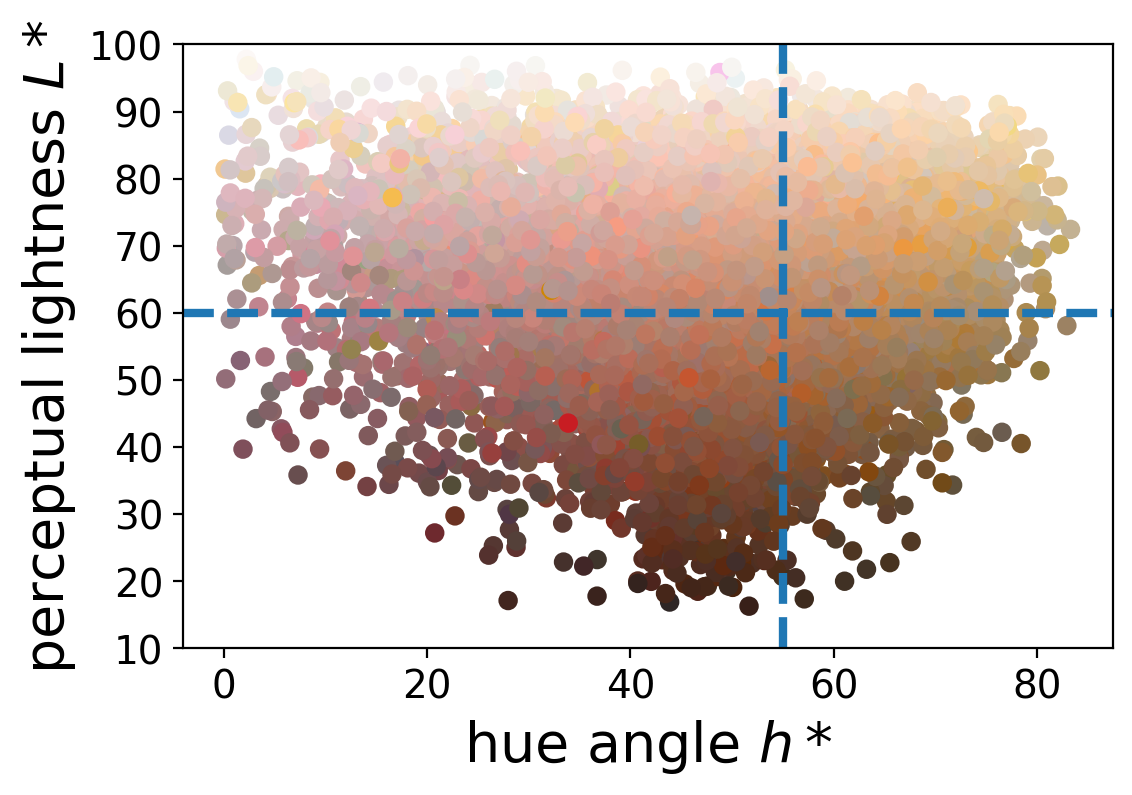}
  \caption{CelebAMask-HQ}
  \end{subfigure}\hfill
  \begin{subfigure}{.5\linewidth}
  \centering
  \includegraphics[width=\linewidth]{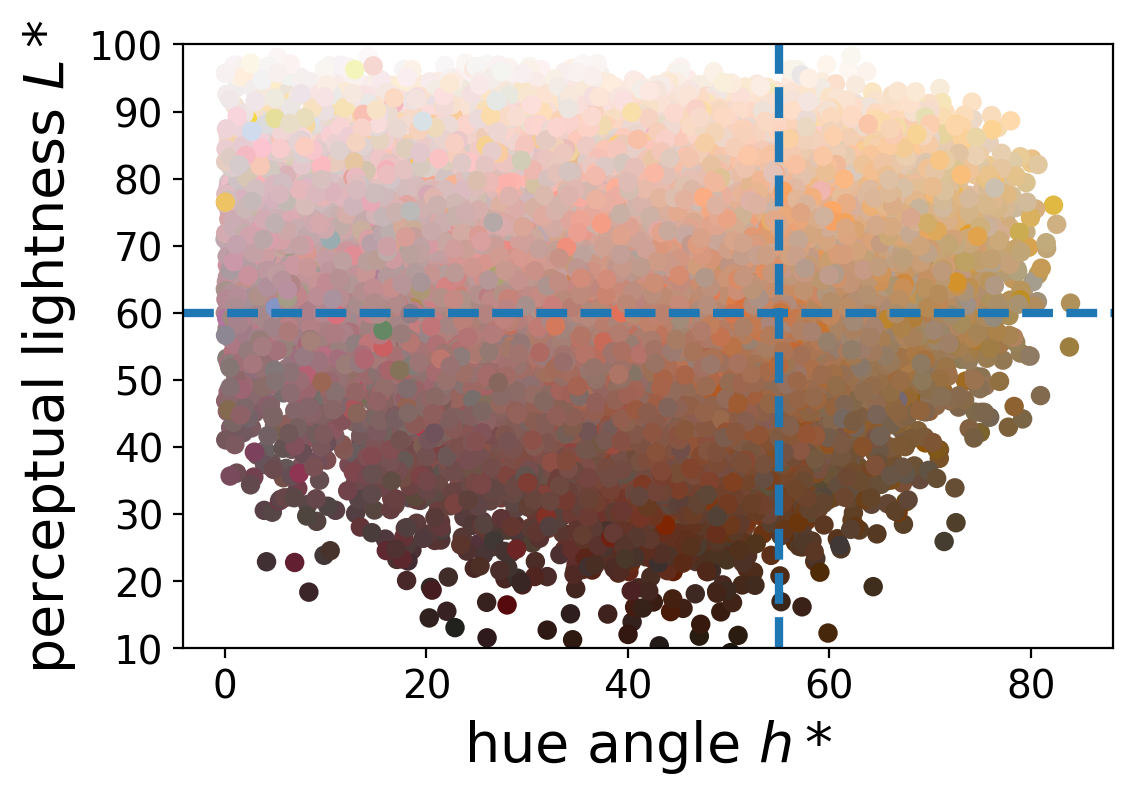}
  \caption{FFHQ}
  \end{subfigure}\hfill
  \caption{\textbf{Skin color distribution} on common face datasets.
  Every dot in the scatter plot corresponds to an image sample in the dataset.
  The skin tone threshold is at value 60 (light \vs dark), and the hue threshold at value 55\degree~(red \vs yellow).
  }
  \label{fig:illustration}
\end{figure}

Figure~\ref{fig:illustration} depicts the distribution in terms of perceptual lightness $L^*$ and hue angle $h^*$. When $L^*$  is over 60, it corresponds to a light skin tone (and conversely for a dark skin tone). When $h^*$ is over 55\degree, it corresponds to a skin turning towards yellow (and conversely for a skin turning towards red). Threshold values are taken from Ly~\etal~\cite{ly2020research}.

\begin{table}[t]
    \tablestyle{2pt}{1.1}
    \begin{subtable}[h]{0.48\linewidth}
        \centering
        \begin{tabular}{c c | c | c | c}
         & & \multicolumn{2}{c|}{\textbf{Skin tone}} \\
         & & Light & Dark & Total \\
        \hline 
        \multirow{2}{*}{\textbf{Hue}} & Red & \textbf{46.95} & 19.28 & \textbf{66.23} \\
        & Yellow & 27.48 & 06.29 & 33.77 \\
        \hline
        & Total & \textbf{74.43} & 25.57 & 100
       \end{tabular}
       \caption{CelebAMask-HQ}
       \label{tab:celeba}
    \end{subtable}
    \hfill
    \tablestyle{2pt}{1.1}
    \begin{subtable}[h]{0.48\linewidth}
        \centering
        \begin{tabular}{c c | c | c | c}
         & & \multicolumn{2}{c|}{\textbf{Skin tone}} \\
         & & Light & Dark & Total \\
        \hline 
        \multirow{2}{*}{\textbf{Hue}} & Red & \textbf{52.82} & 29.06 & \textbf{81.87} \\
        & Yellow &  13.44 & 04.68 & 18.12 \\
        \hline
        & Total & \textbf{66.26} & 33.74 & 100
       \end{tabular}
       \caption{FFHQ}
       \label{tab:ffhq}
    \end{subtable}
     \caption{\textbf{Skin color bias} in common face datasets (in \%). Images are skewed towards a light skin tone and a red skin hue while dark and yellow skin colors are underrepresented.}
     \label{tab:illustration}
\end{table}

Table~\ref{tab:illustration} provides a quantitative assessment of the skin tone and the skin hue. As commonly hypothesized in the literature~\cite{karras2019style,karras2021stylegan3}, both datasets are skewed towards light skins. This paper enables to quantify such skin color bias. Measuring the hue angle further shows that both datasets are also skewed towards red skins.
We also explore in Section~\ref{sec:app:eth} the skin color distribution per ethnicity along the tone and hue dimensions. We show the relevance of skin hue to better represent different ethnical groups, as some ethnicities can have a similar skin tone but different skin hue profile.
We also provide additional histogram plots in Section~\ref{sec:app:data}, where we go beyond a simple binary thresholding for both $L^*$ and $h^*$. The different thresholds confirm the over-representation of light skin tones and red skin hues.

Additionally, we measure the skin color bias on the output of generative models trained on FFHQ. To achieve this, we generate 10,000 images with a generative adversarial network (StyleGAN3~\cite{karras2021stylegan3}) and a diffusion model (P2~\cite{choi2022perception}).
Table~\ref{tab:gan} shows that both StyleGAN3 and P2 are reproducing the skin color bias present in the FFHQ dataset, with P2 slightly amplifying the bias over the original dataset distribution.
In future dataset collections, we recommend tracking both skin tone and skin hue for skin color scores, such that all subgroups are well balanced, which in turn makes the dataset more diverse.

\begin{table}[t]
    \tablestyle{2pt}{1.1}
    \begin{subtable}[h]{0.48\linewidth}
        \centering
        \begin{tabular}{c c | c | c | c}
         & & \multicolumn{2}{c|}{\textbf{Skin tone}} \\
         & & Light & Dark & Total \\
        \hline 
        \multirow{2}{*}{\textbf{Hue}} & Red & \textbf{53.05} & 28.71 & \textbf{81.76} \\
        & Yellow & 13.46 & 04.78 & 18.24 \\
        \hline
        & Total & \textbf{66.51} & 33.49 & 100
       \end{tabular}
       \caption{GAN -- StyleGAN3}
       \label{tab:celeba}
    \end{subtable}
    \hfill
    \tablestyle{2pt}{1.1}
    \begin{subtable}[h]{0.48\linewidth}
        \centering
        \begin{tabular}{c c | c | c | c}
         & & \multicolumn{2}{c|}{\textbf{Skin tone}} \\
         & & Light & Dark & Total \\
        \hline 
        \multirow{2}{*}{\textbf{Hue}} & Red & \textbf{53.29} & 30.08 & \textbf{83.37} \\
        & Yellow & 12.01 & 04.62 & 16.63  \\
        \hline
        & Total & \textbf{65.30} & 34.70 & 100
       \end{tabular}
       \caption{Diffusion -- P2}
       \label{tab:ffhq}
    \end{subtable}
     \caption{\textbf{Skin color bias} in the output of generative models when trained on FFHQ (in \%, for 10,000 images). Both StyleGAN3 and P2 reproduce the skin color bias of FFHQ.}
     \label{tab:gan}
\end{table}

\subsection{Skin color bias in models}

\subsubsection{Saliency-based image cropping}

\begin{figure*}[t]
\centering
  \begin{subfigure}{.24\textwidth}
  \centering
  \includegraphics[height=140px]{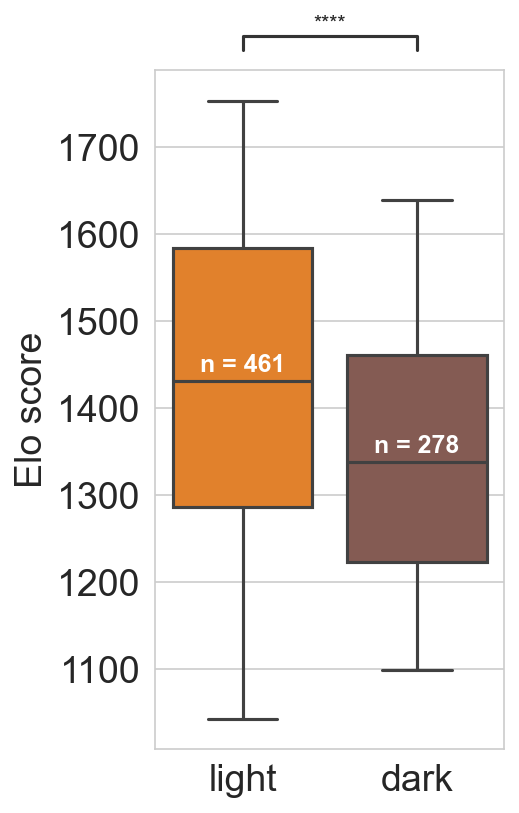}
  \caption{Skin tone}
  \label{fig:twitter:tone}
  \end{subfigure}\hfill
  \begin{subfigure}{.24\textwidth}
  \centering
  \includegraphics[height=140px]{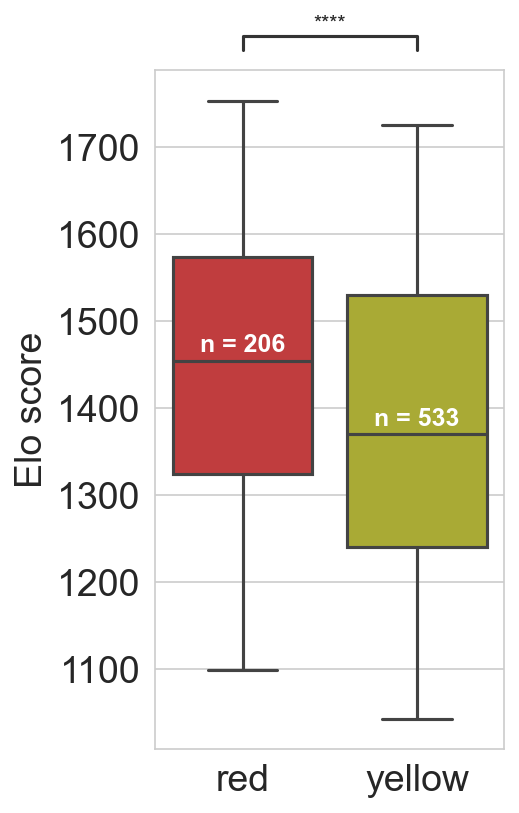}
  \caption{Skin hue}
  \label{fig:twitter:hue}
  \end{subfigure}\hfill
  \begin{subfigure}{.4\textwidth}
  \vspace{-10px}
  \centering
  \includegraphics[height=150px]{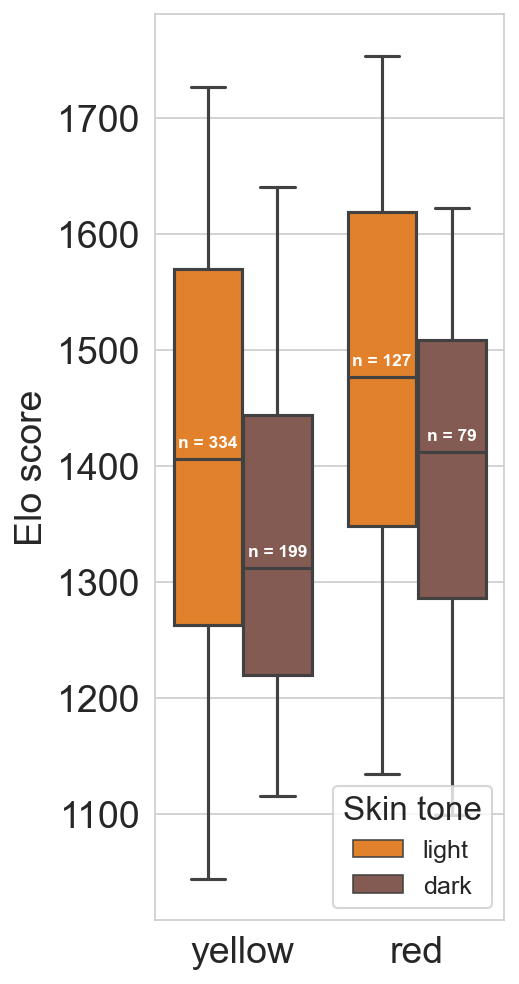}
  \includegraphics[height=150px]{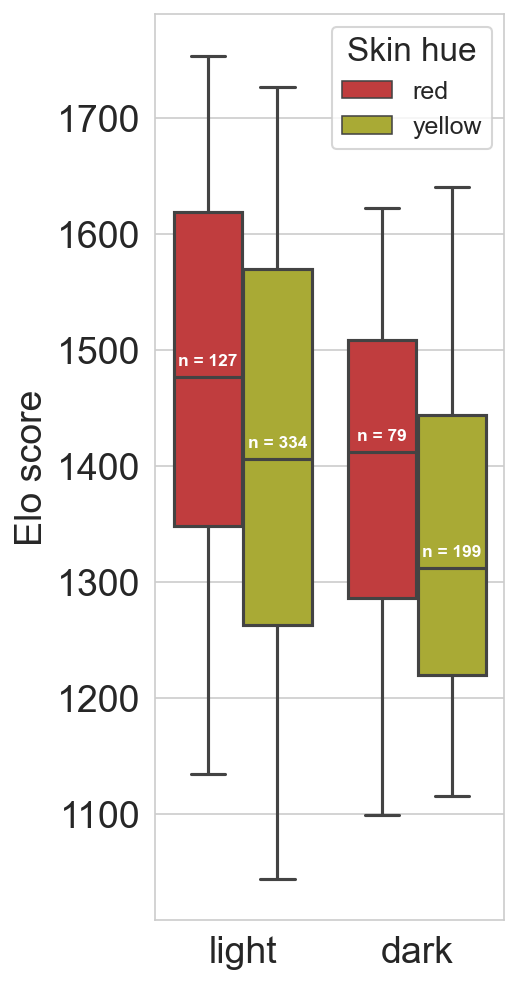}
  \caption{Intersection between skin tone and hue}
  \label{fig:twitter:intersection}
  \end{subfigure}
  \caption{\textbf{Saliency-based image cropping} on CFD, where the model is fed an image with two faces and predicts which one to keep. (a) Performance differences between light and dark skin tones are statistically significant ($p<0.0001$). (b) Differences between red and yellow skin hues are also statistically significant ($p<0.0001$). (c) The intersectional analysis also reveals statistically significant differences ($p<0.01$), except between light-yellow and dark-red skin colors. Complementary to the skin tone, the skin hue reveals additional differences in performance.}
  \label{fig:twitter}
\end{figure*}

\paragraph{Task.}
The saliency-based image cropping task produces a thumbnail based on a saliency map.
Such cropping algorithms are notably useful to select which region of a large image to display on a smaller screen~\cite{yan2013learning,deng2017image,wang2018deep}.
In the context of this paper, we are inspired by previous works~\cite{yee2021image,birhane2022auditing} and focus on cropping images with two faces. This setting can then be related to a pairwise comparison where the output decides which face to keep to produce the thumbnail.

Discrimination can happen in case the cropping algorithm favors a specific member of a protected attribute. For example, a method could consistently prefer a facial image of a light-skinned person over a dark-skinned one.
In fact, Yee~\etal~\cite{yee2021image} have shown that the Twitter cropping algorithm is prone to gender or skin tone biases, as well as male-gaze-like artifacts.
In this paper, we focus on assessing the skin color bias on the saliency-based image cropping task.

\paragraph{Method.}
To build a benchmark for saliency-based image cropping, we start from the one initially proposed by Birhane~\etal~\cite{birhane2022auditing}. We extend it to multiple race labels and propose a quantitative score to measure the probability of a face to be selected by the cropping algorithm.
We are given a dataset $\{x_i, a_i\}_{i=1}^N$ of $N$ facial images $x$ with their associated race label $a$.
We also extract skin color scales corresponding to perceptual lightness $y^{L^*}$ and the hue angle $y^{h^*}$ for every facial image.
To evaluate the cropping algorithm, we build images to include a unique pair of two facial images $x_i$ and $x_j$ with $i\neq j$.

We propose to compute the Elo rating~\cite{elo1978rating} for every facial image. Every pairwise comparison is considered as a game between two facial images $x_i$ and $x_j$. The objective is to get a rating for every facial image, \ie, $R_{i}$ and $R_{j}$, which indicates the probability of the face to be selected by the cropping algorithm. The probability for image $x_i$ to be chosen is defined as $p_{i}{=}{{1}/({1+10^{(R_{j}-R_{i})/M}})}$, where $M$ acts as a temperature for the sigmoid function. Intuitively, if there is a difference of $M$ points between $i$ and $j$, this means that $i$ has 10 times more chance to be chosen. Conversely, for image $x_i$ we have $p_{j}{=}1-p_{i}$.
The outcome of the cropping algorithm $S_{i}$ for $x_i$ is equal to 1 if $i$ wins and 0 if $j$ looses, and can be used to update the player score with $R_{i}^{\prime }=R_{i}+K(S_{i}-p_{i})$, and conversely for $x_j$.
Following common practice in chess playing~\cite{elo1978rating}, we set $M=400$ and $K=16$ and we initialize scores at 1400.

For benchmarking, we rely on Chicago Face Dataset (CFD)~\cite{ma2015chicago} and CFD-India~\cite{lakshmi2021india}, for a total of 739 unique facial images acquired in a controlled setting available for non-commercial research purposes. The dataset includes a self-reported gender label: 359 females and 380 males; as well as self-reported ethnic labels: 109 Asians, 197 Blacks, 142 Indians, 108 Latinos, 183 Whites. Individuals have given their informed consent for data collection.
We extract the skin masks with DeepLabV3~\cite{chen2018encoder}, trained on CelebAMask-HQ~\cite{lee2020CelebAMask} as done in FFHQ-Ageing~\cite{or2020lifespan}.
From CFD, we sample pairs of facial images that are equally distributed with respect to gender and ethnicity. The product of both gender and ethnicity label sets results in 10 intersectional groups, forming a total of 45 pairwise combinations. We sample 500 pairwise comparisons for each combination, for a total of 22,500. To form the final image, we concatenate both facial images vertically with a white space in between and preserve their aspect ratio.
This setting is highly inspired by Birhane~\etal~\cite{birhane2022auditing}; it differs by the inclusion of many ethnic groups instead of only two and it preserves the aspect ratio of images, as we noticed effects on the cropping algorithm when resizing images.

\paragraph{Results.}
Figure~\ref{fig:twitter} presents the results of the saliency-based cropping algorithm of Twitter on CFD.
We report the Elo score of the 739 individuals after 22,500 pairwise comparisons for image cropping.
A high Elo score indicates a preference of the algorithm to select the individual to be kept for cropping.
We find that light skin tones are preferred over dark skin tones (Figure~\ref{fig:twitter:tone}) with an average Elo score indicating a 60.73\% preference, which is confirmed with a statistically significant independent \textit{t}-test ($p<0.0001$).
The algorithm also prefers red skin hues over yellow ones with an average Elo score indicating a 58.25\% preference, which again is statistically significant (Figure~\ref{fig:twitter:hue}).
We further consider intersectional groups of skin tone and hue (Figure~\ref{fig:twitter:intersection}). When performing independent \textit{t}-tests with a Bonferroni correction, we find that all pairwise groups are statistically different ($p<0.01$), except for the light-yellow and dark-red skin colors.
Overall, this benchmark reveals a multidimensional hierarchy on the skin color preference of the Twitter cropping algorithm, with light-red skin colors being favored and dark-yellow skin colors being disfavored.

\subsubsection{Face verification}

\paragraph{Task.}
The face verification task compares a pair of facial images to verify whether they belong to the same individual or not.
In the context of this paper, we evaluate non-commercial and publicly available facial recognition models.
Note that, similar to the previous experiment, we focus on exposing potential skin color biases rather than improving the selected face verification models.

\paragraph{Method.}
We adopt a standard benchmark in face verification, where models are evaluated on their accuracy to predict whether a pair of facial images corresponds to the same individual.
Following previous works (\eg,~\cite{deng2019arcface,schroff2015facenet}), we evaluate on the Labeled Faces in the Wild (LFW) dataset~\cite{LFWTech}, which contains 1,000 test pairs (500 positives and 500 negatives).
We rely on the pre-processed version of LFW available in scikit-learn~\cite{pedregosa2011scikit}, and on the DeepFace repository\footnote{\href{https://github.com/serengil/deepface}{https://github.com/serengil/deepface}} to run ArcFace~\cite{deng2019arcface}, FaceNet~\cite{schroff2015facenet} and Dlib~\cite{dlib09} face recognition models.

\begin{table}[t]
\centering
\tablestyle{1.9pt}{1.1}
\begin{tabular}{l|c|cc|cc|cc|cc}
\multirow{3}{*}{Model}
& \multirow{3}{*}{Overall}
& \multicolumn{4}{c|}{Independent groups}
& \multicolumn{4}{c}{Intersectional groups} \\

& 
& \multicolumn{2}{c|}{Skin tone}
& \multicolumn{2}{c|}{Skin hue}
& \multirow{2}{*}{L+R} & \multirow{2}{*}{L+Y}
& \multirow{2}{*}{D+R} & \multirow{2}{*}{D+Y}
\\
 & & Light & Dark & Red & Yellow & & & &\\
\shline
ArcFace & 95.20 & \textbf{95.52} & 94.39 & \textbf{95.36} & 95.07 & \textbf{96.12} & 94.82 & 92.55 & \textbf{96.77} \\
FaceNet & 94.40 & \textbf{94.55} & 94.04 & \textbf{94.89} & 93.81 & \textbf{95.61} & 93.30 & 93.17 & \textbf{95.16}\\
Dlib & 94.20 & \textbf{94.41} & 93.68 & \textbf{94.71} & 93.58 & \textbf{94.83} & 93.90 & \textbf{94.41} & 92.74 \\
\end{tabular}
\caption{\textbf{Face verification} on LFW.
Individuals with a light skin tone or a red skin hue are better identified by all face verification models.
Intersectional results confirm this pattern for lighter tones, but differ for darker tones.
}
\label{tab:face-verification}
\end{table}

\begin{figure*}[t]
\centering
  \begin{subfigure}{\imgWidth\textwidth}
  \centering
  {\scriptsize $L^*{=}67.29$ $h^*{=}42.72$}
  \includegraphics[width=0.95\linewidth]{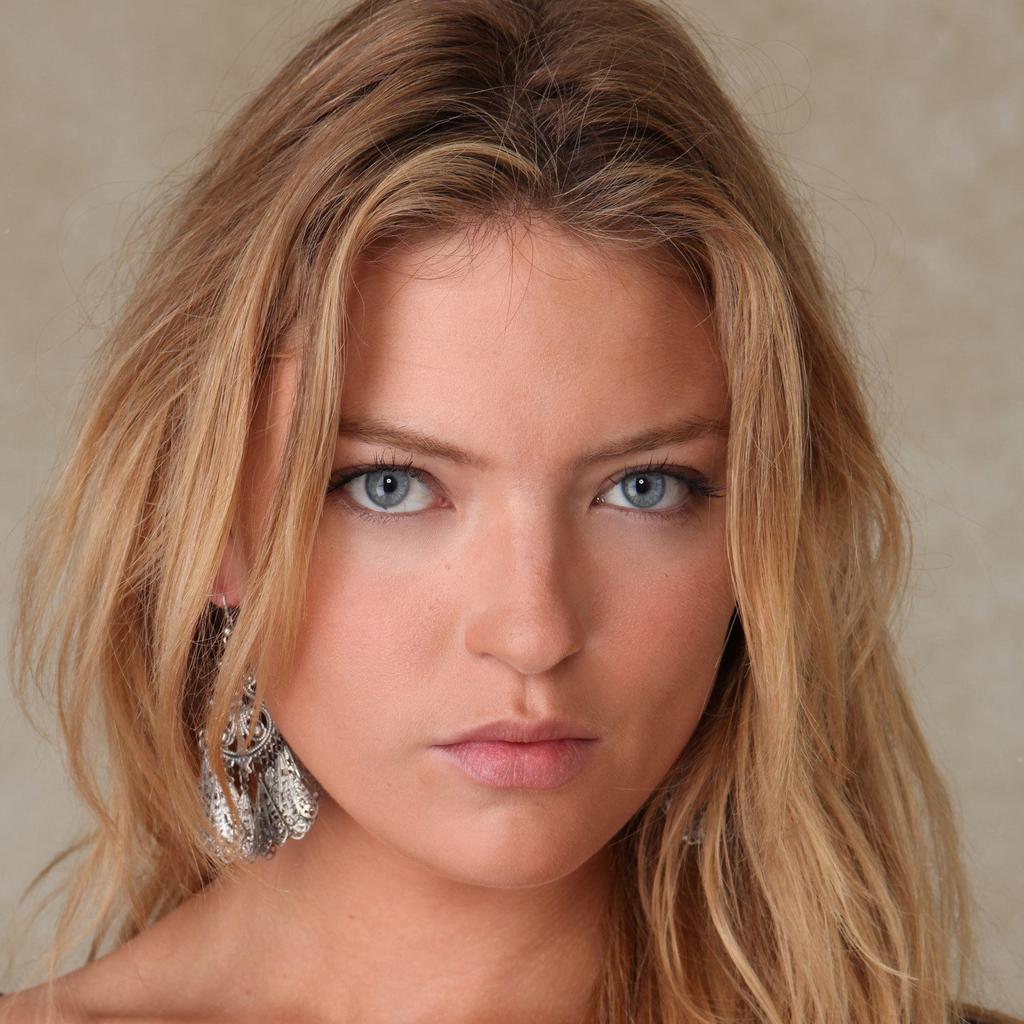}
  {\scriptsize $L^*{=}45.51$ $h^*{=}68.98$}
  \includegraphics[width=0.95\linewidth]{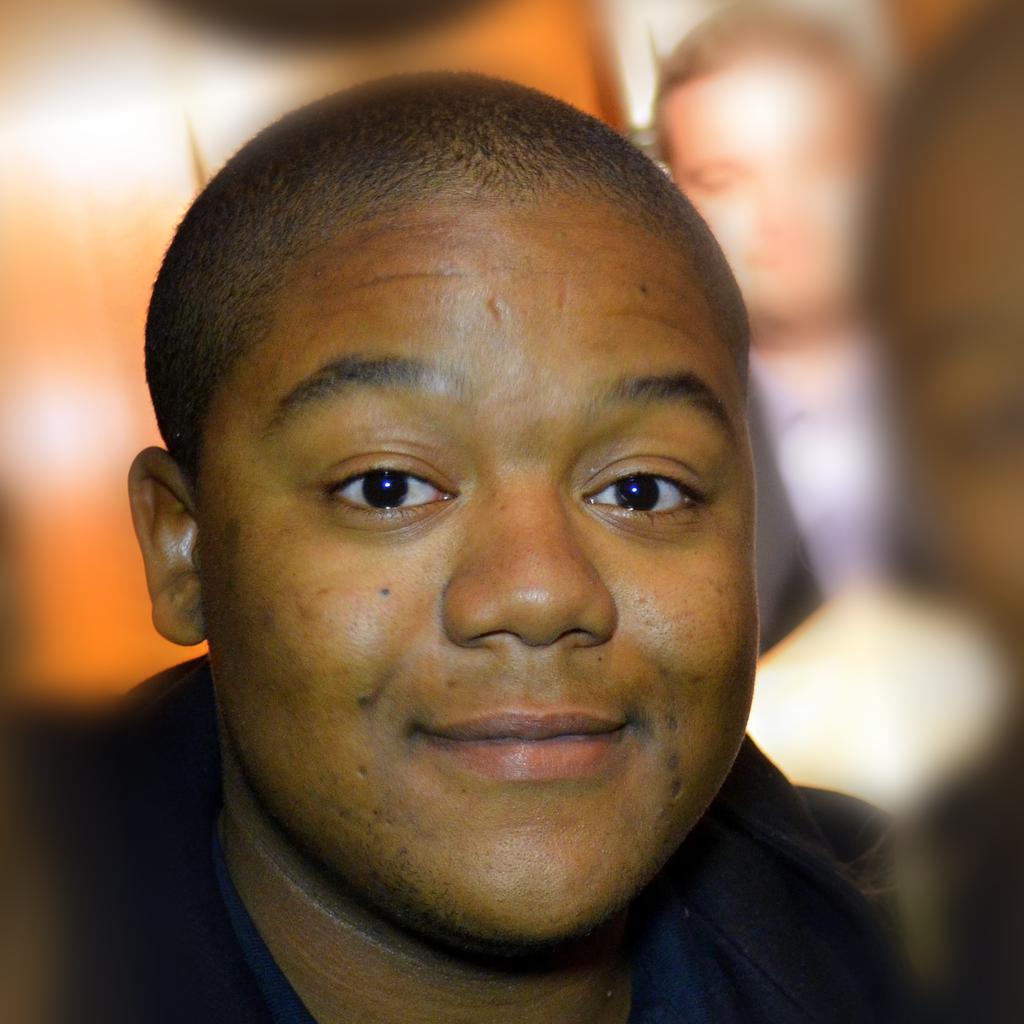}
  {\scriptsize $L^*{=}74.19$ $h^*{=}59.07$}
  \includegraphics[width=0.95\linewidth]{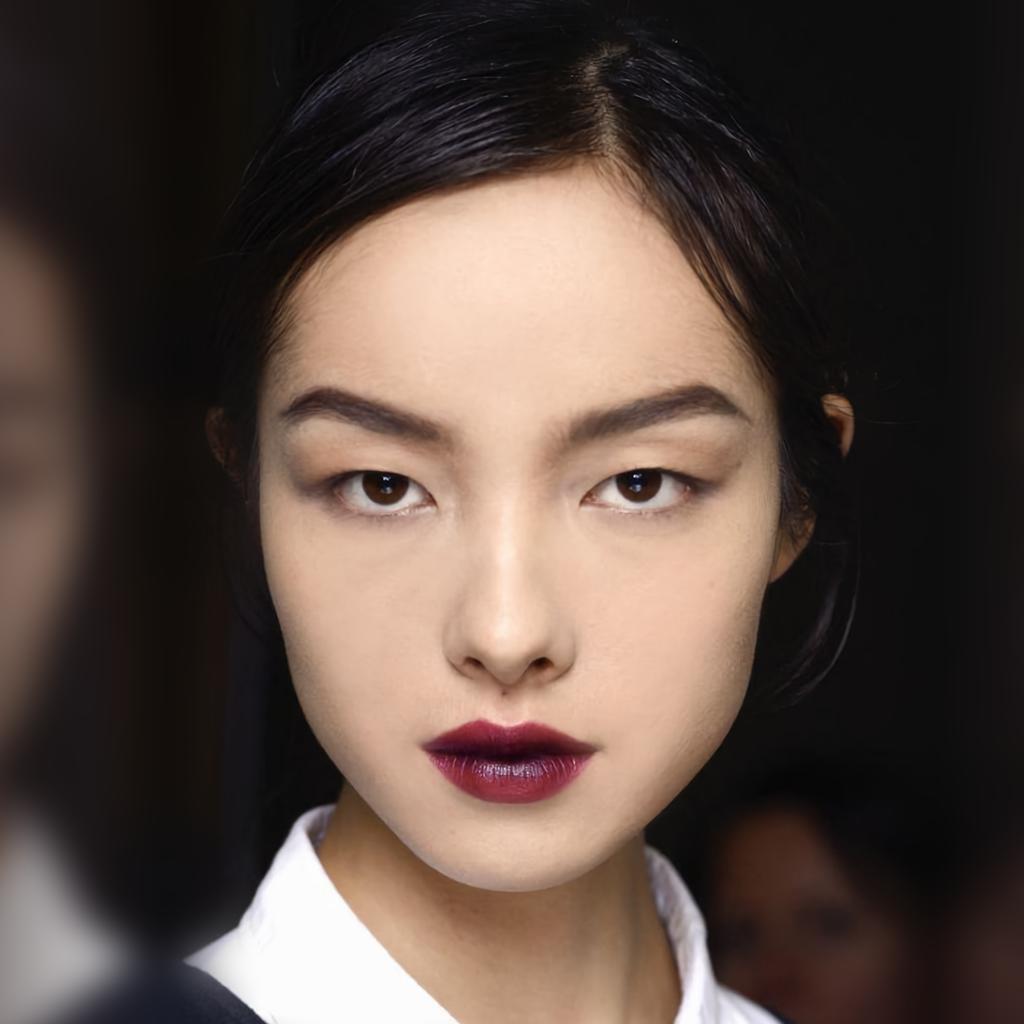}
  \caption{\footnotesize Original \label{fig:manipulations:ori}}
  \end{subfigure}\hfill
  \begin{subfigure}{\imgWidth\textwidth}
  \centering
  {\scriptsize $L^*{=}61.92$ $h^*{=}46.76$}
  \includegraphics[width=0.95\linewidth]{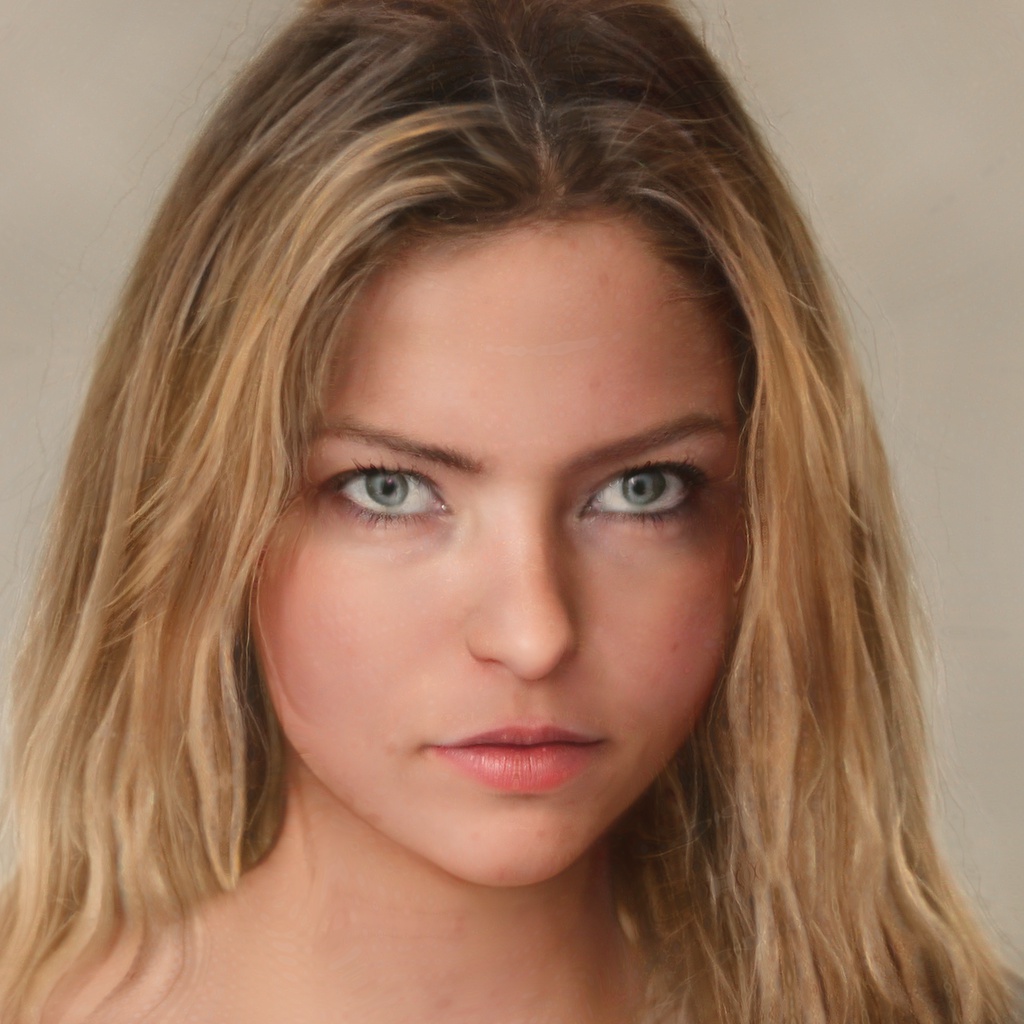}
  {\scriptsize $L^*{=}44.43$ $h^*{=}70.19$}
  \includegraphics[width=0.95\linewidth]{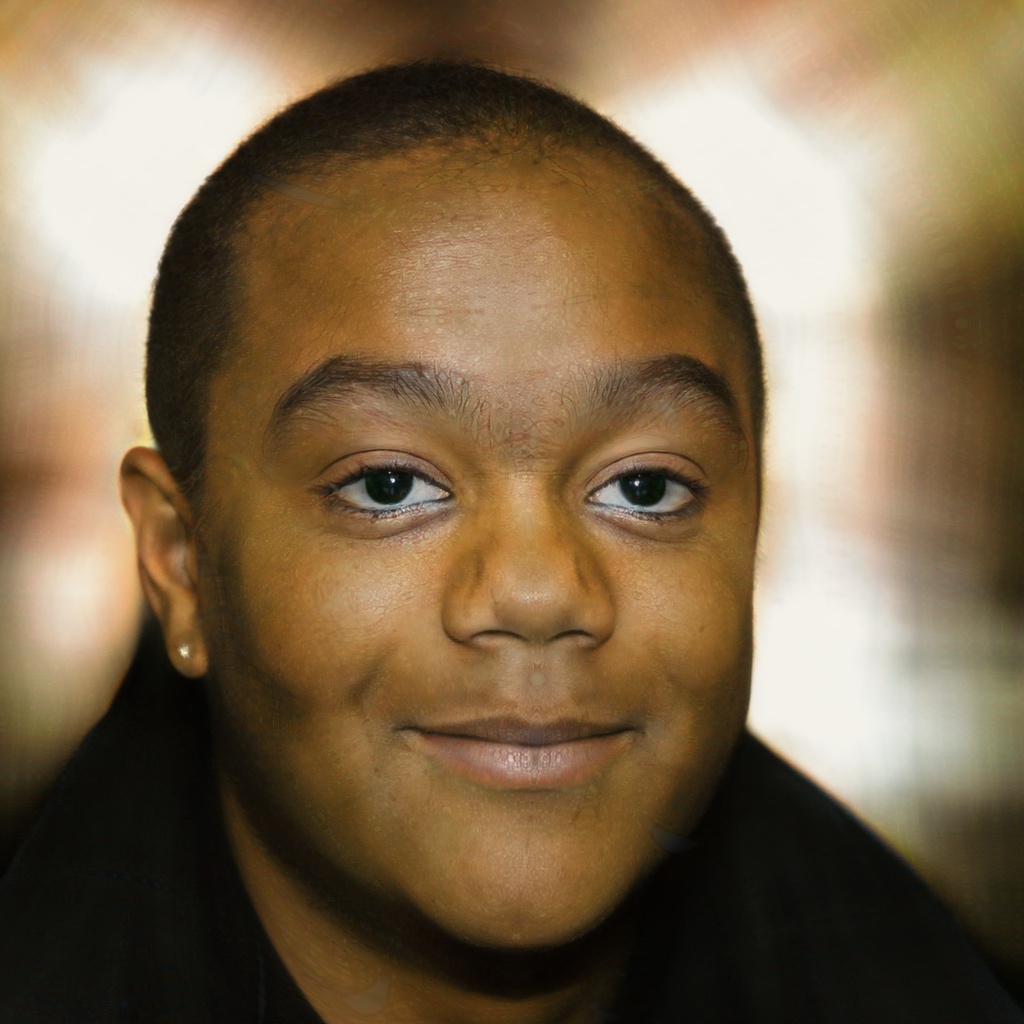}
  {\scriptsize $L^*{=}72.15$ $h^*{=}55.96$}
  \includegraphics[width=0.95\linewidth]{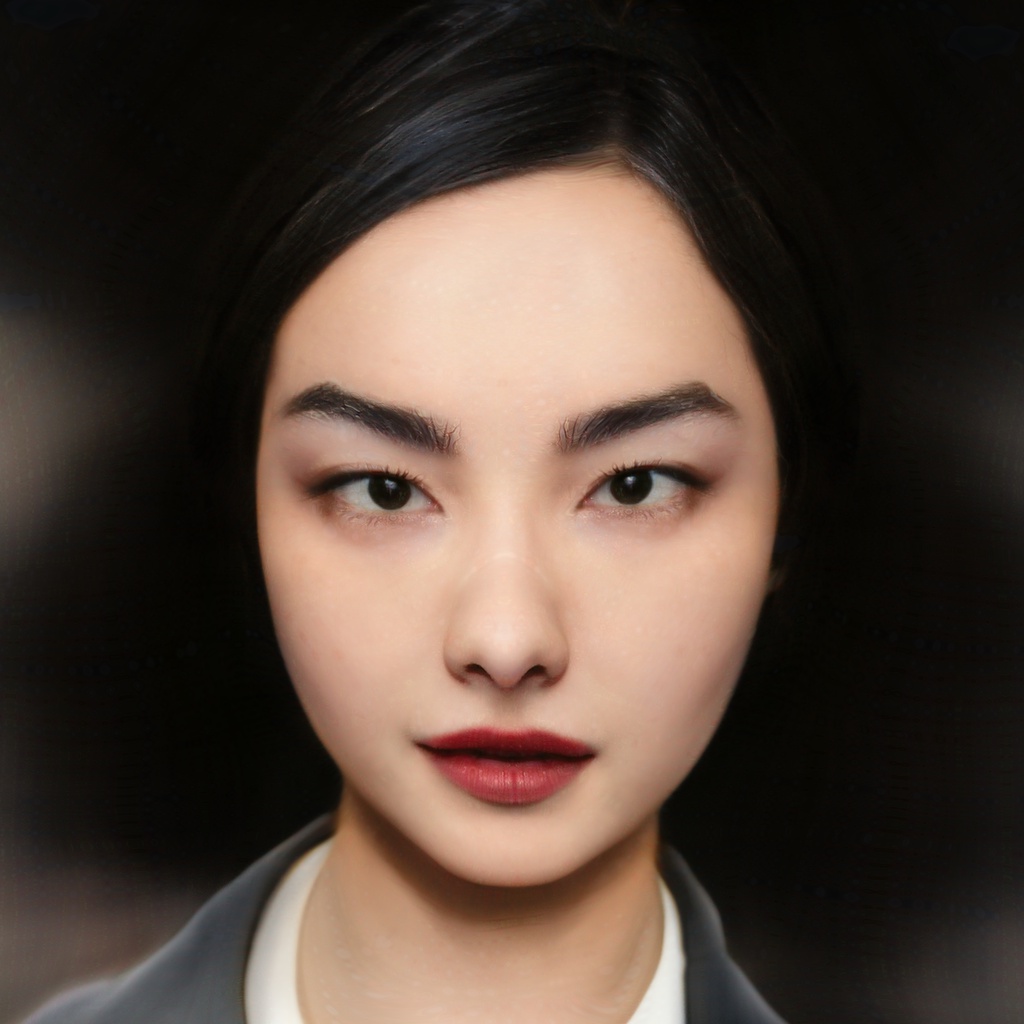}
  \caption{\footnotesize Reconstructed \label{fig:manipulations:recon}}
  \end{subfigure}\hfill
  \begin{subfigure}{\imgWidth\textwidth}
  \centering
  {\scriptsize $L^*{=}71.74$ $h^*{=}45.76$}
  \includegraphics[width=0.95\linewidth]{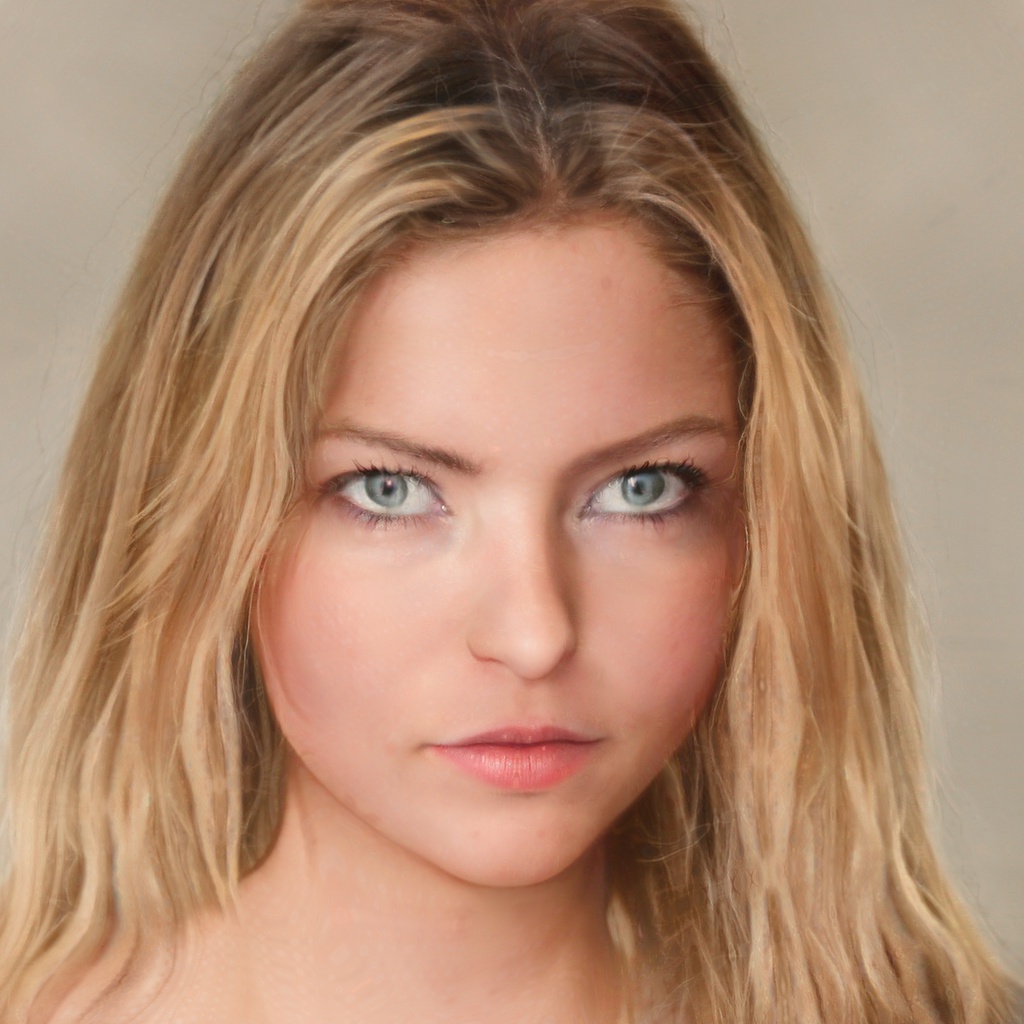}
  {\scriptsize $L^*{=}58.73$ $h^*{=}68.82$}
  \includegraphics[width=0.95\linewidth]{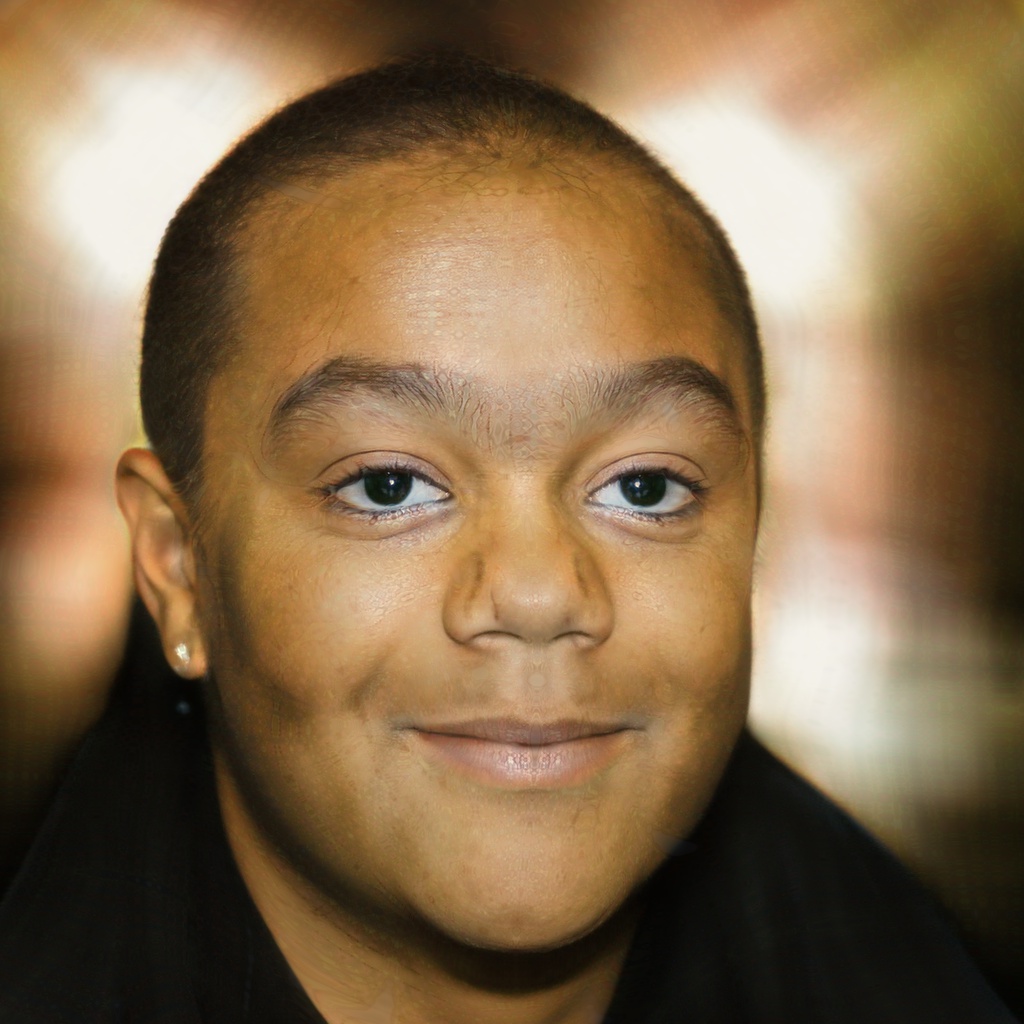}
  {\scriptsize $L^*{=}77.43$ $h^*{=}55.88$}
  \includegraphics[width=0.95\linewidth]{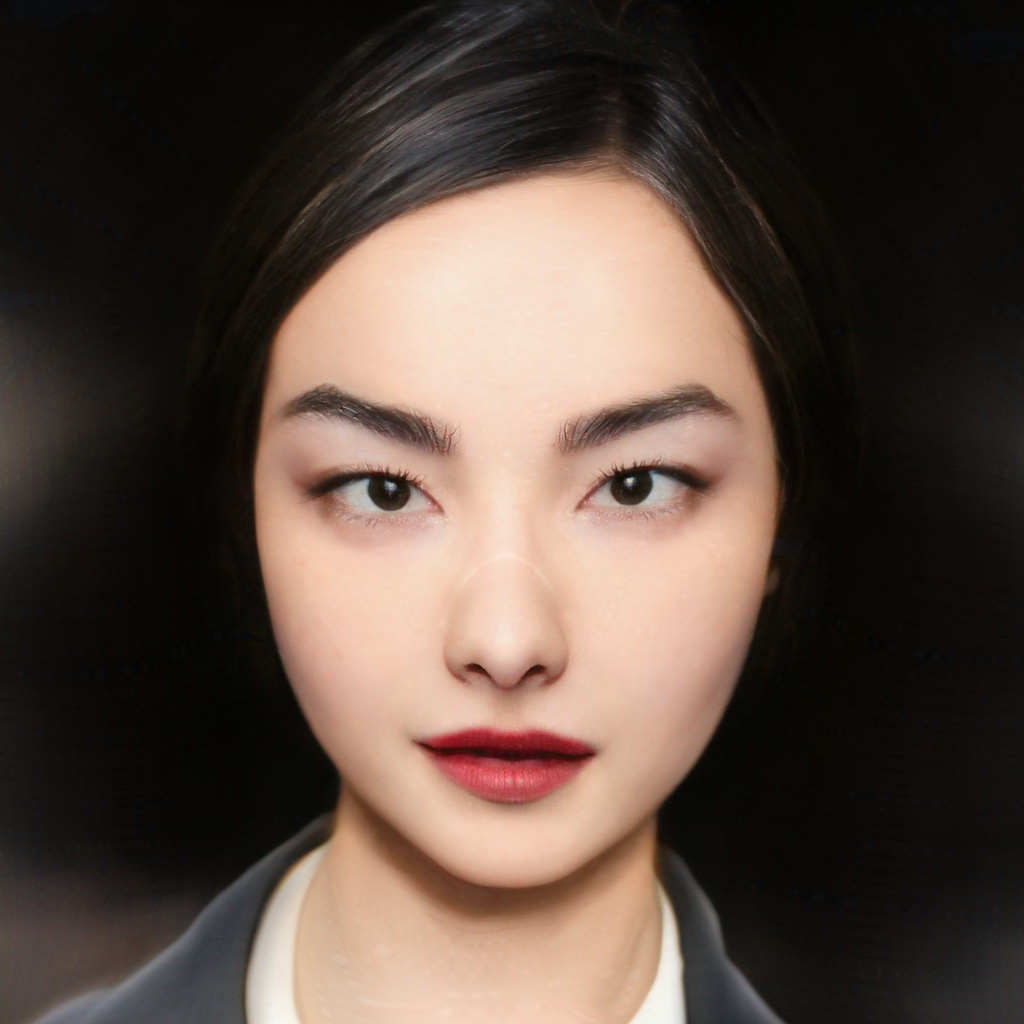}
  \caption{\footnotesize +light \label{fig:manipulations:tone:light}}
  \end{subfigure}\hfill
  \begin{subfigure}{\imgWidth\textwidth}
  \centering
  {\scriptsize $L^*{=}48.20$ $h^*{=}46.69$}
  \includegraphics[width=0.95\linewidth]{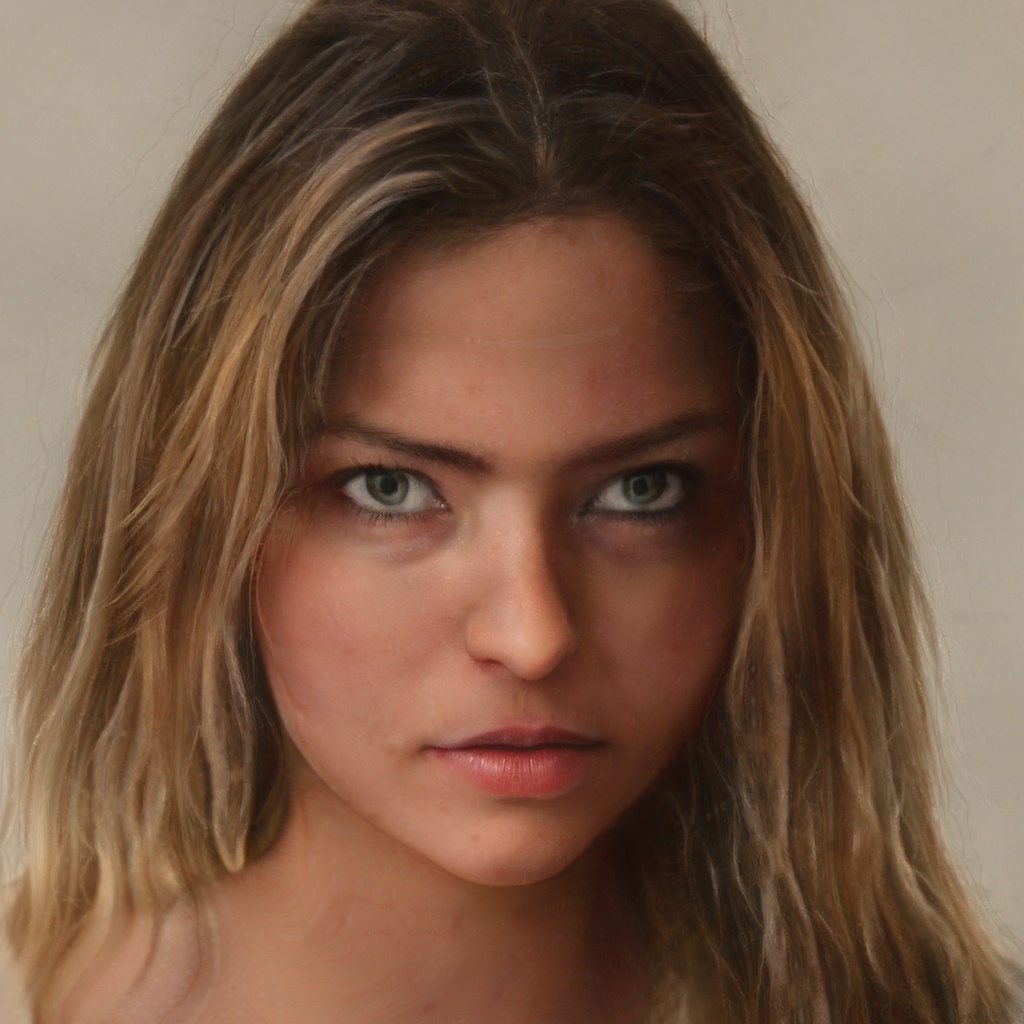}
  {\scriptsize $L^*{=}23.78$ $h^*{=}73.71$}
  \includegraphics[width=0.95\linewidth]{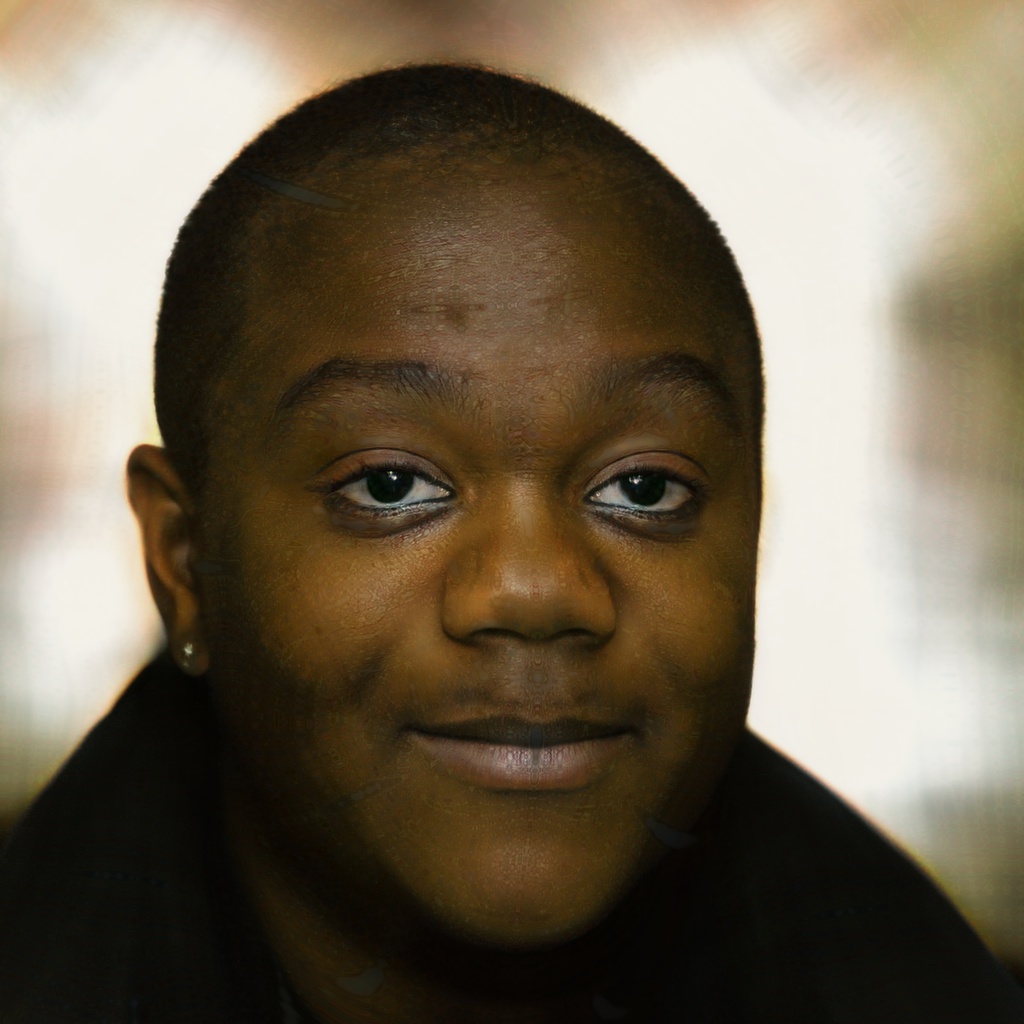}
  {\scriptsize $L^*{=}62.40$ $h^*{=}56.84$}
  \includegraphics[width=0.95\linewidth]{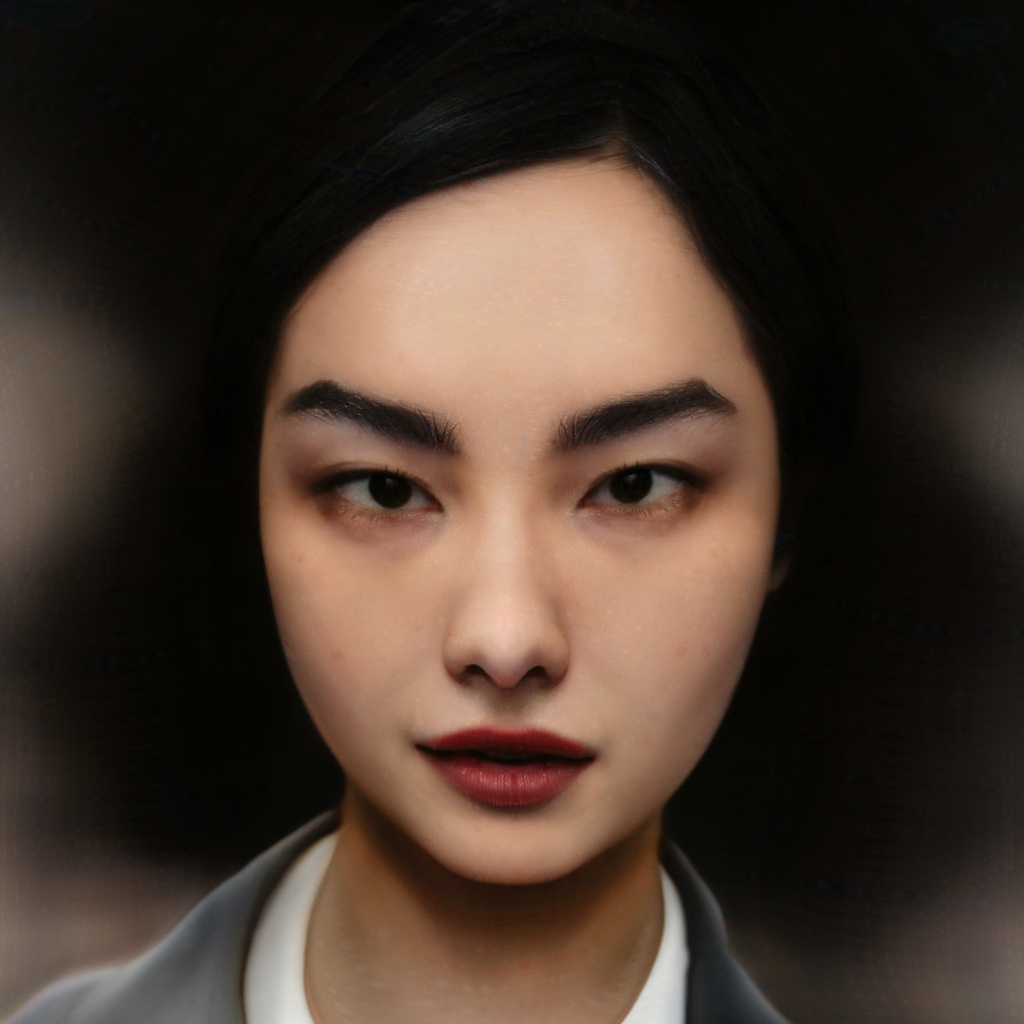}
  \caption{\footnotesize +dark \label{fig:manipulations:tone:dark}}
  \end{subfigure}\hfill
  \begin{subfigure}{\imgWidth\textwidth}
  \centering
  {\scriptsize $L^*{=}64.09$ $h^*{=}36.76$}
  \includegraphics[width=0.95\linewidth]{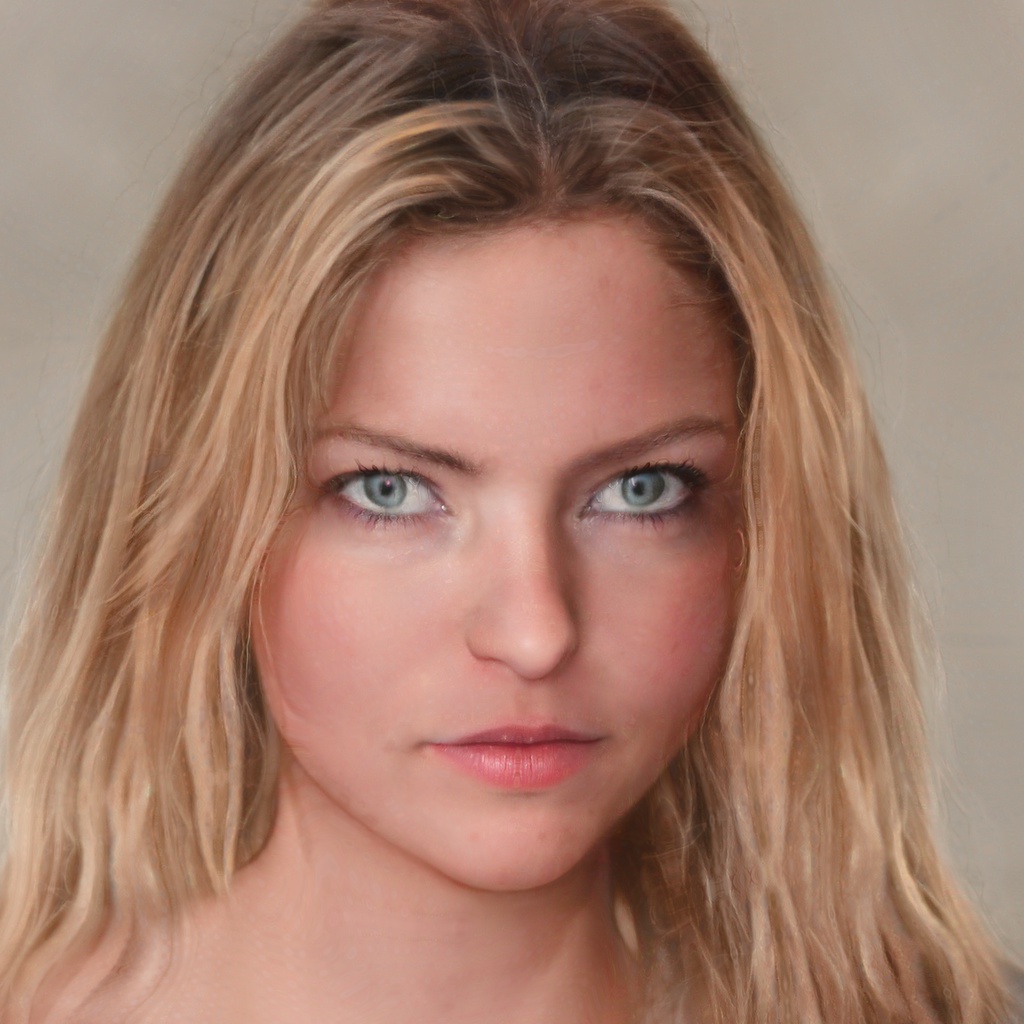}
  {\scriptsize $L^*{=}52.05$ $h^*{=}57.32$}
  \includegraphics[width=0.95\linewidth]{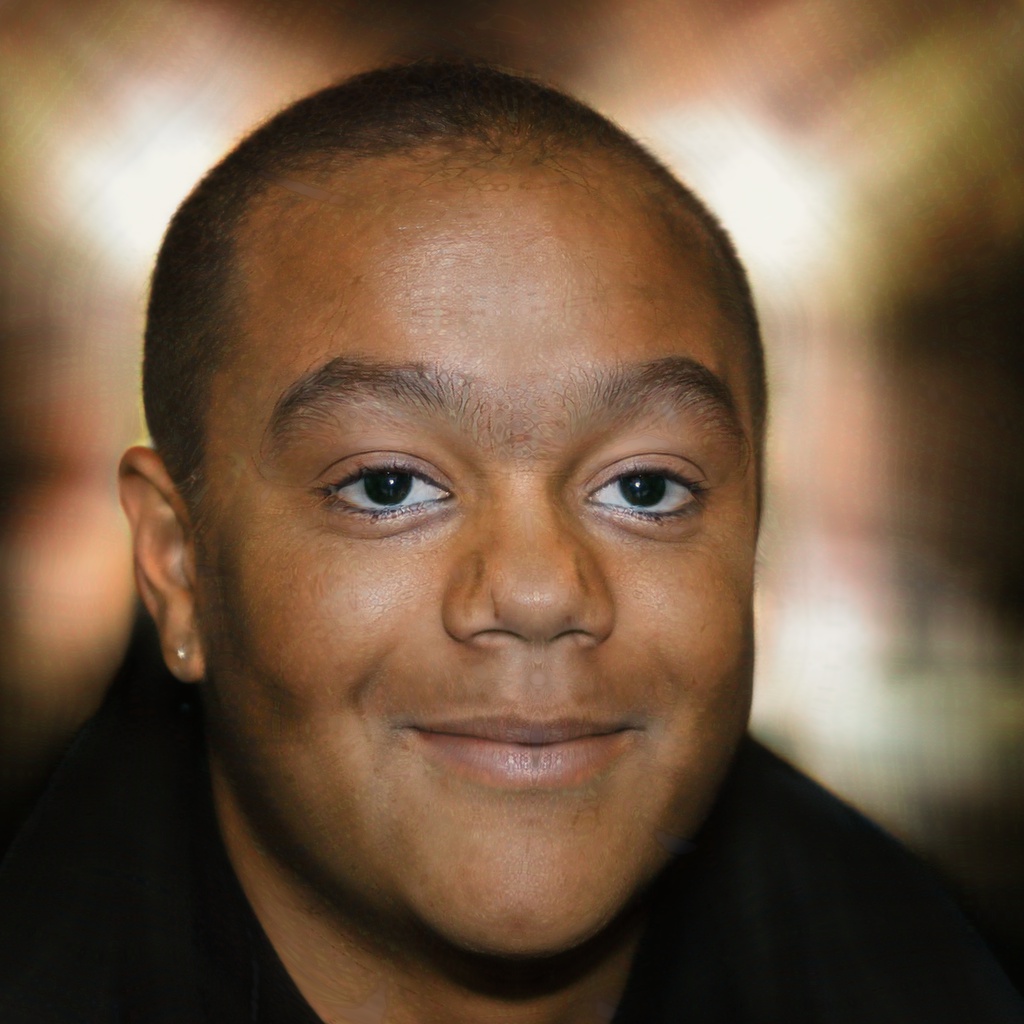}
  {\scriptsize $L^*{=}71.34$ $h^*{=}42.16$}
  \includegraphics[width=0.95\linewidth]{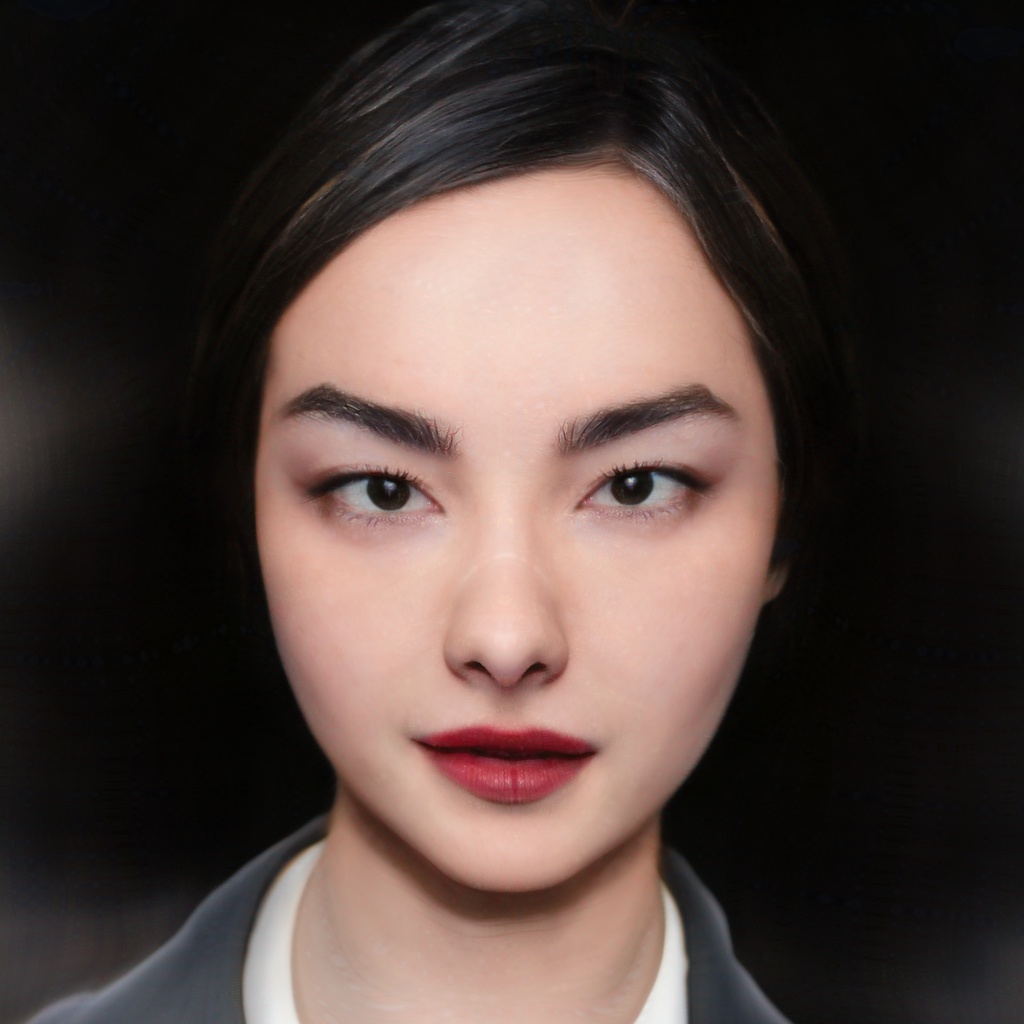}
  \caption{\footnotesize +red \label{fig:manipulations:hue:red}}
  \end{subfigure}\hfill
  \begin{subfigure}{\imgWidth\textwidth}
  \centering
  {\scriptsize $L^*{=}61.50$ $h^*{=}55.27$}
  \includegraphics[width=0.95\linewidth]{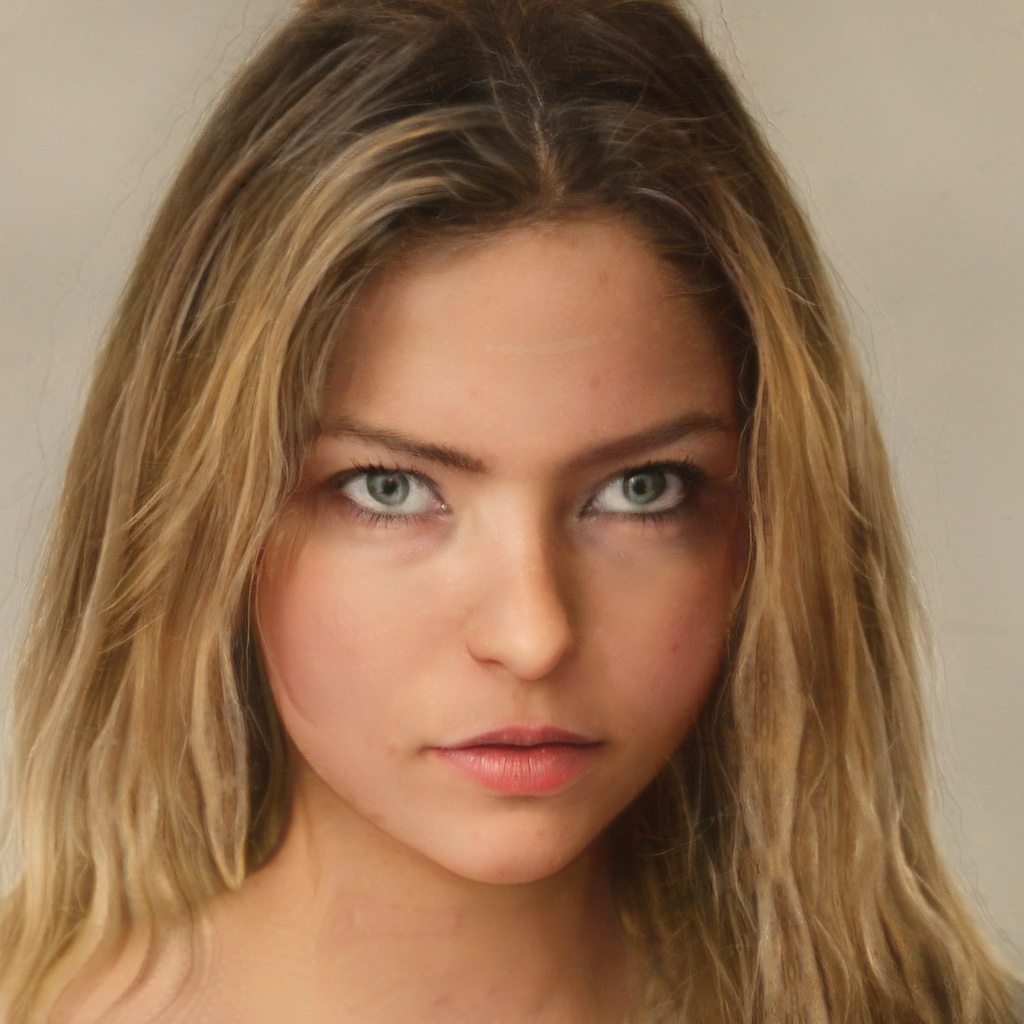}
  {\scriptsize $L^*{=}46.70$ $h^*{=}77.13$}
  \includegraphics[width=0.95\linewidth]{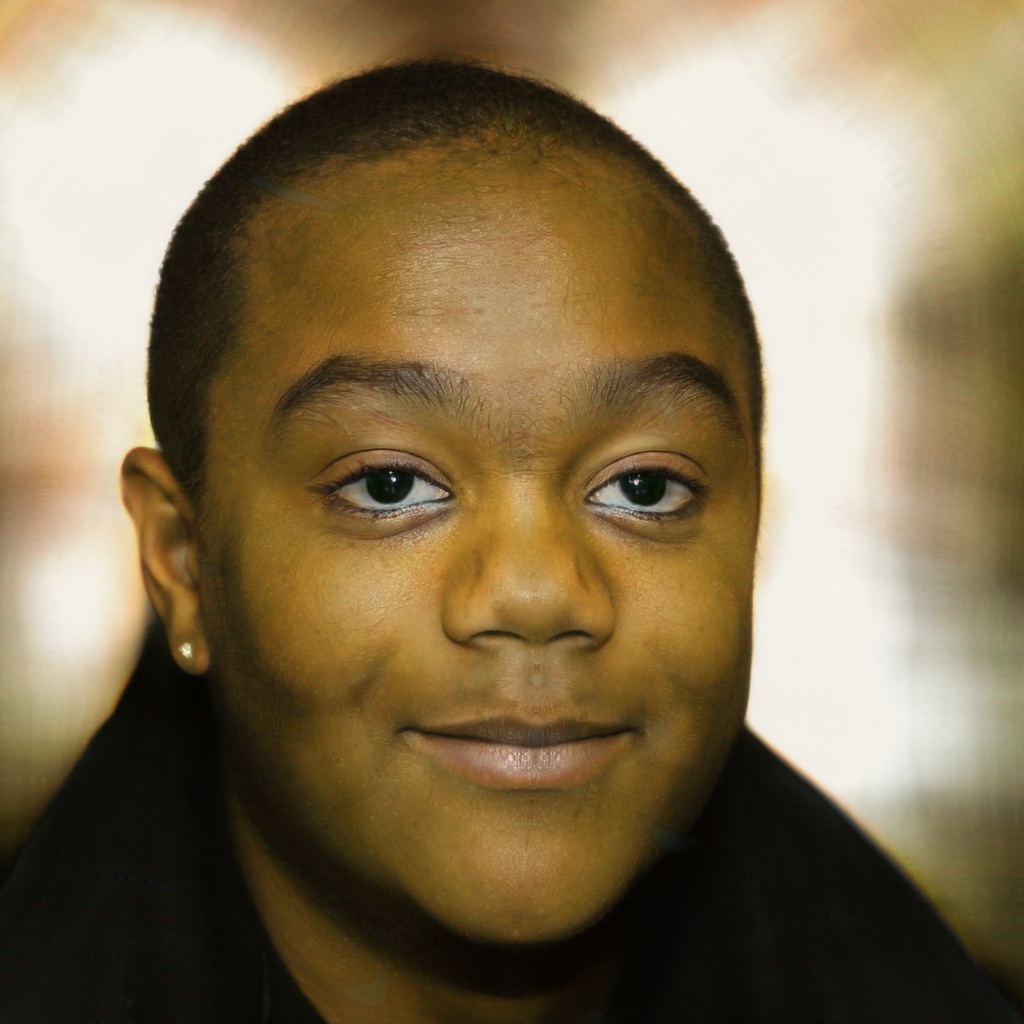}
  {\scriptsize $L^*{=}75.92$ $h^*{=}66.32$}
  \includegraphics[width=0.95\linewidth]{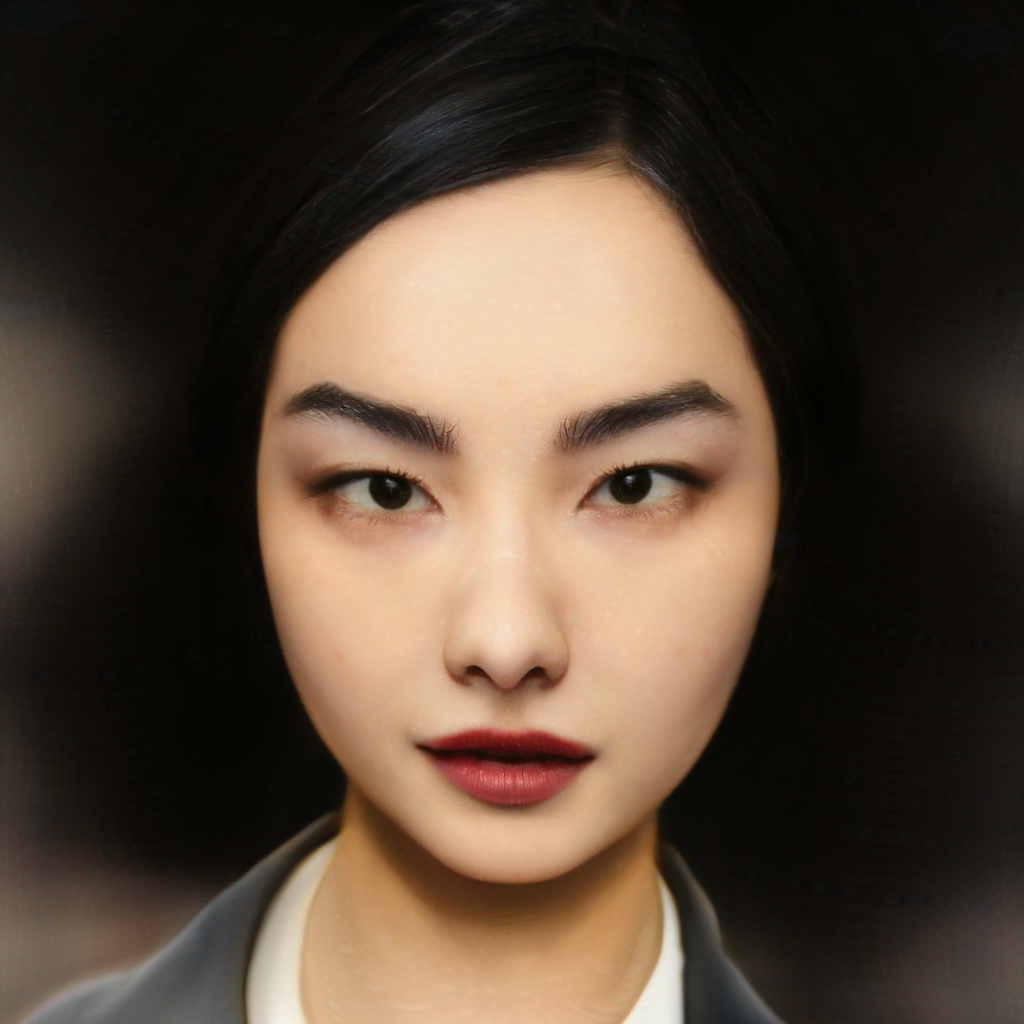}
  \caption{\footnotesize +yellow \label{fig:manipulations:hue:yellow}}
  \end{subfigure}
  \begin{subfigure}{\imgWidth\textwidth}
  \centering
  {\scriptsize $L^*{=}69.56$ $h^*{=}47.87$}
  \includegraphics[width=0.95\linewidth]{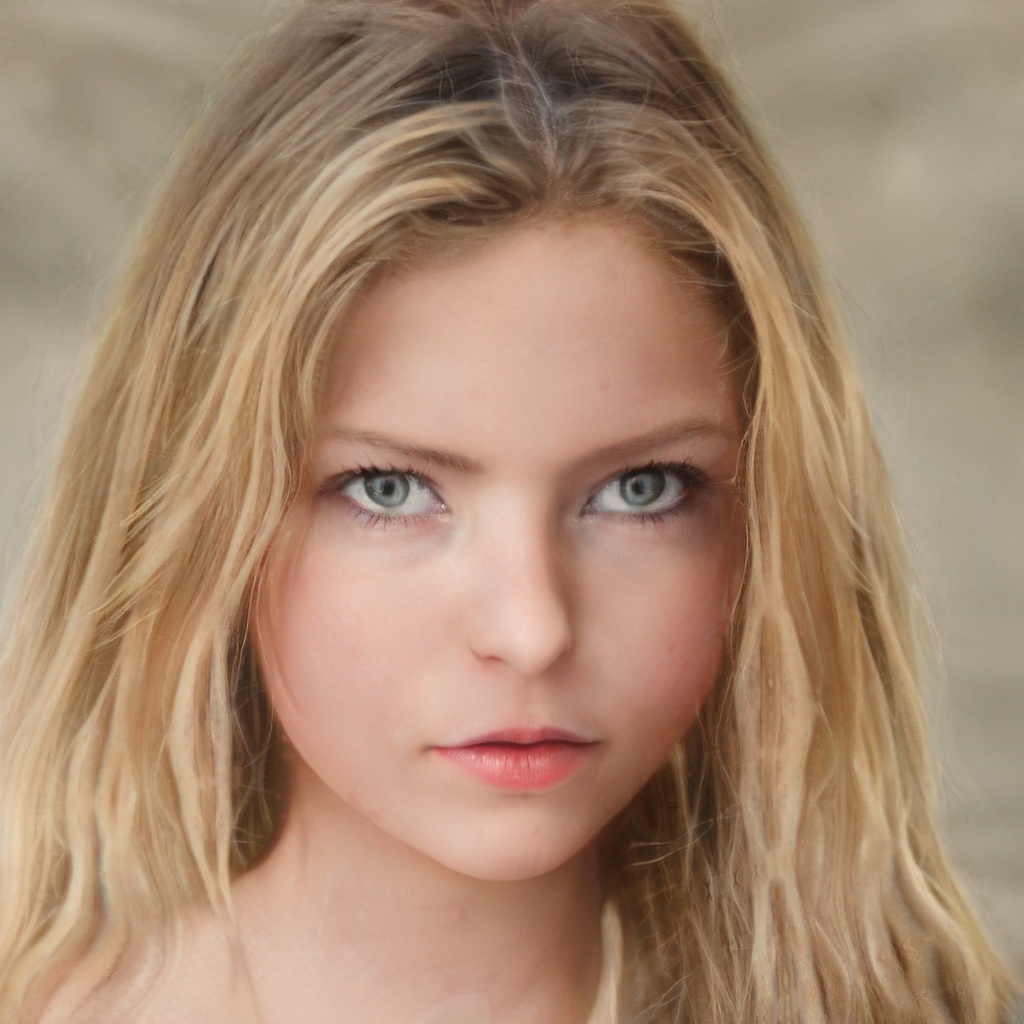}
  {\scriptsize $L^*{=}60.48$ $h^*{=}77.55$}
  \includegraphics[width=0.95\linewidth]{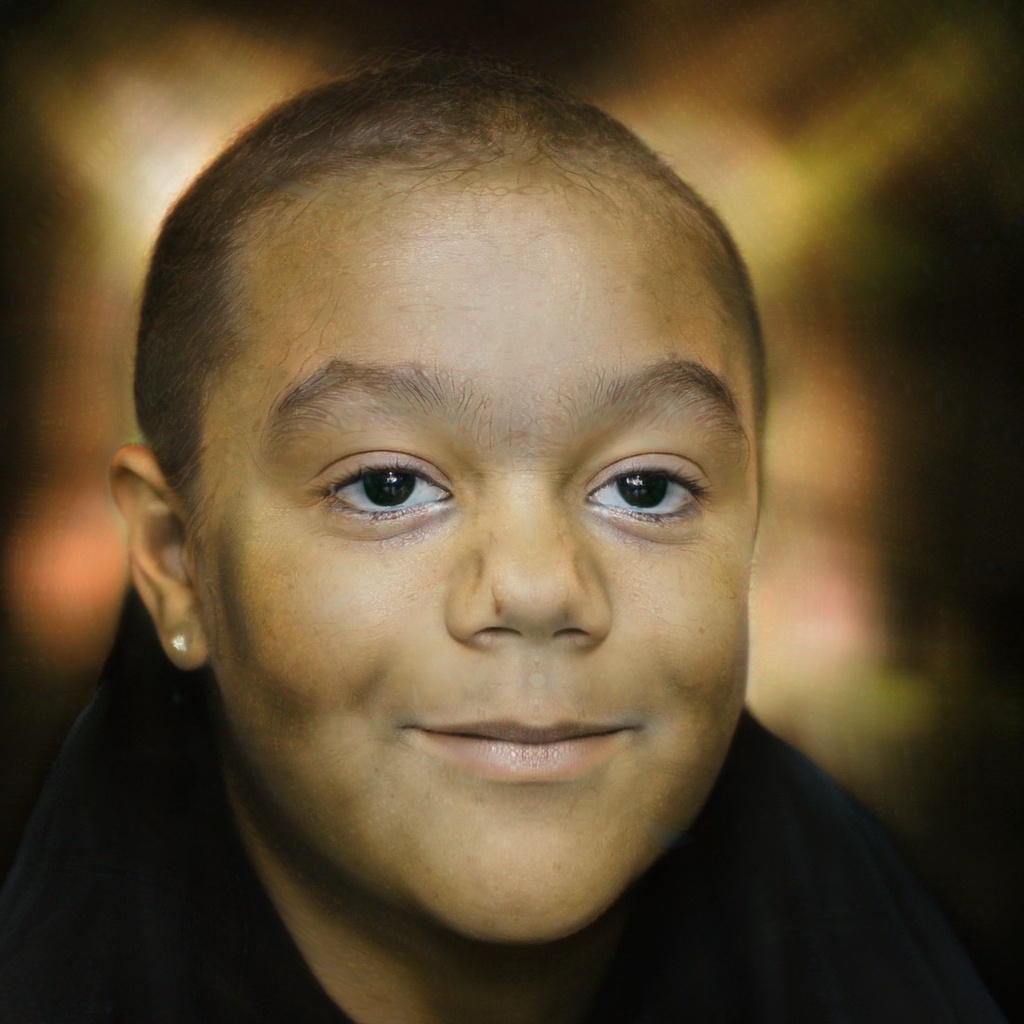}
  {\scriptsize $L^*{=}80.60$ $h^*{=}59.47$}
  \includegraphics[width=0.95\linewidth]{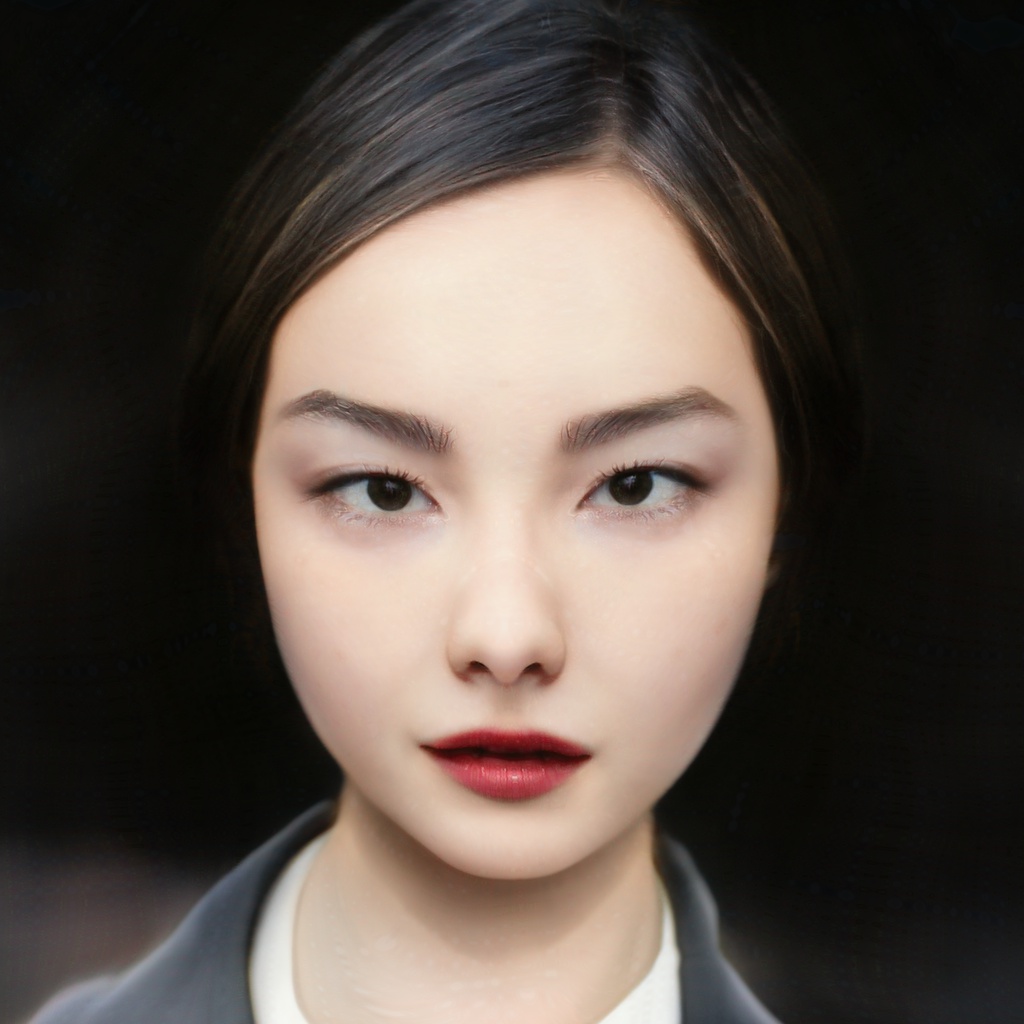}
  \caption{\footnotesize +pale \label{fig:manipulations:pale:plus}}
  \end{subfigure}\hfill
  \begin{subfigure}{\imgWidth\textwidth}
  \centering
  {\scriptsize $L^*{=}54.91$ $h^*{=}43.54$}
  \includegraphics[width=0.95\linewidth]{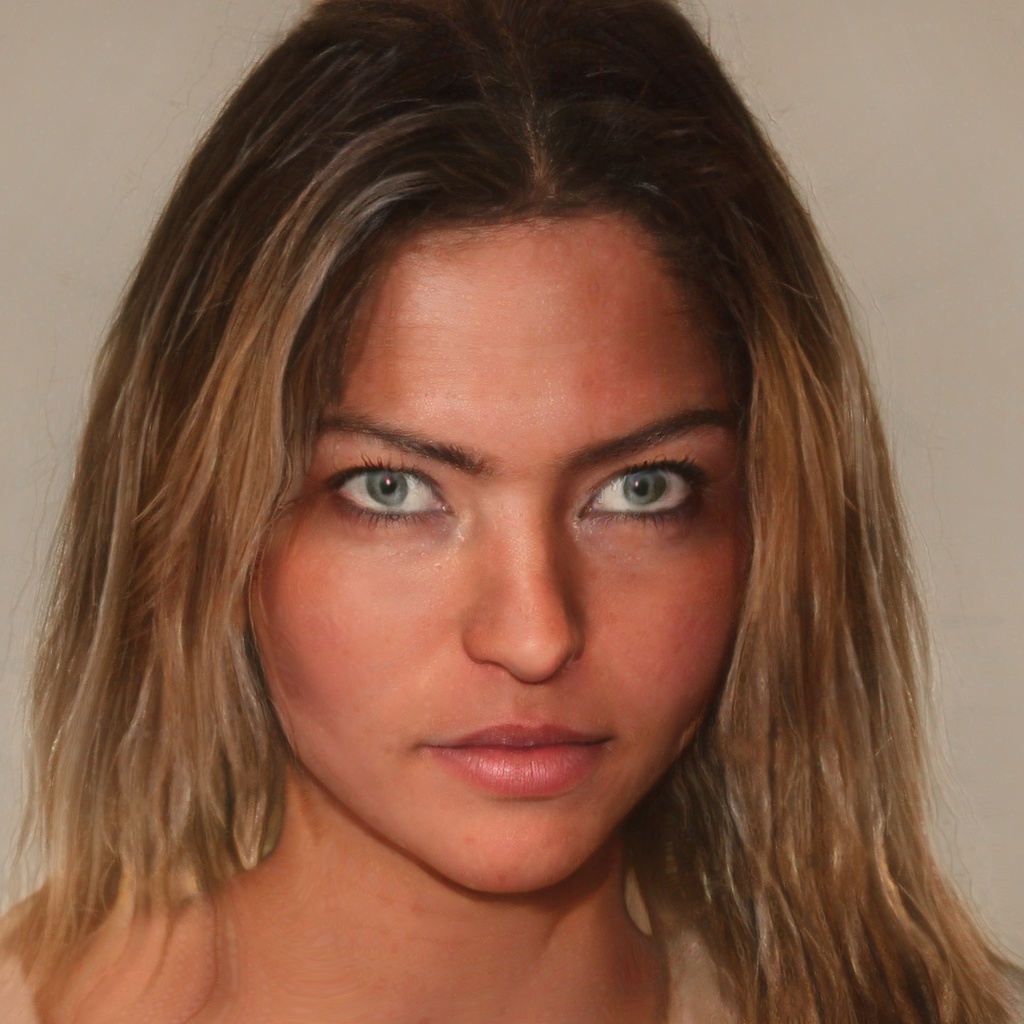}
  {\scriptsize $L^*{=}27.55$ $h^*{=}61.59$}
  \includegraphics[width=0.95\linewidth]{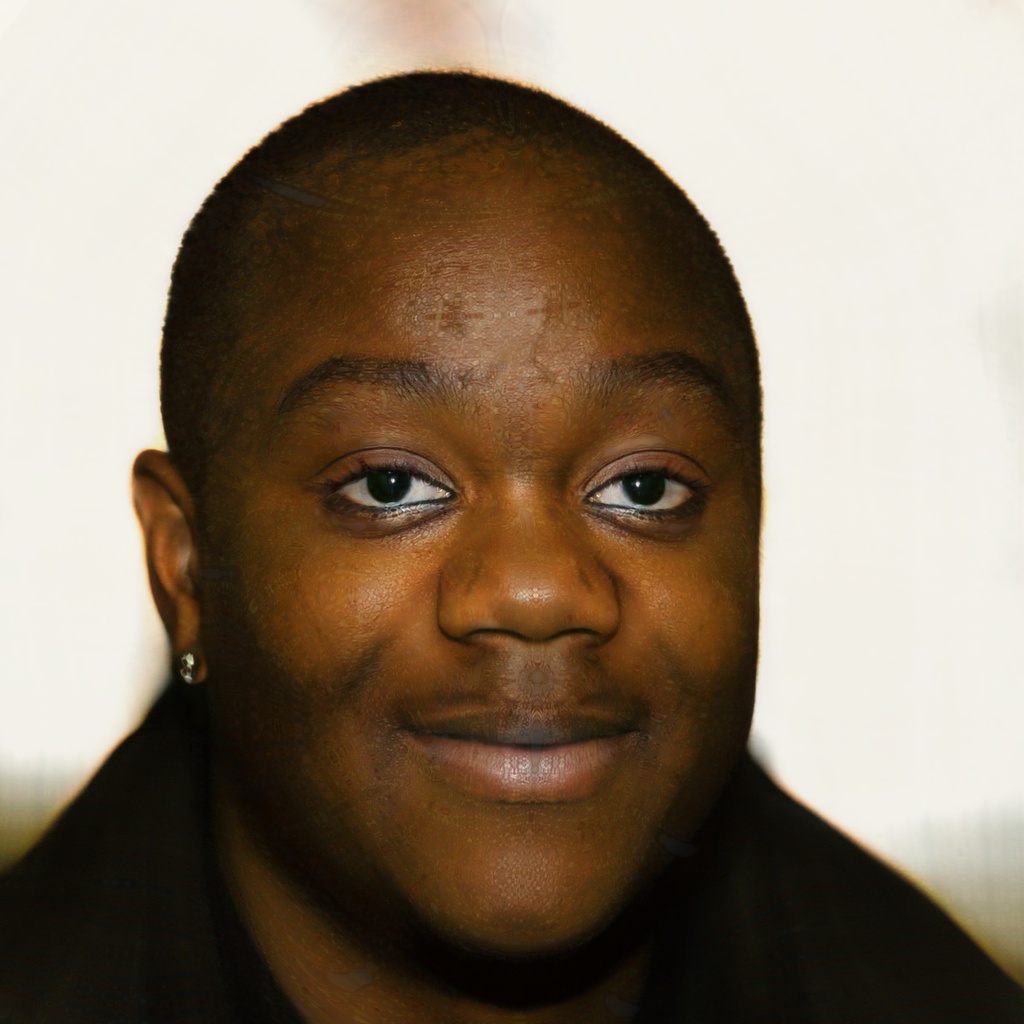}
  {\scriptsize $L^*{=}61.40$ $h^*{=}51.70$}
  \includegraphics[width=0.95\linewidth]{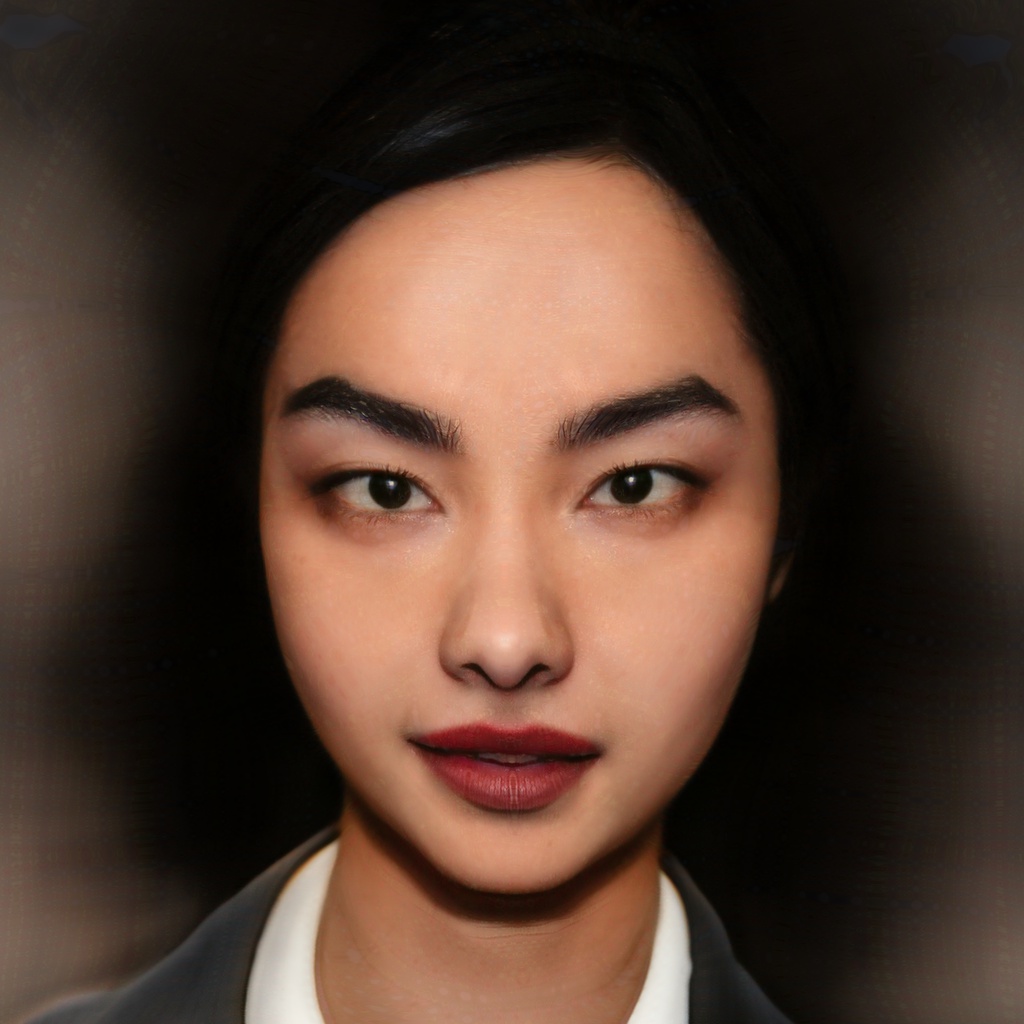}
  \caption{\footnotesize -pale \label{fig:manipulations:pale:minus}}
  \end{subfigure}
  \caption{\textbf{Skin color manipulation} on CelebAMask-HQ. By moving along specific directions in the latent space of StyleGAN3 (b), we manipulate the skin tone (c-d) or skin hue (e-f). When modifying the skin tone, this affects the perceptual lightness $L^*$ and preserves the hue angle $h^*$ (and conversely when modifying the skin hue).
  This differs from the attribute ``pale'' in CelebA, where a manipulation leads to changes in both $L^*$ and $h^*$ (g-h).
  As such, our proposal (c-f) is better for measuring the multidimensional causal effect of skin color as we have an independent control over $L^*$ and $h^*$.
  }
  \label{fig:manipulations}
\end{figure*}

\paragraph{Results.}
Table~\ref{tab:face-verification} presents the face verification results of several methods on LFW, broken down by skin tone (light \vs dark) and by skin hue (red \vs yellow).
All models tend to prefer light skin tones and red skin hues.
Specifically, both ArcFace and Dlib models are affected by skin tone differences, as they better verify the identify of light-skinned individuals. FaceNet has a different behaviour, it is more robust to skin tone differences but as prone to skin hue differences as Dlib.
When looking at intersectional groups, ArcFace and FaceNet perform lower for light and yellow skin colors, as well as dark and red skin colors; Dlib has a decreased performance of dark and yellow skin colors.
Overall, this benchmark confirms that in well-established methods for face verification, there exist performance differences in both skin tones and skin hues, which reiterates the importance of a multidimensional measure of skin color.

\subsection{Skin color causal effect in models}\label{sec:causal}

\paragraph{Task.}
Given a facial image $x$, the objective is to predict the presence or absence of an attribute $a$.
In the context of this paper, and following previous works~\cite{gebru2021datasheets,raji2019actionable,raji2020saving}, we focus on commercial systems, as well as publicly available models, to predict the gender and the presence of a smile.

Empowered with the ability to measure skin color quantitatively, we propose to manipulate an existing dataset by changing its skin tone and skin hue.
We are inspired by the idea of transects~\cite{balakrishnan2021towards} to reveal causal effects by manipulating one particular property in the image at a time. In the context of this paper, we are interested in manipulating skin color and observing its causal effect on attribute prediction performance.
For example, we modify all images in a dataset to have a lighter skin tone and compare its performance with respect to the original dataset version. Any discrepancy would then corresponds to the effect of a bias towards light skin tone in the model.

\begin{table*}[t]
\centering
\begin{subtable}[h]{\textwidth}
    \centering
    \tablestyle{4pt}{1.1}
    \begin{tabular}{l|ccc|ccc|ccc|ccc|ccc}
    \multirow{3}{*}{Model}
    & \multicolumn{3}{c|}{\multirow{2}{*}{Reconstructed}}
    & \multicolumn{6}{c|}{Skin tone}
    & \multicolumn{6}{c}{Skin hue} \\
    & & & & \multicolumn{3}{c|}{+light} & \multicolumn{3}{c|}{+dark} & \multicolumn{3}{c|}{+red} & \multicolumn{3}{c}{+yellow} \\
      & F & M & \textit{All} & F & M & \textit{All} & F & M & \textit{All} & F & M & \textit{All} & F & M & \textit{All} \\
    \shline
    AWS & 99.67 & 94.82 & 97.88 & 99.84 & 90.66 & 96.45 & 98.84 & 97.73  & 98.27 & 99.69 & 94.88  & 97.92 & 99.63 & 94.56 & 97.76 \\
    Azure & 99.78 & 95.80 & 98.31 & 99.91 & 92.59 & 97.22 & 99.11 & 97.72 & 98.60 & 99.75 & 96.01 & 98.37 & 99.78 & 95.45 & 98.18 \\
    Clarifai & 95.31 & 85.94 & 91.85 & 96.04 & 80.14 & 90.18 & 93.04 & 92.20 & 92.73 & 94.72 & 86.52 & 91.70 & 96.02 & 84.53 & 91.79\\
    FairFace & 99.49 & 92.91 & 97.06 & 99.77 & 87.56 & 95.27 & 98.21 & 96.57 & 97.61 & 99.52 & 93.10 & 96.97 & 99.42 & 92.77 & 97.16 \\
    \end{tabular}
    \caption{\textbf{Gender classification accuracy}, where F denotes female samples, M denotes male samples and \textit{All} denotes all samples.}
    \label{tab:attribute:gender}    
\end{subtable}
\newline
\begin{subtable}[h]{\textwidth}
    \centering
    \tablestyle{4pt}{1.1}
    \begin{tabular}{l|ccc|ccc|ccc|ccc|ccc}
    \multirow{3}{*}{Model}
    & \multicolumn{3}{c|}{\multirow{2}{*}{Reconstructed}}
    & \multicolumn{6}{c|}{Skin tone}
    & \multicolumn{6}{c}{Skin hue} \\
     & & & & \multicolumn{3}{c|}{+light} & \multicolumn{3}{c|}{+dark} & \multicolumn{3}{c|}{+red} & \multicolumn{3}{c}{+yellow} \\
      & S & NS & \textit{All} & S & NS  & \textit{All} & S & NS  & \textit{All} & S & NS  & \textit{All} & S & NS  & \textit{All}\\
    \shline
    AWS & 92.80 & 93.66 & 93.26 & 94.22 & 91.58 & 92.82 & 88.80 & 96.02 & 92.63 & 96.88 & 86.38  & 91.32 & 87.97 & 96.70 & 92.60 \\
    Azure & 97.29 & 80.84 & 88.57  & 98.13 & 76.36 & 86.60 & 95.59 & 86.11 & 90.57 & 99.00 & 69.76  & 83.51 & 95.18 & 87.22  & 90.97\\
    \end{tabular}
    \caption{\textbf{Smile classification accuracy}, where S denotes smiling samples, NS denotes non-smiling samples and \textit{All} denotes all samples.}
    \label{tab:attribute:smile}    
\end{subtable}
\caption{\textbf{Attribute prediction} on CelebAMask-HQ, with disaggregated results for positive and negative samples.
(a) In gender classification, models classify individuals as more feminine when their skin tone becomes lighter.
(b) In smile classification, models classify individuals as smiling when their skin tone becomes lighter or when their skin hue becomes redder.}
\label{tab:attribute:intersection}
\end{table*}

\paragraph{Method.}
To build a benchmark for attribute prediction, we propose to manipulate images with an image generation method. Specifically, we consider an encoder-decoder scheme, which encodes an image $x$ into a latent vector $w$, and then decodes $w$ to provide a reconstructed image $\hat{x}$.
Modifying $x$ then consists of moving in the latent space towards specific and meaningful directions $D$, \ie, $w + D$.

Following Alahuf~\etal~\cite{alaluf2022third}, we rely on e4e~\cite{tov2021designing} to encode images and on StyleGAN3~\cite{karras2021stylegan3} to generate images. Finding directions in the latent space relies on the InterfaceGAN~\cite{shen2020interpreting} method.
Concretely, we first train a classifier on CelebAMask-HQ~\cite{lee2020CelebAMask} to predict the skin tone (light \vs dark) and skin hue (red \vs yellow). We then sample 500,000 images from StyleGAN3 randomly and use the trained classifier to infer skin color labels. Two linear SVMs, one for skin tone and one for skin hue, are finally trained on the most confident positive and negative predictions to produce decision boundaries to be used to compute the latent directions $D$.

For benchmarking, we consider face recognition models of commercial systems such as AWS Rekognition~\cite{aws}, Microsoft Azure~\cite{azure} and Clarifai~\cite{clarifai}. Additionally, we consider the publicly available model trained on the FairFace dataset~\cite{karkkainen2021fairface}.
Note that all models have been released after the seminal Gender Shades paper~\cite{buolamwini2018gender}, which highlighted bias issues in commercial systems, and we use their latest available version for attribute prediction. As a result, biases we observe in this paper might be reduced compared to the Gender Shades paper.

We focus on gender and smile classifications, as both attributes can be predicted with the above systems and come with ground truth labels annotated by a professional company in the CelebA dataset~\cite{liu2015faceattributes}.
We follow previous literature initiated by the Gender Shades paper~\cite{buolamwini2018gender} and follow-up works~\cite{raji2019actionable,jaiswal2022two,raji2020saving}, which perform gender and smile classification to highlight ethical concerns. Such fairness benchmarking promotes model transparency, which in turn creates accountability that could lead to the discontinuation of harmful models~\cite{raji2020saving}.
As such, we stress that we do not condone the gender classification task, as it causes harm to non-binary and transgendered individuals by reducing gender to a binary value~\cite{keyes2018misgendering}, but rather report it to examine additional ethical issues around this commonly used task in the fairness literature.

We consider CelebAMask-HQ~\cite{lee2020CelebAMask} for benchmarking. Modifications of the images involve making the skin tone lighter or darker, as well as making the skin hue more red or yellow. Figure~\ref{fig:manipulations} depicts qualitative samples of the reconstructed image with e4e~\cite{tov2021designing} and StyleGAN3~\cite{karras2021stylegan3}, as well as manipulations in the latent space with InterfaceGAN~\cite{shen2020interpreting}. We observe that when skin tone changes, the hue angle stays the same (\ref{fig:manipulations:tone:light}-\ref{fig:manipulations:tone:dark}); and conversely when manipulating the hue angle (\ref{fig:manipulations:hue:red}-\ref{fig:manipulations:hue:yellow}).
Modifying the image in one direction does not affect the skin color score in the other direction, making them orthogonal.
The same does not hold when selecting the \say{pale} attribute available in CelebA~\cite{liu2015faceattributes} as a comparison.
While the \textit{pale} direction is effective at manipulating the skin tone, it actually also alters the skin hue (\ref{fig:manipulations:pale:plus}-\ref{fig:manipulations:pale:minus}). In other words, both $L^*$ and $h^*$ shift when modifying the \textit{pale} attribute in the images, which makes such metric impractical for measuring the causal effect of skin color as we cannot control its effect.
Thus for benchmarking, we modify all images in the CelebAMask-HQ datasets by increasing and decreasing directions of $L^*$ and $h^*$, for a total of $4{\times}30,000$ images, and report the binary accuracy of evaluated models on gender and smile classifications.

A current limitation of this setup lies in the image reconstruction and manipulation methods, which struggle with spurious correlations. This results in attributes not being preserved, which can be depicted by an altered background or eye gaze (see original \vs reconstructed in Figure~\ref{fig:manipulations:ori}--\ref{fig:manipulations:recon}).
As such, causal experiments in this section are dependent on the quality of synthetic data, akin to related previous works (\eg,~\cite{ramaswamy2021fair,balakrishnan2021towards}).
To avoid capturing performance discrepancies due to imperfect reconstruction, we compare the performance of manipulated images with the \textit{reconstructed} version rather than the \textit{original} one in Table~\ref{tab:attribute:intersection}.

\paragraph{Results.}

Table~\ref{tab:attribute:intersection} presents the attribute prediction performance of several methods on CelebAMask-HQ.
When measuring the performance on gender classification in Table~\ref{tab:attribute:gender}, we observe that manipulating the skin color to have a lighter skin tone decreases the gender classification accuracy while the skin hue does not seem to have a large effect.
The skin tone bias occurs because models are prone to classify people as feminine when the tone is lighter. For example, in the gender predictions of AWS for male samples, the accuracy drops from 94.82\% to 90.66\% while it stays relatively the same for female samples.
Interestingly, manipulating the skin to be redder or darker in male sample results in a increase in accuracy for all models.
When measuring the performance on smile classification in Table~\ref{tab:attribute:smile}, we observe that manipulating the skin color to have a lighter skin tone or a redder skin hue decreases the accuracy in non-smiling individuals as they tend to be predicted as smiling. For example, the accuracy for non-smiling individuals with Azure drops from 80.84\% to 69.76\% when the skin hue becomes redder while it stays the same for smiling individuals.
Conversely, a darker skin tone or a yellower skin hue decreases the accuracy in smiling individuals.
Overall, this benchmark reveals a bias towards a light skin tone when predicting if the individual belongs to the female gender, and a bias towards light or red skin hue when predicting the presence of a smile, which illustrates the importance of a~multidimensional measure of skin color.

\section{Conclusion}\label{sec:conclusion}

Measuring apparent skin color requires a multidimensional score to capture its variation and provide a comprehensive representation of its constitutive complexity.
In this paper, we first focus on the perceptual lightness $L^*$, as a measure of skin tone ranging from light to dark, and the hue angle $h^*$, as a measure of skin hue ranging from red to yellow. Our proposal serves as a simple, yet effective, first step towards a multidimensional skin color score.
Second, we reveal biases related to skin color in image datasets and computer vision models, previously invisible.
Our multidimensional skin color scale offers a more representative assessment to surface socially relevant biases due to skin color effects in computer vision.
While the paper considers face-related tasks, assessing skin color could also be applied to other human-centric tasks (\eg, pose estimation, segmentation, etc).
This would help to (i) enhance the diversity in the data collection process, by encouraging specifications to better represent skin color variability; and (ii) improve the identification of dataset and model biases in fairness benchmarking, by highlighting their limitations and leading to fairness-aware training methods.
Therefore, we recommend the usage of a multidimensional skin color measure as a fairness tool to assess the computer vision pipeline, from data collection to model deployment.

{\small
\bibliographystyle{ieee_fullname}
\bibliography{egbib}
}

\clearpage
\input{supp}

\end{document}

%% file: supp.tex
\begin{strip}
\vspace*{-22pt}
   \begin{center}
      {\Large \bf Beyond Skin Tone: A Multidimensional Measure of Apparent Skin Color \par}
      \vspace*{24pt}
      {
      \large
      \lineskip .5em
      \begin{tabular}[t]{c@{\hskip 3em}c@{\hskip 3em}c}
         \ificcvfinal 
        William Thong & Przemyslaw Joniak & Alice Xiang\\
         Sony AI & The University of Tokyo & Sony AI
         \else Anonymous ICCV submission\\
         \vspace*{1pt}\\
        Paper ID \iccvPaperID \fi
      \end{tabular}
      \par
      }
      \vskip .5em
      \vspace*{12pt}
   \end{center}
\end{strip}

\appendix

\section{Supplementary Material}
\label{sec:supp}\noindent

\setcounter{equation}{0}
\setcounter{figure}{0}
\setcounter{table}{0}
\setcounter{page}{1}
\makeatletter
\renewcommand{\theequation}{S\arabic{equation}}
\renewcommand{\thefigure}{S\arabic{figure}}
\renewcommand{\thetable}{S\arabic{table}}

We provide additional materials to supplement our main paper.
Section~\ref{sec:app:challenges} describes challenges in estimating skin color in in-the-wild images.
Section~\ref{sec:app:extracting} details our method to extract skin color scores in images and performs a robustness analysis with different illuminations.
Section~\ref{sec:app:eth} explores the skin color distribution per ethnicity while Section~\ref{sec:app:data} characterizes skin color bias in datasets beyond binary thresholding.

\subsection{Factors influencing skin color in the wild}\label{sec:app:challenges}

Compared with images acquired in a controlled setting, images in the wild have a wide range of variation. External factors can influence the visual output, which would affect in turn the measure of the \say{true} skin color of the individual. Characterizing the effect of external factors remains an open challenge. As such, similar to previous works, we focus in this paper on the \say{apparent} skin color observed in the image for our fairness evaluation.

\begin{figure}[t]
\centering
\hfill
  \begin{subfigure}{.19\linewidth}
  \centering
  {\scriptsize $L^*{=}37.10$ $h^*{=}-25.38$}
  \includegraphics[width=0.95\linewidth]{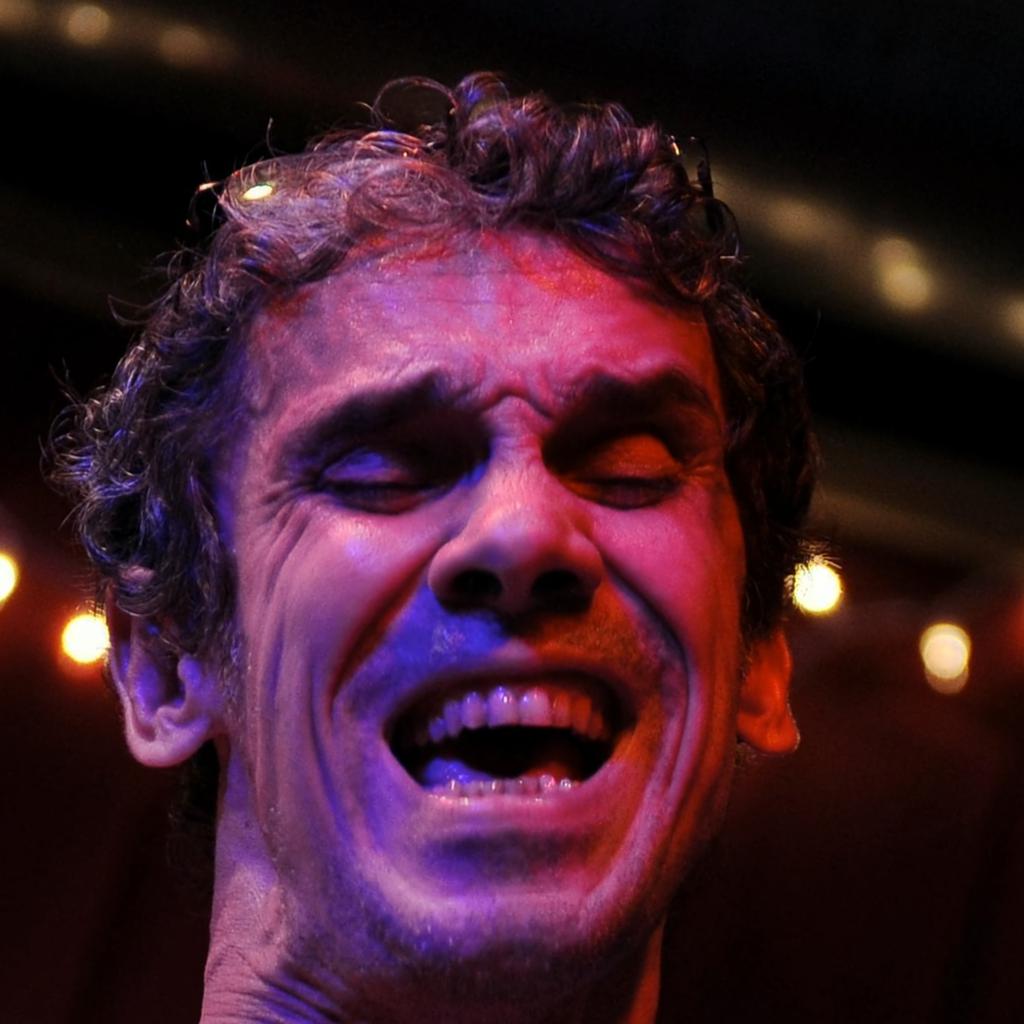}
  \captionsetup{justification=centering}
  \caption{\\Color cast}
  \end{subfigure}\hfill
  \begin{subfigure}{.19\linewidth}
  \centering
  {\scriptsize $L^*{=}19.62$ $h^*{=}45.38$}
  \includegraphics[width=0.95\linewidth]{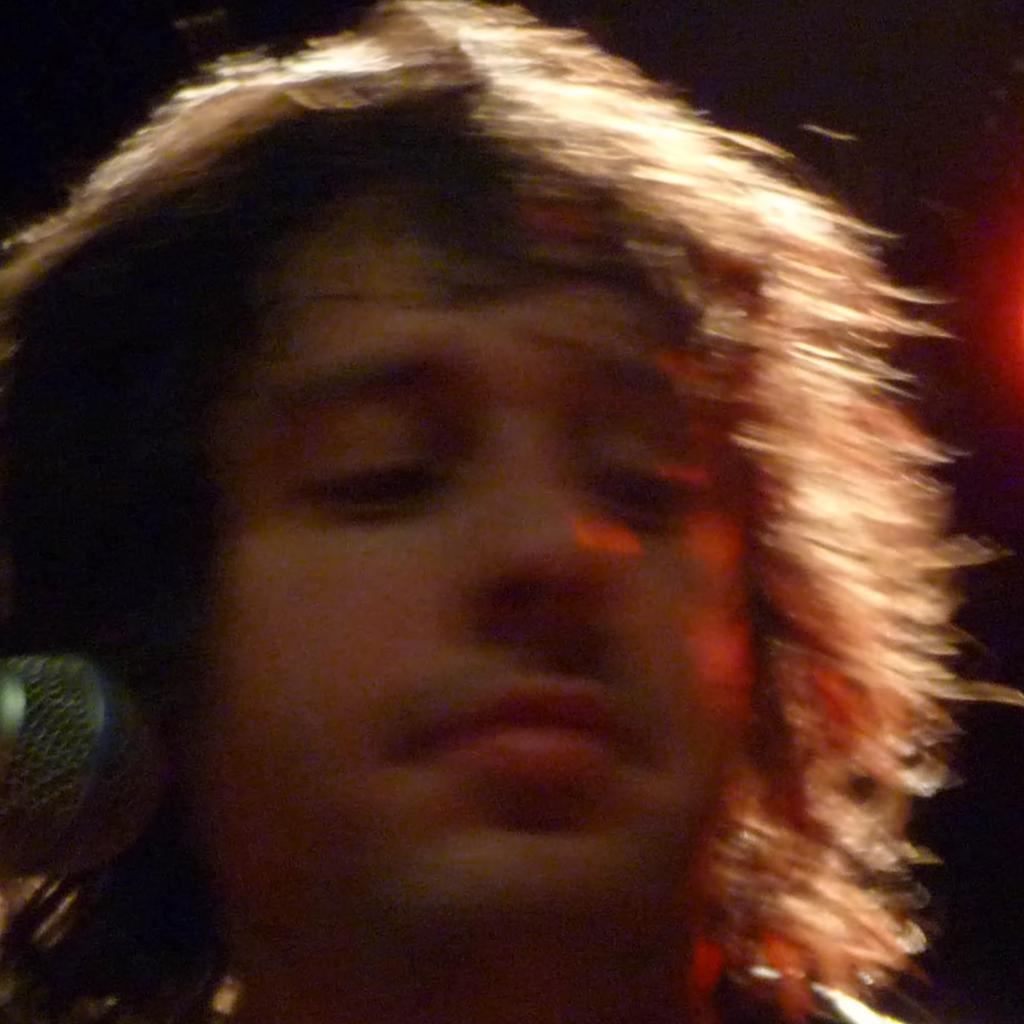}
  \captionsetup{justification=centering}
  \caption{\\Illumination}
  \end{subfigure}\hfill
  \begin{subfigure}{.19\linewidth}
  \centering
  {\scriptsize $L^*{=}18.93$ $h^*{=}-14.70$}
  \includegraphics[width=0.95\linewidth]{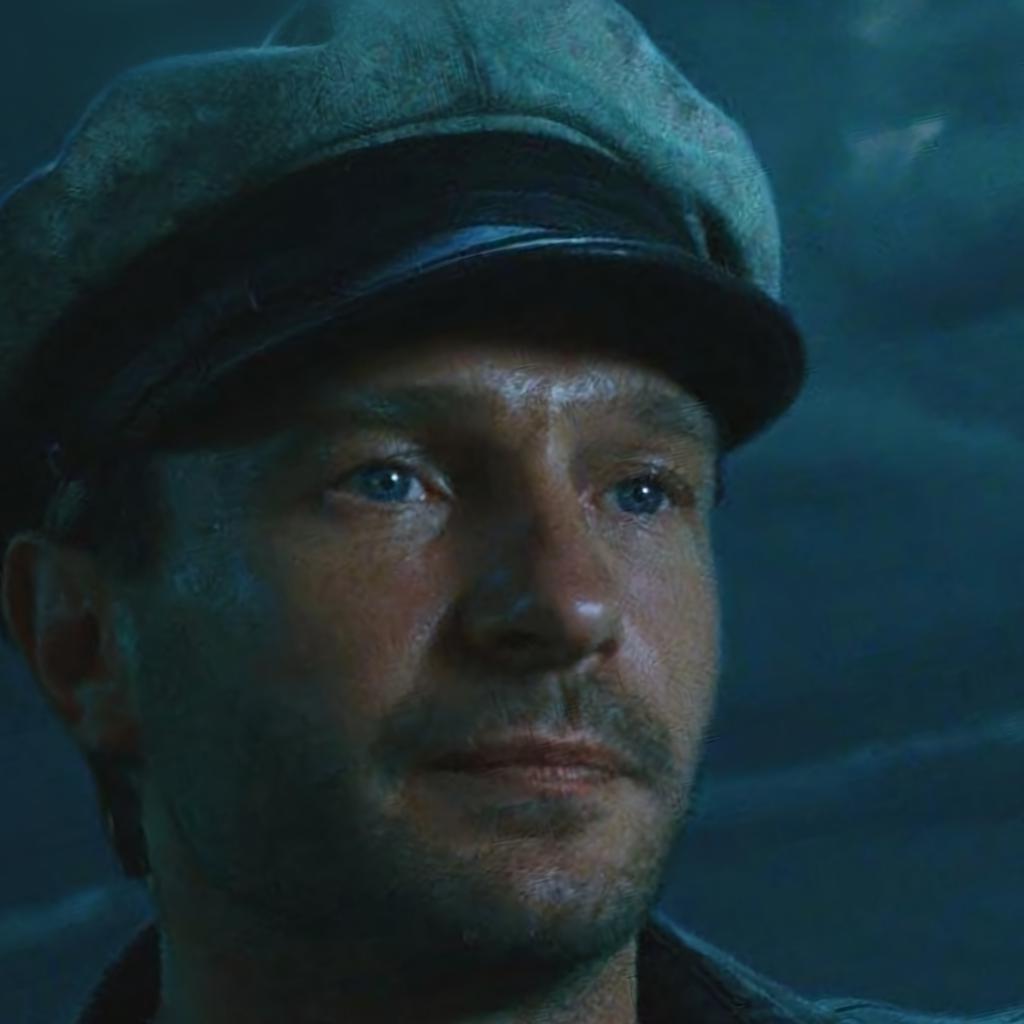}
  \captionsetup{justification=centering}
  \caption{\\Low-light}
  \end{subfigure}\hfill
  \begin{subfigure}{.19\linewidth}
  \centering
  {\scriptsize $L^*{=}46.40$ $h^*{=}-0.70$}
  \includegraphics[width=0.95\linewidth]{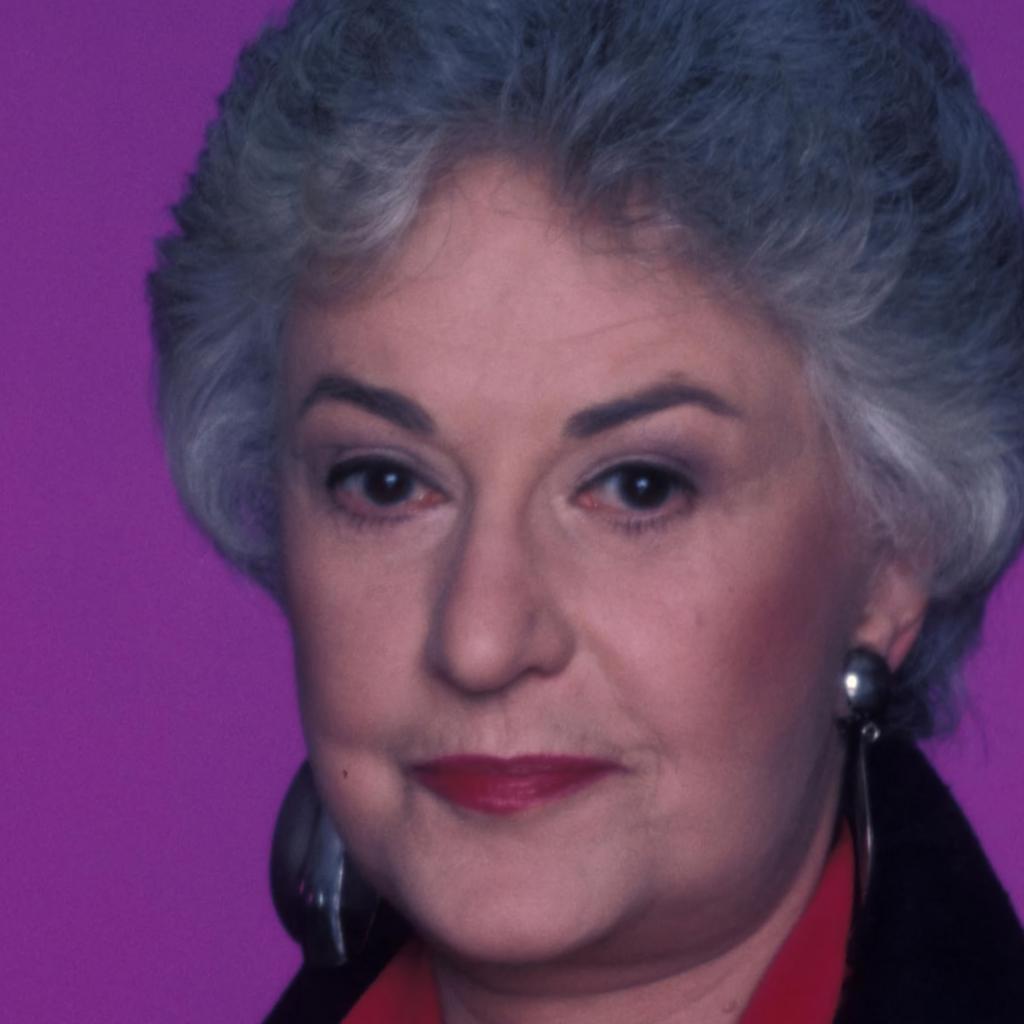}
  \captionsetup{justification=centering}
  \caption{\\Makeup}
  \end{subfigure}\hfill
  \begin{subfigure}{.19\linewidth}
  \centering
  {\scriptsize $L^*{=}81.29$ $h^*{=}-4.22$}
  \includegraphics[width=0.95\linewidth]{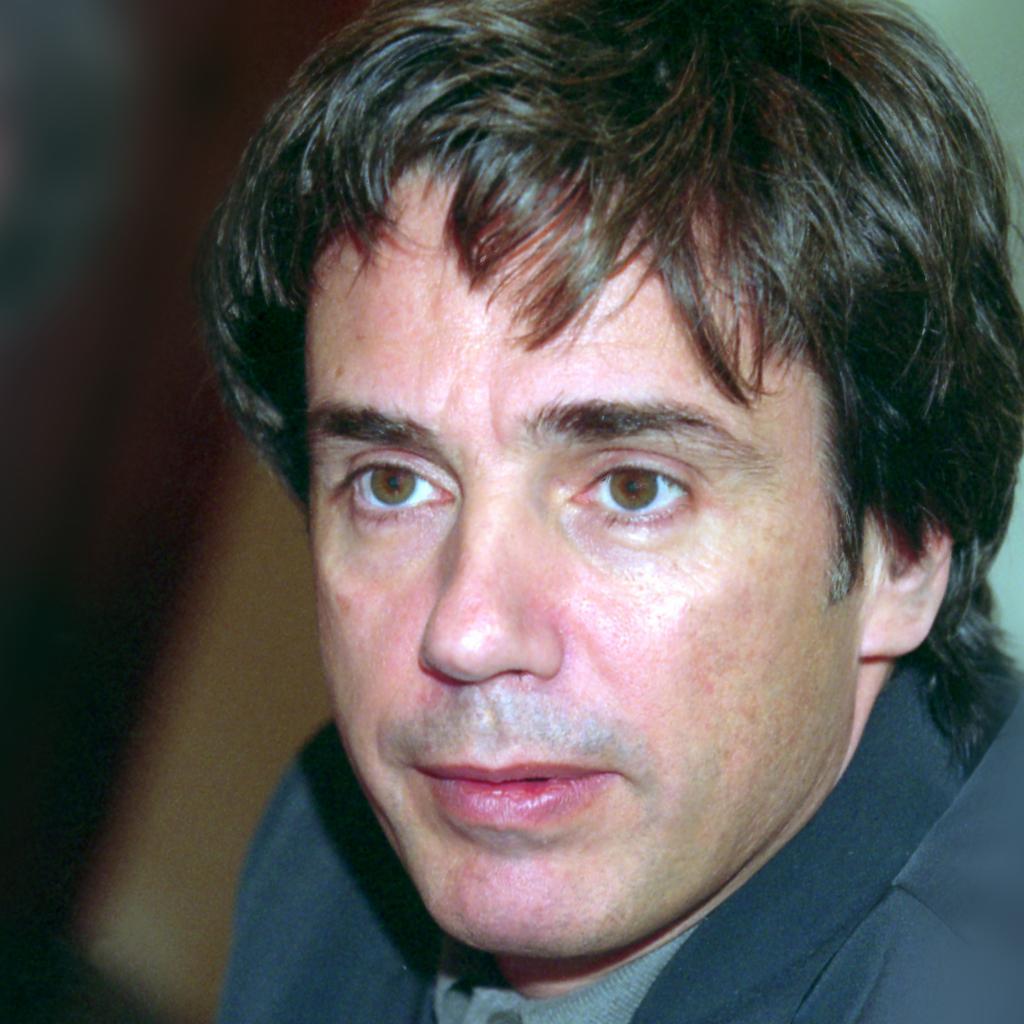}
  \captionsetup{justification=centering}
  \caption{Face \\flushing}
  \end{subfigure}\hfill\null
  \caption{\textbf{Factors influencing skin color measurement} in CelebAMask-HQ. External factors coming from scene affect the skin color (a-c), as well as structural ones from the individuals (d-e).}
  \label{fig:challenges}
\end{figure}

Figure~\ref{fig:challenges} highlights some of the external factors that affect the skin color in the CelebAMask-HQ dataset.
We observe that the \say{apparent} skin color can be affected by external factors in the scene environment or the camera setting (\eg, color cast, intensity and orientation of the illumination, low-light environment, etc), but also by structural factors of the subjects (\eg, having makeup or face flushing). As a result, the skin color in these images can, for example, appear to be much darker or much redder than their \say{true} color.
The effect of these external and structural factors make the measure of skin color in in-the-wild images an open challenging problem.
Still, we consider the \say{apparent} skin color as this what computer vision models are seeing.

\subsection{Extracting skin color from skin pixels}\label{sec:app:extracting}

Skin color scores provide a quantitative measure to characterize the appearance of the skin in an image. Extracting such measures helps to identify potential biases towards skin color subgroups in model performance.
Our objective differs from the cosmetics or dermatology fields, which requires an accurate assessment of constitutive skin color from cutaneous measurements~\cite{ly2020research}.
We focus, instead, on the \say{apparent} skin color in images acquired from any camera, with varying acquisition parameters or lighting conditions.
The main challenge resides in extracting skin color scores from skin pixels in an image. We propose a framework that starts from a facial image $x$ and outputs a~final scalar scoring value $y$ or a set of scalar scoring values $y=\{y^1, \cdots, y^Y\}$.

\begin{figure}
\centering
  \begin{subfigure}{.2\linewidth}
  \centering
  \includegraphics[width=0.94\linewidth]{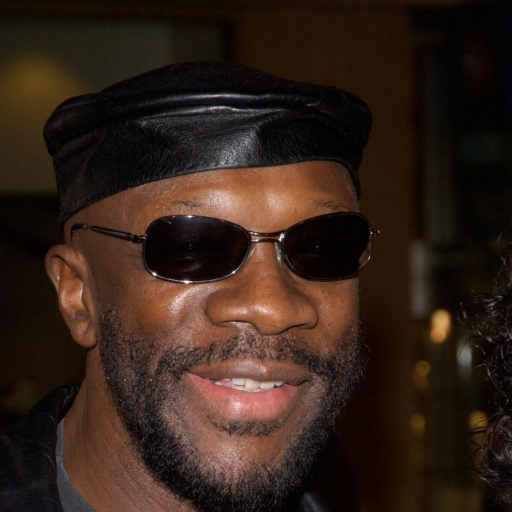}
  \captionsetup{justification=centering}
  \caption{Original\\image}
  \end{subfigure}\hfill
  \begin{subfigure}{.2\linewidth}
  \centering
  \includegraphics[width=0.94\linewidth]{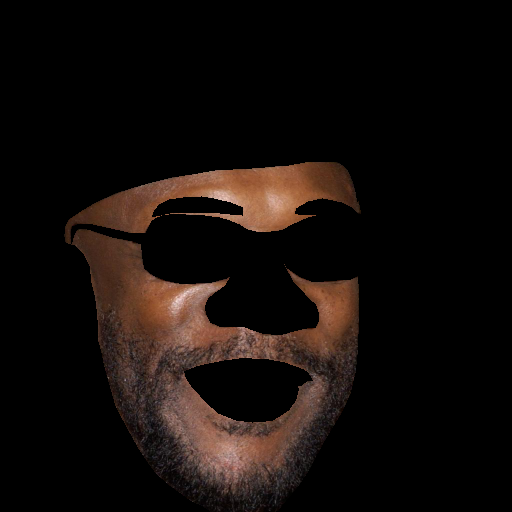}
  \captionsetup{justification=centering}
  \caption{Skin\\pixels}
  \end{subfigure}\hfill
  \begin{subfigure}{.4\linewidth}
  \centering
  \includegraphics[width=0.47\linewidth]{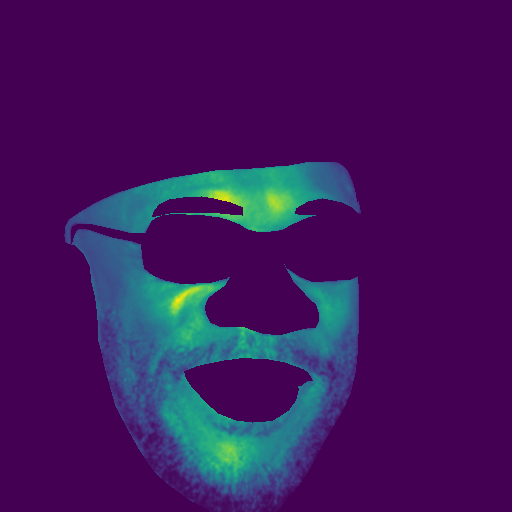}
  \includegraphics[width=0.47\linewidth]{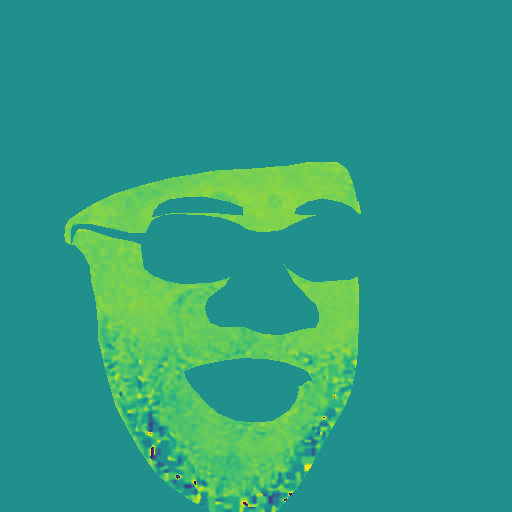}
  \caption{$L^*$ and $h^*$\\~}
  \end{subfigure}\hfill
  \begin{subfigure}{.2\linewidth}
  \centering
  \includegraphics[width=0.94\linewidth]{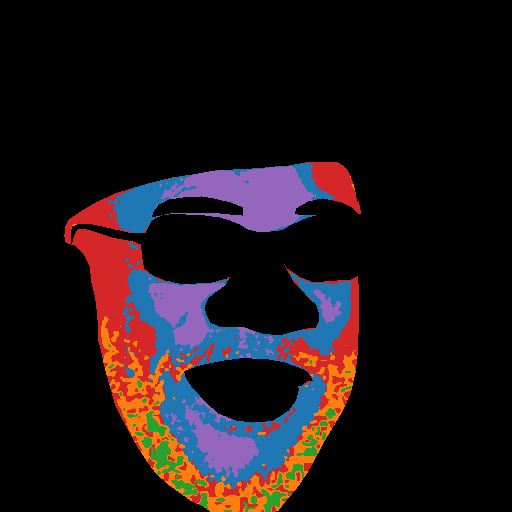}
  \caption{Clusters\\~}
  \end{subfigure}
  \caption{\textbf{Extracting skin color scores} method overview. Given an input image, we isolate the skin pixels and compute a per-pixel skin color score measurement. Pixels are then clustered together and we perform a weighted average of skin color scores of each cluster to get the f}
  \label{fig:overview}
\end{figure}

\begin{figure*}[t]
\centering
  \begin{subfigure}{.12\textwidth}
  \centering
  {\scriptsize $L^*{=}66.63$ $h^*{=}59.12$}
  \includegraphics[width=0.95\linewidth]{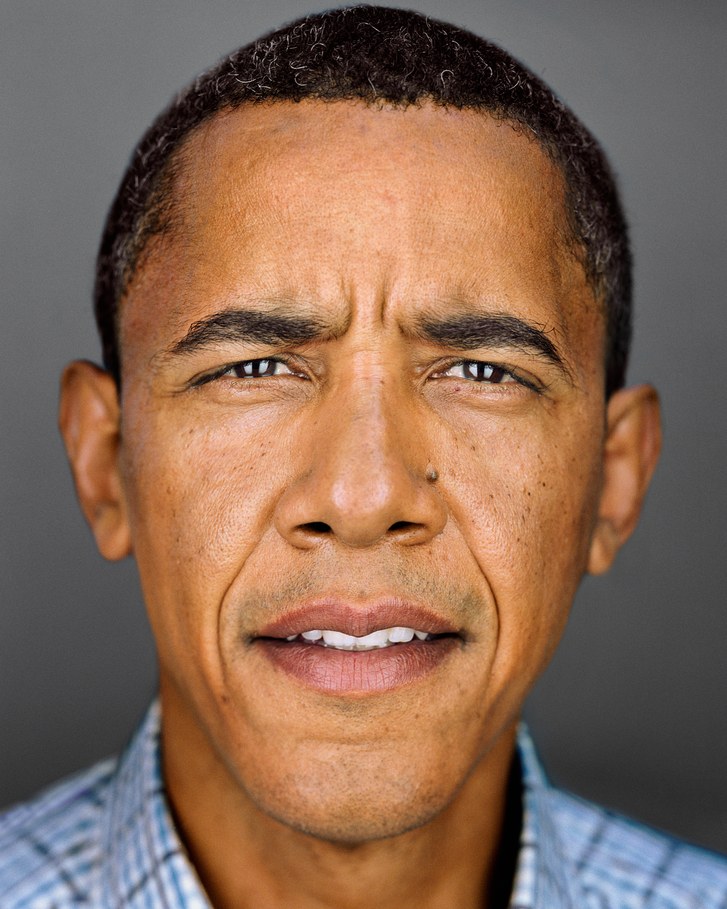}
  \includegraphics[width=0.95\linewidth]{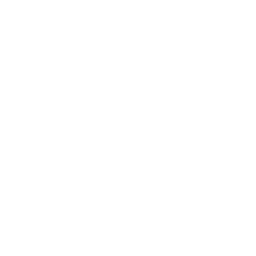}
  \end{subfigure}\hfill
  \begin{subfigure}{.12\textwidth}
  \centering
  {\scriptsize $L^*{=}69.24$ $h^*{=}59.08$}
  \includegraphics[width=0.95\linewidth]{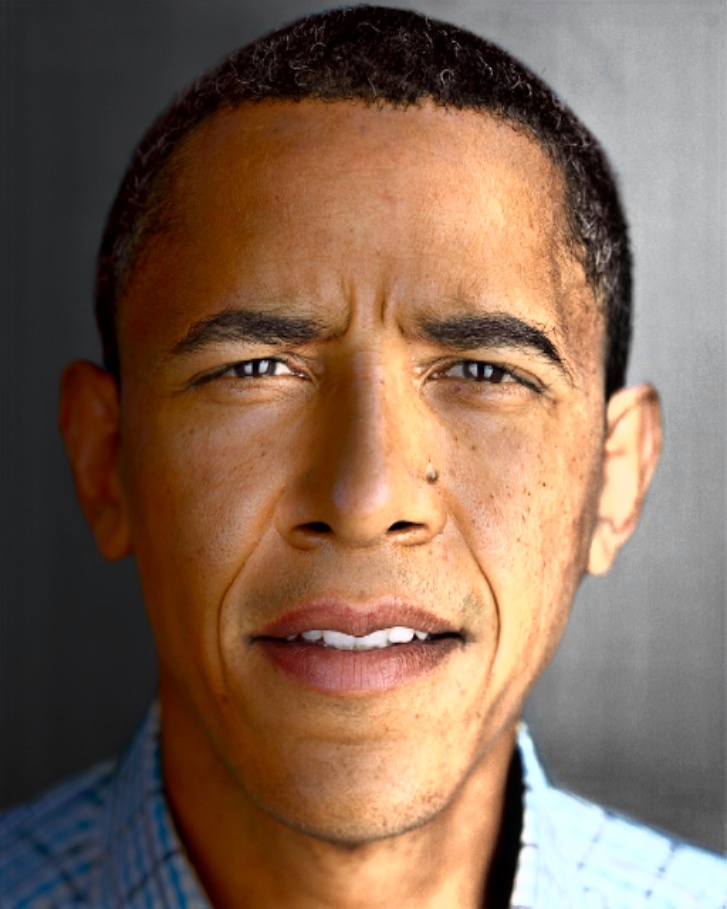}
  \includegraphics[width=0.95\linewidth]{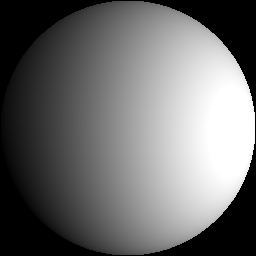}
  \end{subfigure}\hfill
  \begin{subfigure}{.12\textwidth}
  \centering
  {\scriptsize $L^*{=}69.66$ $h^*{=}58.80$}
  \includegraphics[width=0.95\linewidth]{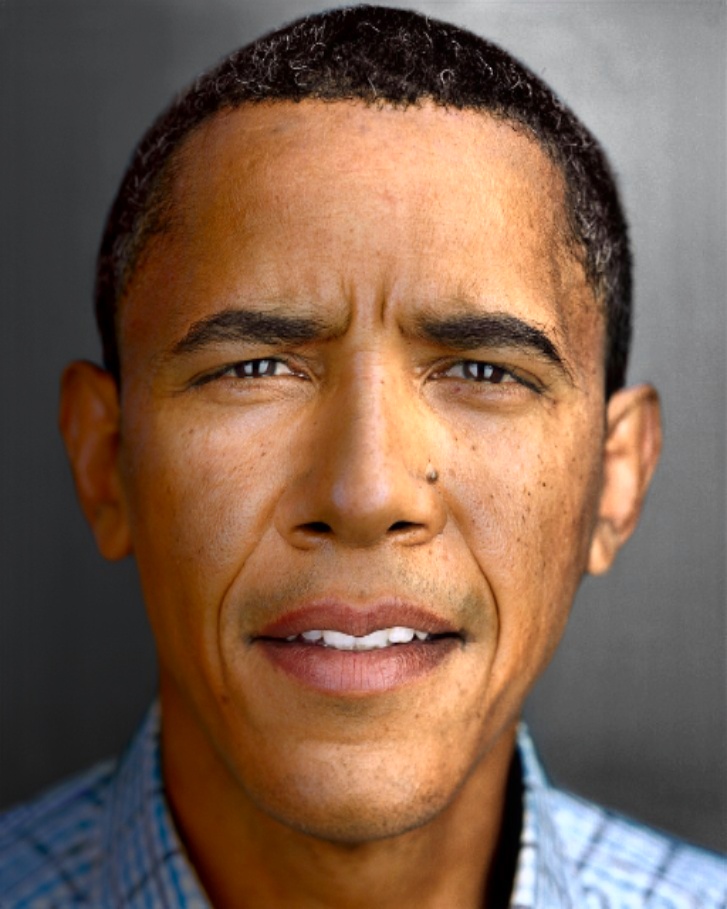}
  \includegraphics[width=0.95\linewidth]{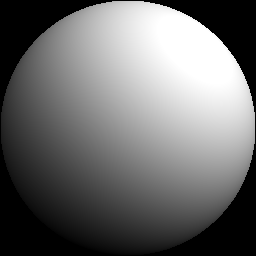}
  \end{subfigure}\hfill
  \begin{subfigure}{.12\textwidth}
  \centering
  {\scriptsize $L^*{=}63.49$ $h^*{=}58.14$}
  \includegraphics[width=0.95\linewidth]{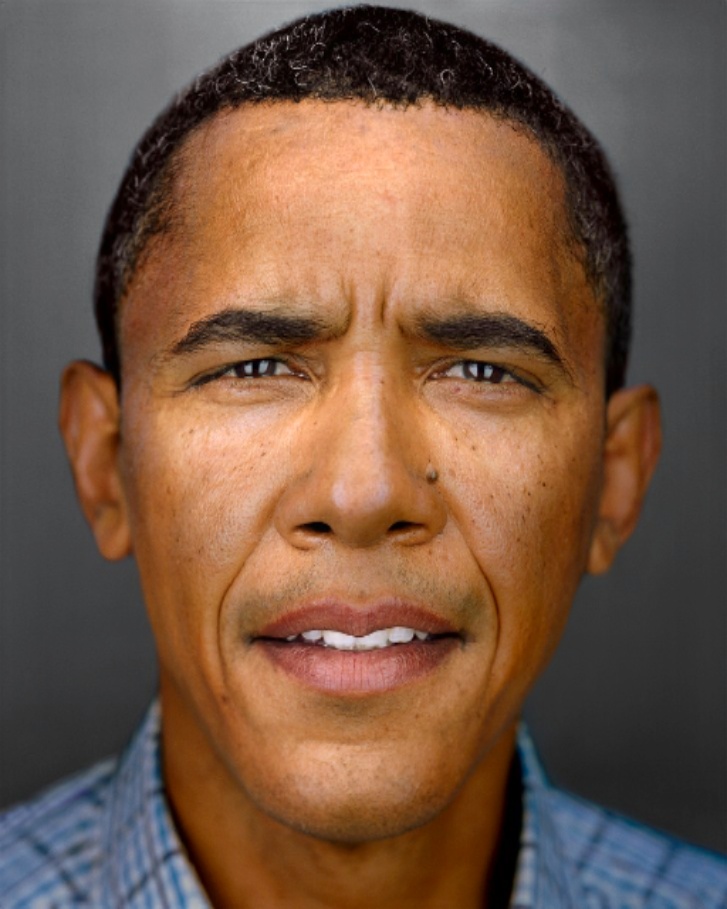}
  \includegraphics[width=0.95\linewidth]{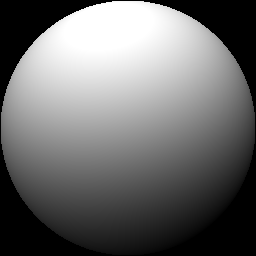}
  \end{subfigure}\hfill
  \begin{subfigure}{.12\textwidth}
  \centering
  {\scriptsize $L^*{=}70.48$ $h^*{=}58.37$}
  \includegraphics[width=0.95\linewidth]{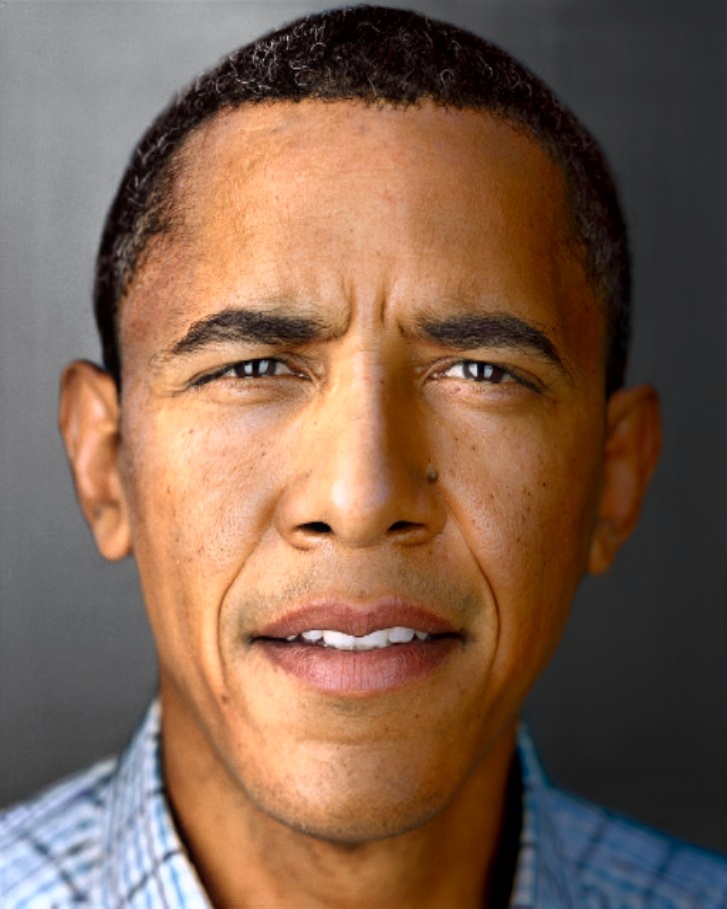}
  \includegraphics[width=0.95\linewidth]{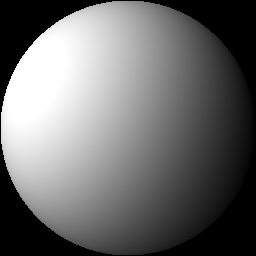}
  \end{subfigure}\hfill
  \begin{subfigure}{.12\textwidth}
  \centering
  {\scriptsize $L^*{=}71.34$ $h^*{=}57.08$}
  \includegraphics[width=0.95\linewidth]{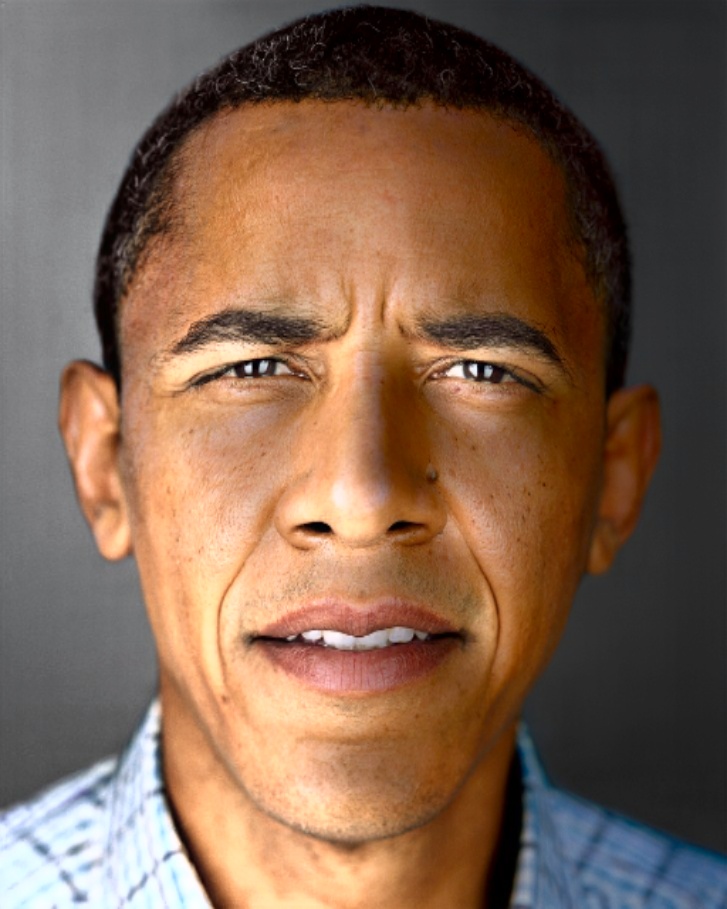}
  \includegraphics[width=0.95\linewidth]{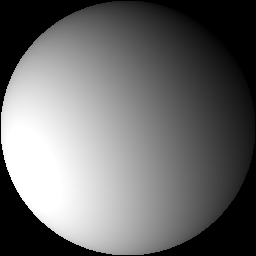}
  \end{subfigure}\hfill
  \begin{subfigure}{.12\textwidth}
  \centering
  {\scriptsize $L^*{=}59.41$ $h^*{=}58.08$}
  \includegraphics[width=0.95\linewidth]{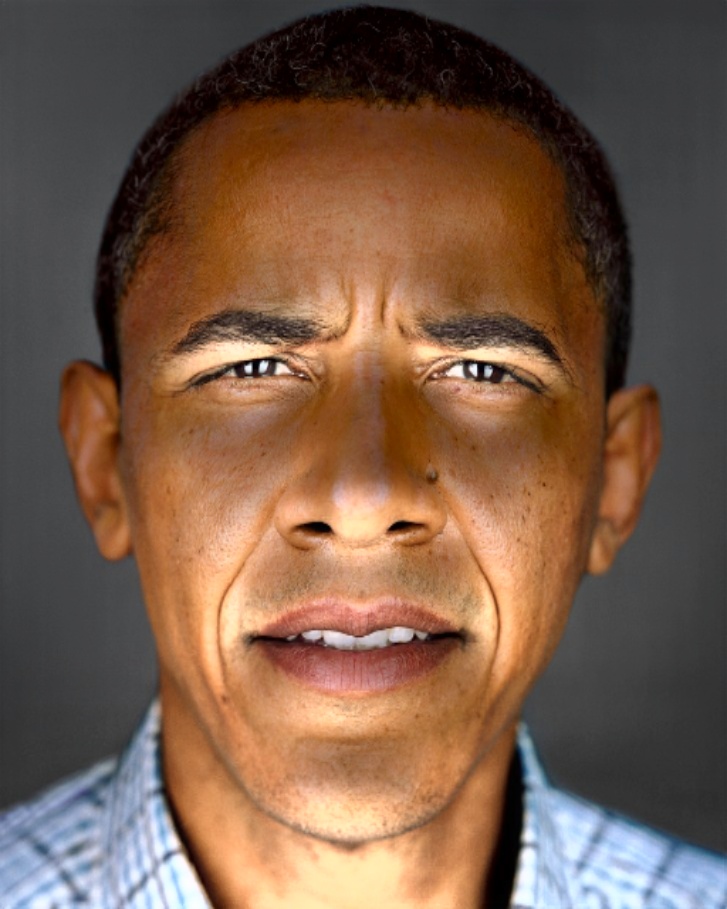}
  \includegraphics[width=0.95\linewidth]{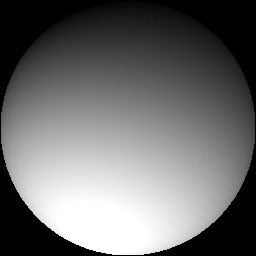}
  \end{subfigure}\hfill
  \begin{subfigure}{.12\textwidth}
  \centering
  {\scriptsize $L^*{=}64.58$ $h^*{=}57.40$}
  \includegraphics[width=0.95\linewidth]{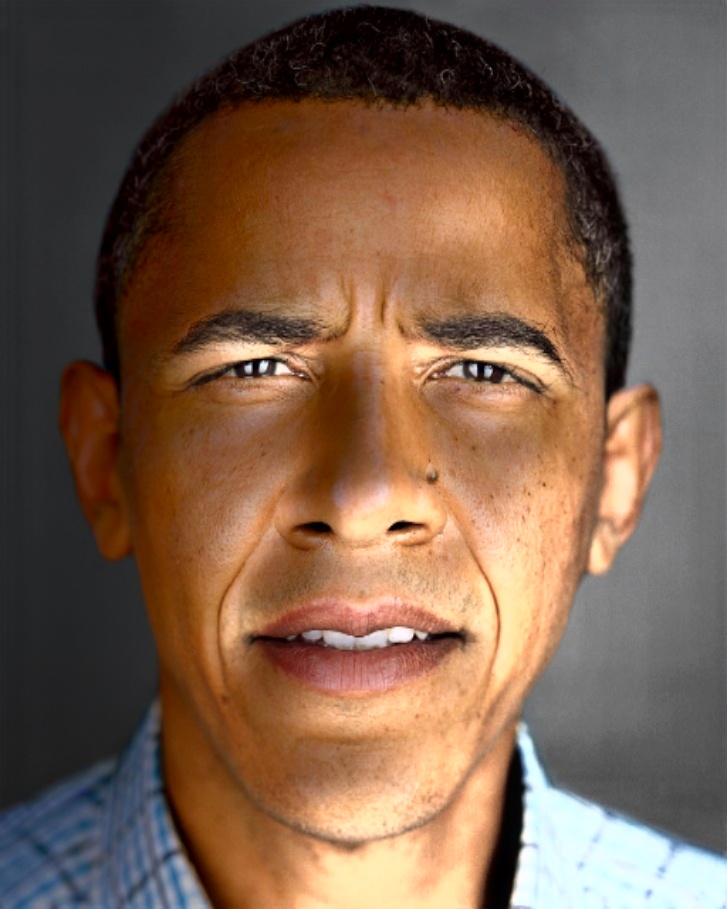}
  \includegraphics[width=0.95\linewidth]{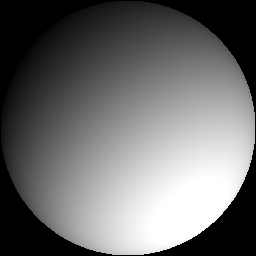}
  \end{subfigure}
  \caption{
  \textbf{Robustness of skin color scores} to face relighting.
  When changing the direction and intensity of the illumination with~\cite{DPR}, the perceptual lightness $L^*$ varies by up to 6 points w.r.t. the original image.
  while the hue angle $h^*$ remains stable.
  Given that we strive to measure the ``apparent'' skin color, as seen by a computer vision model, rather than the ``true'' skin color of individuals, changes in perceptual lightness are expected.
  }
  \label{fig:robustness}
\end{figure*}

\paragraph{Method.}
To extract skin color scores from a facial image, we are inspired by the algorithm initially proposed by Merler~\etal~\cite{merler2019diversity} for the Diversity in Faces dataset (see Section 4.6 in their paper), and generalize their method to handle any scalar scoring value, any face pose and facial variations.
Figure~\ref{fig:overview} presents the steps to extract a skin color score in human-centric images:

\begin{enumerate}[nolistsep,label=(\alph*)]
\item We are given an input image of a subject.
\item As we are interested in skin color, we start by segmenting the skin pixels. Segmentation can be done manually by an annotator or predicted by a skin segmentation model.
\item Once skin pixels have been identified, they are converted from the standard RGB space to the target space of the desired scoring values. We convert to the CIELAB space to extract the $L^*$ component, and further use the $a^*$ and $b^*$ components to compute the hue angle $h^*$. This results in a point measurement of $L^*$ and $h^*$ for every skin pixel in the image.
\item We then apply a clustering algorithm, such as K-Means~\cite{lloyd1982least}, to group the skin pixels. For every cluster, we compute a histogram of distribution and set the number of bins with the Sturges formula~\cite{sturges1926choice}. The mode of the histogram is then used to assign a scalar value for each considered skin color score. In our case, this results in $L^*$ and $h^*$ scalar values for every cluster.
\end{enumerate}

To obtain the final scalar scores for the image, we average the scalar values of every group normalized by their pixel size.
However, as some parts of the face can skew the results towards darker values (\eg, facial hair or shaded regions), we prefer to exclude some groups which yield a very low $L^*$. We cluster the face skin into five groups and keep the top-3 groups with the highest $L^*$ to compute the final $L^*$ and $h^*$ scalar scores for the image.
Such approach is inspired by how human artists would perform value grouping in five different groups when simplifying an image~\cite{publishing2022artists}.

Similar to Merler~\etal~\cite{merler2019diversity}, we start from an image of a subject and require a segmentation of facial parts to obtain a skin mask.
Such segmentation can be obtained via manual labeling or automatically via a model for skin segmentation.
The difference lies mainly in steps (c) and (d).
In step (c), we consider any skin color score and not only the individual typology angle.
In step (d), we remove the need for facial landmarks by relying on skin pixel clustering. This better deals with atypical facial poses, as clustering can handle faces that are unaligned or from the side (\ie, without visible landmarks). Moreover, clustering can identify shaded areas of the face or facial hair, which we remove to avoid contaminating the final skin color scores.

\begin{figure}[t]
\centering
\hfill
  \begin{subfigure}{.19\linewidth}
  \centering
  {\scriptsize $L^*{=}72.15$ $h^*{=}55.96$}
  \includegraphics[width=0.95\linewidth]{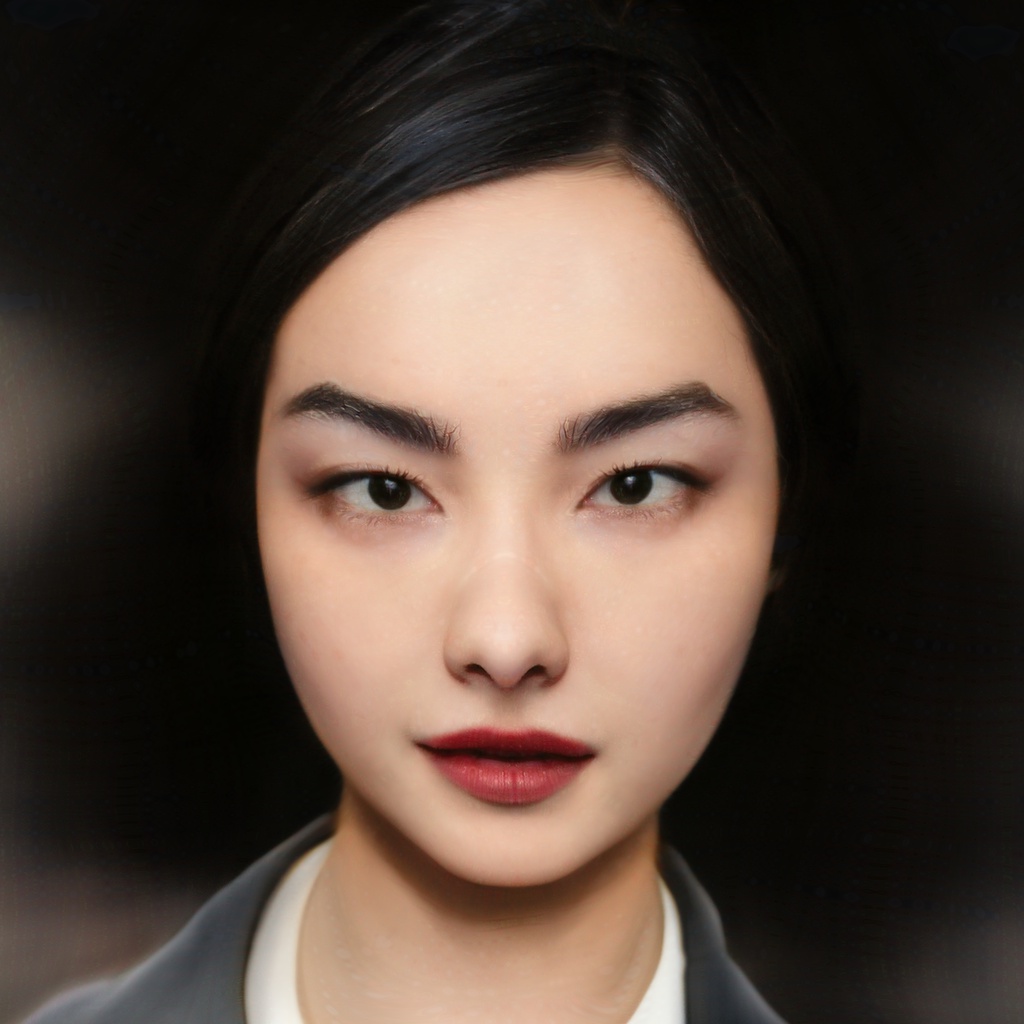}
  \captionsetup{justification=centering}
  \caption{Reconstructed}
  \end{subfigure}\hfill
  \begin{subfigure}{.19\linewidth}
  \centering
  {\scriptsize $L^*{=}77.43$ $h^*{=}55.88$}
  \includegraphics[width=0.95\linewidth]{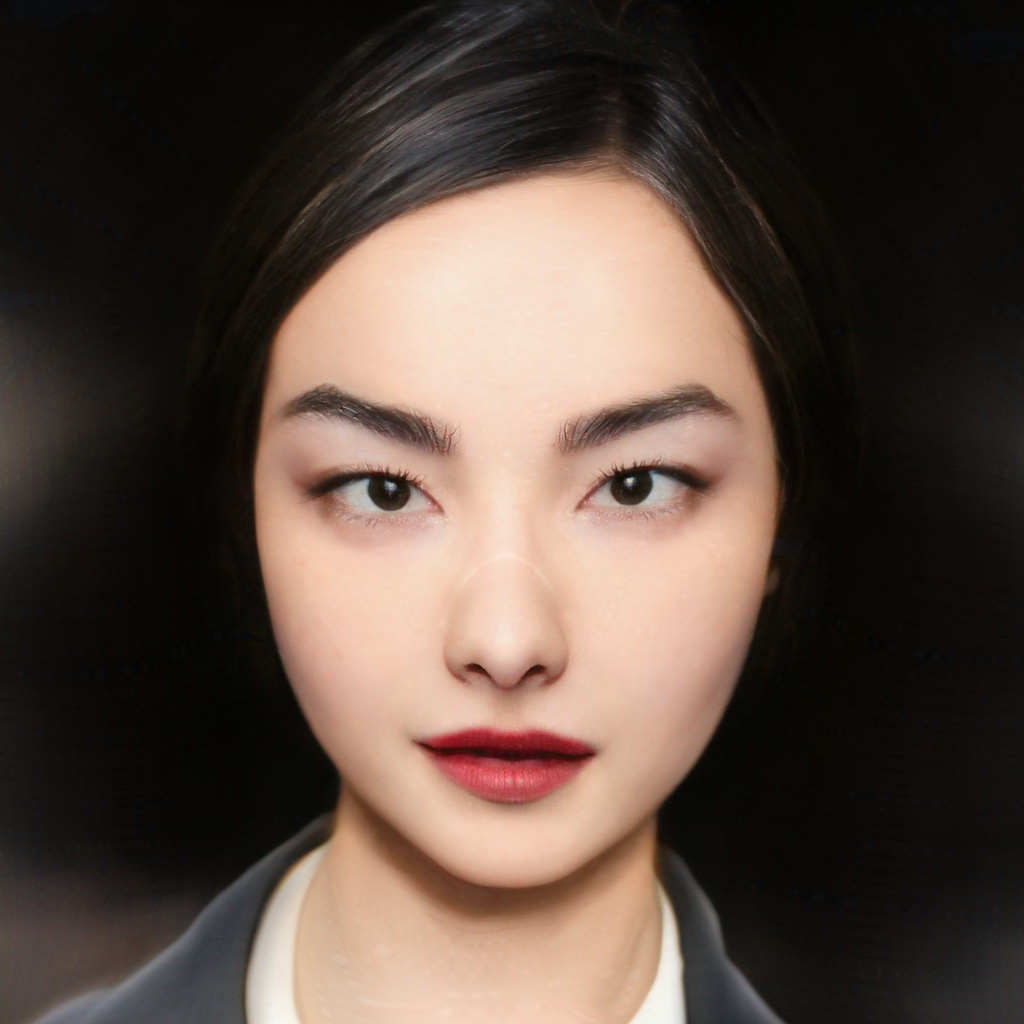}
  \captionsetup{justification=centering}
  \caption{\\+light}
  \end{subfigure}\hfill
  \begin{subfigure}{.19\linewidth}
  \centering
  {\scriptsize $L^*{=}62.40$ $h^*{=}56.84$}
  \includegraphics[width=0.95\linewidth]{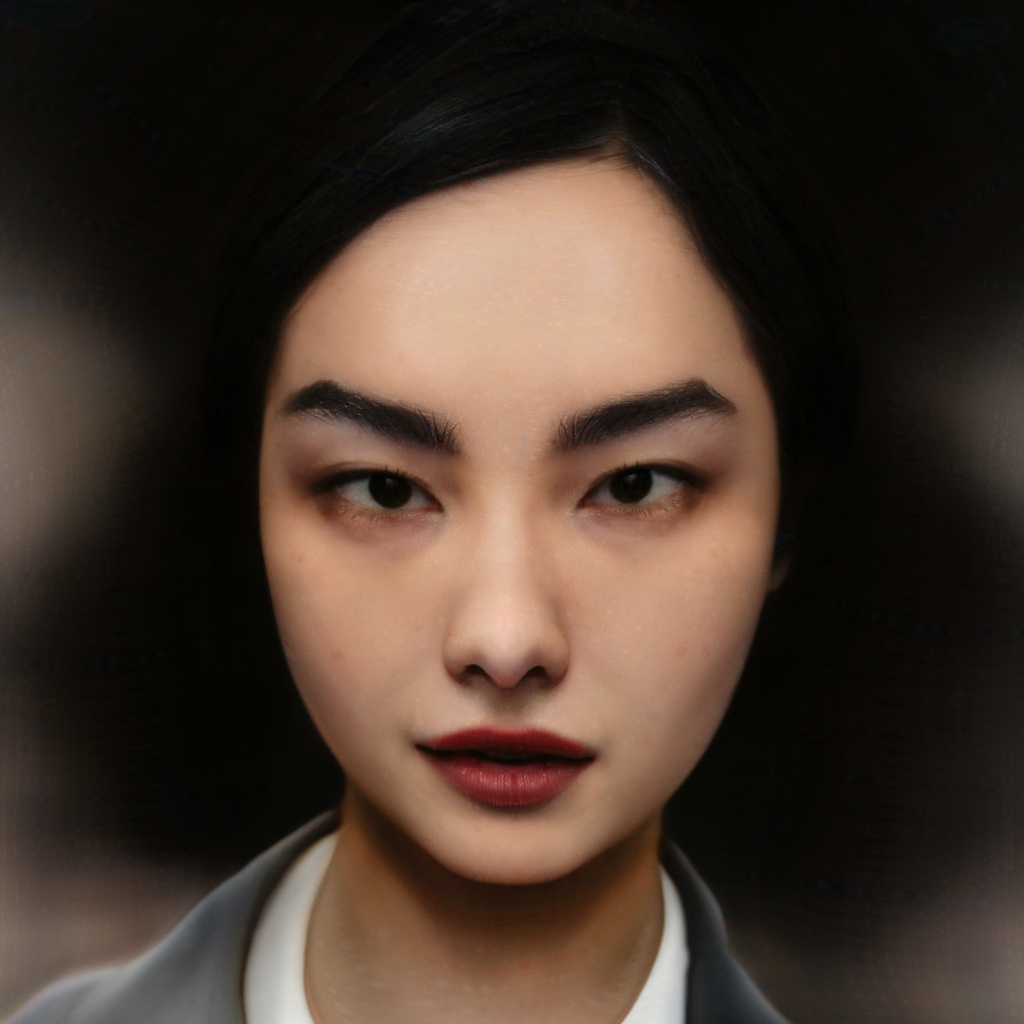}
  \captionsetup{justification=centering}
  \caption{\\+dark}
  \end{subfigure}\hfill
  \begin{subfigure}{.19\linewidth}
  \centering
  {\scriptsize $L^*{=}91.74$ $h^*{=}14.05$}
  \includegraphics[width=0.95\linewidth]{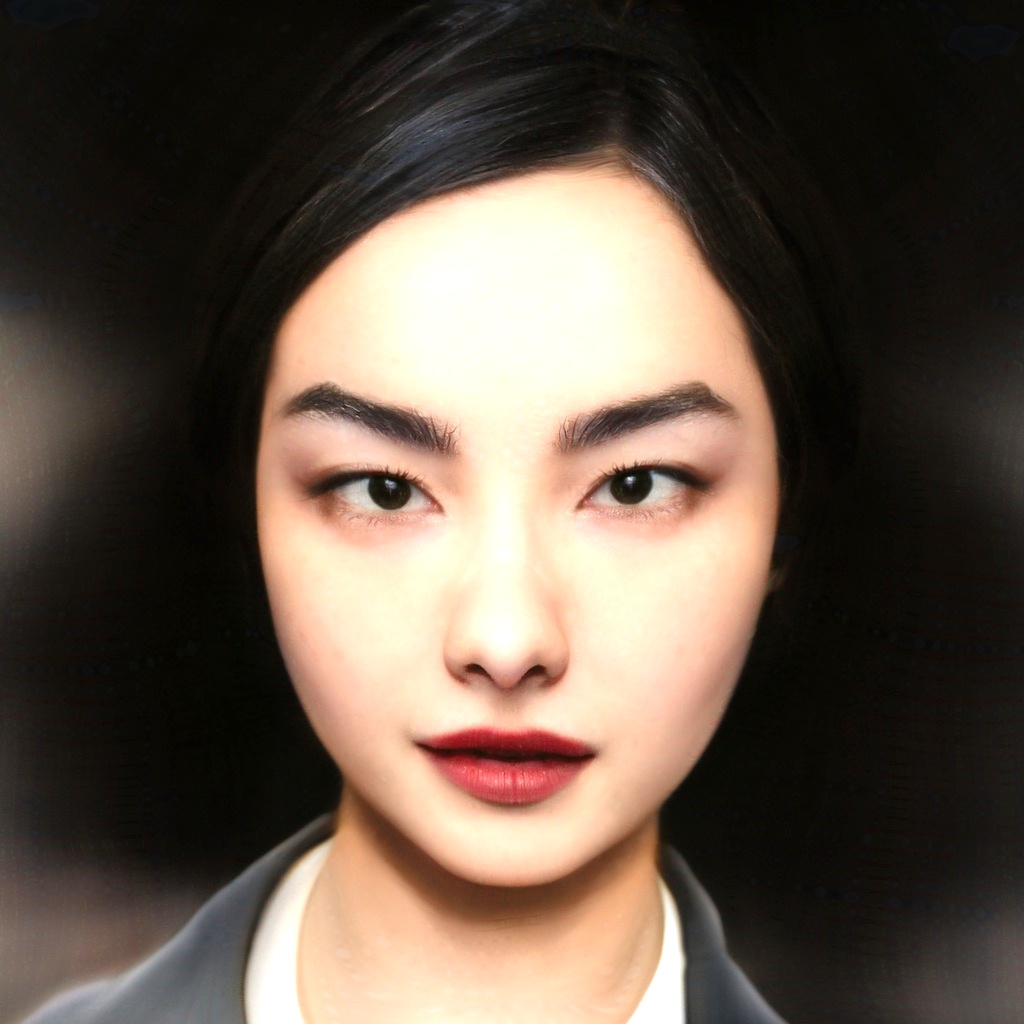}
  \captionsetup{justification=centering}
  \caption{\\+contrast}
  \end{subfigure}\hfill
  \begin{subfigure}{.19\linewidth}
  \centering
  {\scriptsize $L^*{=}51.10$ $h^*{=}56.56$}
  \includegraphics[width=0.95\linewidth]{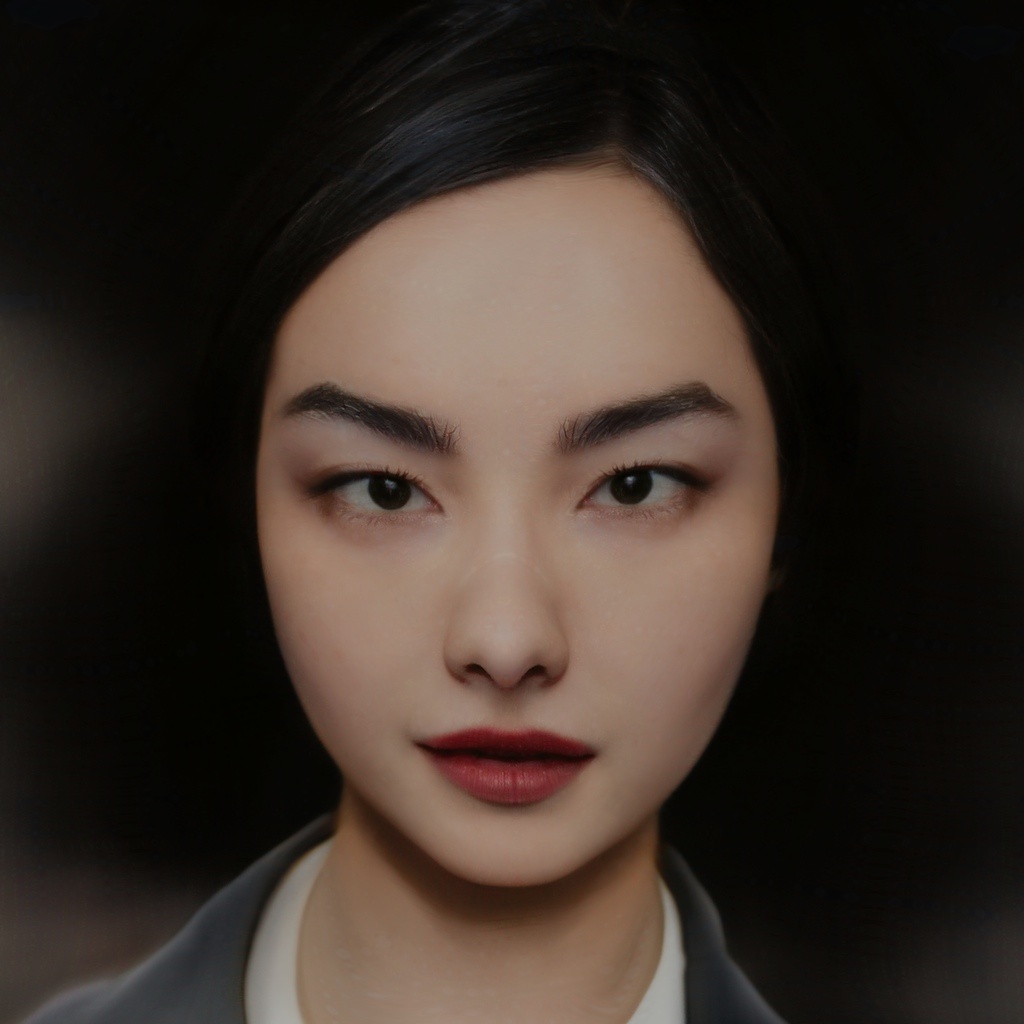}
  \captionsetup{justification=centering}
  \caption{\\-contrast}
  \end{subfigure}\hfill\null
  \caption{\textbf{Skin color \vs contrast manipulation} on CelebAMask-HQ. When modifying the image contrast (d-e), faces appear less realistic and the hue angle can change.}
  \label{fig:contrast}
\end{figure}

\paragraph{Robustness.}
To gain insights about the robustness of the proposed method, we extract skin color scores for a series of images in which the illumination changes in terms of direction and intensity for a given individual.
To achieve this, we rely on samples coming from the work of Zhou~\etal~\cite{DPR} where a deep neural network produces different face relighting images depending on the target lighting.
Figure~\ref{fig:robustness} reports for the perceptual lightness $L^*$ and hue angle $h^*$ for every sample. Face relighting affects $L^*$, which can differ from up to 6 points with respect to the original image. Interestingly, $h^*$ remains stable and robust to face relighting, which confirms that it provides complementary and orthogonal information about skin color than the skin tone.
Differences in $L^*$ are expected as we measure the \say{apparent} skin color in images, which is affected by the illumination, rather than the \say{true} skin color of individuals.

\paragraph{Contrast.}
To gain additional insights on the relevance of the proposed method, we measure the root mean square contrast on the whole image using the perceptual lightness channel.
To achieve this, we consider the manipulated images in Section~\ref{sec:causal}. Reconstructed images have an average contrast of 0.644 while images manipulated to have a darker skin tone are at 0.630 and the ones to have a lighter skin tone at 0.652. Manipulating images has an effect on the overall contrast.
Furthermore, we compare our skin color manipulation with a contrast manipulation by scaling the pixel values.
Figure~\ref{fig:contrast} shows that image contrast should not be conflated with the skin tone. Increasing or decreasing the image contrast does not result in the same visual modifications, as manipulated faces appear less realistic and the hue angle is not preserved.

\subsection{Skin color distribution}\label{sec:app:eth}

\begin{figure}[t]
\centering
  \hfill
  \begin{subfigure}{.5\linewidth}
  \centering
  \includegraphics[width=\linewidth]{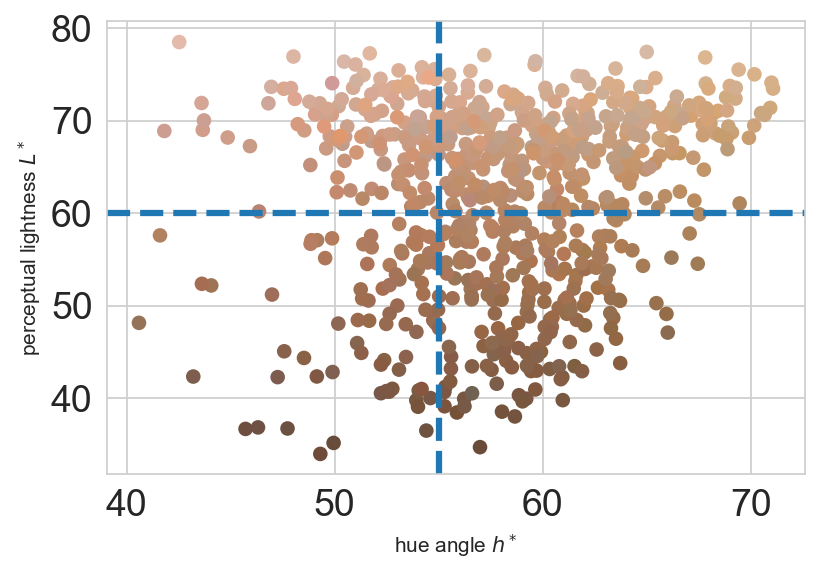}
  \caption{Skin color representation}\label{fig:cfd:color}
  \end{subfigure}\hfill
  \begin{subfigure}{.5\linewidth}
  \centering
  \includegraphics[width=\linewidth]{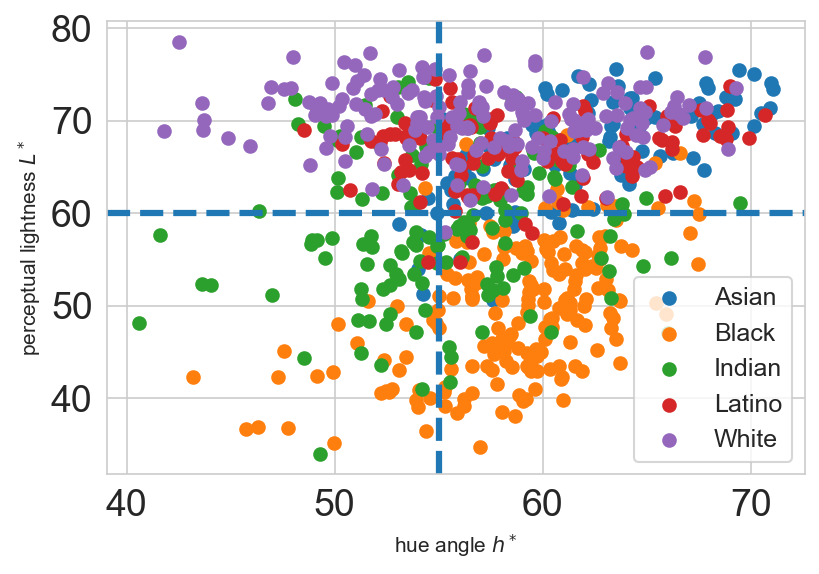}
  \caption{Ethnicity representation}\label{fig:cfd:eth}
  \end{subfigure}\hfill\null
  \caption{\textbf{Skin color distribution} on CFD dataset w.r.t. perceptual lightness and hue angle.
  Every dot in the scatter plot corresponds to an image sample in the dataset.
  The skin tone threshold is at value 60 (light \vs dark), and the hue threshold at value 55\degree~(yellow \vs red).
  }
  \label{fig:cfd}
\end{figure}

Figure~\ref{fig:cfd} depicts the distribution of the CFD dataset in terms of perceptual lightness and hue angle.
Contrary to the distribution of the CelebAMask-HQ or FFHQ-Ageing datasets, the  CFD dataset depicts less variation (Figure~\ref{fig:cfd:color} \vs. Figure~\ref{fig:illustration}). This is explained by the fact that images in CFD have been captured in a controlled setting, enabling fair comparisons among images.

Another interesting aspect of CFD is the available self-reported ethnic labels. When breaking down the distribution with the ethnic labels (Figure~\ref{fig:cfd:eth}), we observe some trends in the skin color for the subjects included in the dataset.
When comparing White and Black skin color scores, the skin tone --expressed through the perceptual lightness $L^*$-- is sufficient to distinguish both groups.
This explains why the fairness literature (\eg,~\cite{buolamwini2018gender,raji2019actionable}) has focused on skin tone to characterize skin color in images, as it serves as a proxy for White and Black skins.
Nevertheless, when subjects from other ethnicities are included, boundaries become fuzzy and the skin tone is no longer enough to capture the variability in skin colors. Towards this goal, it is relevant to consider the hue angle $h^*$ to assess the skin hue and reveal other skin color differences.
For example,
Indian and Black subjects have a darker skins than the other considered ethnicities, with Black subjects having a darker and more yellow skin color than Indian subjects.
Another difference lies in Asian subjects, which are mostly in the yellow side of skin hue, as opposed to Latino and White subjects, which have the same skin tone but appear to be more spread in terms of skin hue.

In the scenario where a data collection process would only measure the skin tone for diversity, collecting data from White and Black subjects would be enough to cover the whole spectrum. This is an issue because this would ignore other types of skin color coming from Indian or Asian skin colors for example.
Prior work has notably shown that computer vision systems produced in the West often exhibit lower performance for Asian individuals~\cite{phillips2011other}.
Including the hue angle, as proposed in the paper, would avoid such an effect where subgroups could be conflated with others despite different skin color characteristics because it gets collapsed into a single \say{light \vs dark} dimension.
Overall, adding the hue offers a complementary perspective to assess skin color beyond the tone and reveal previously invisible biases.

\begin{figure}[t]
\centering
  \begin{subfigure}{\linewidth}
  \centering
  \hfill
  \includegraphics[width=.5\linewidth]{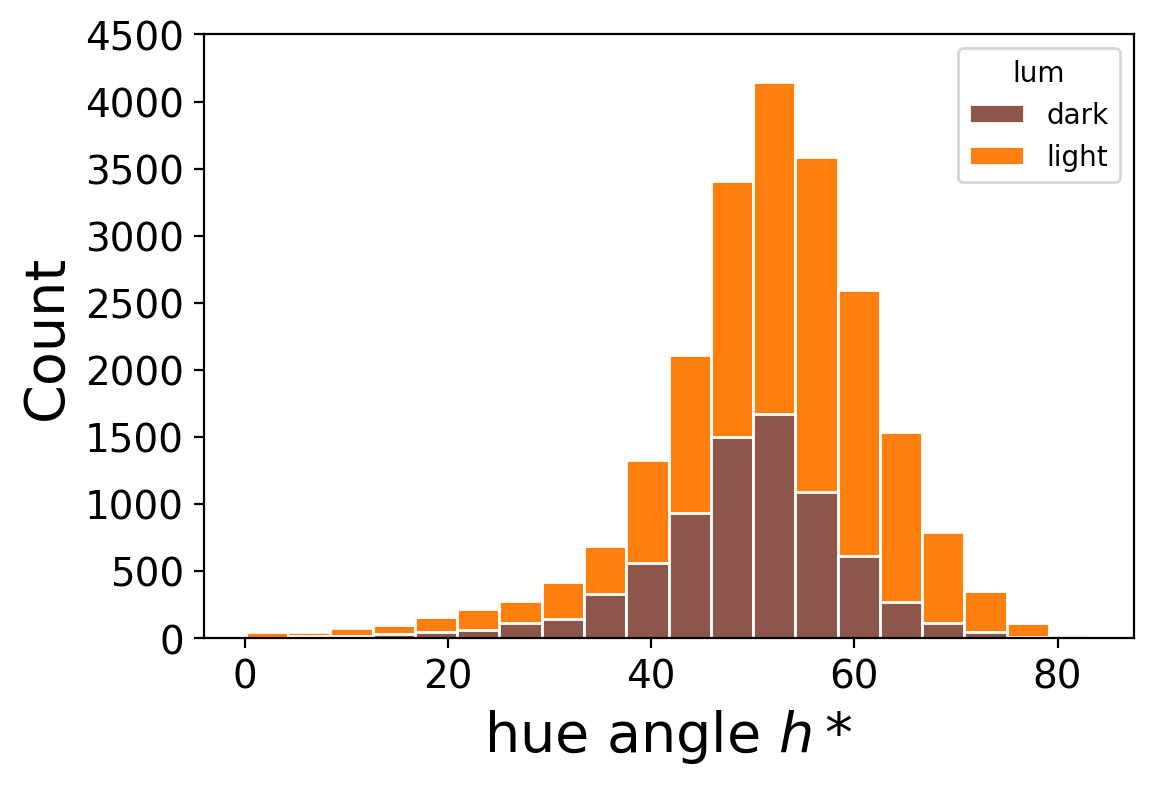}\hfill
  \includegraphics[width=.5\linewidth]{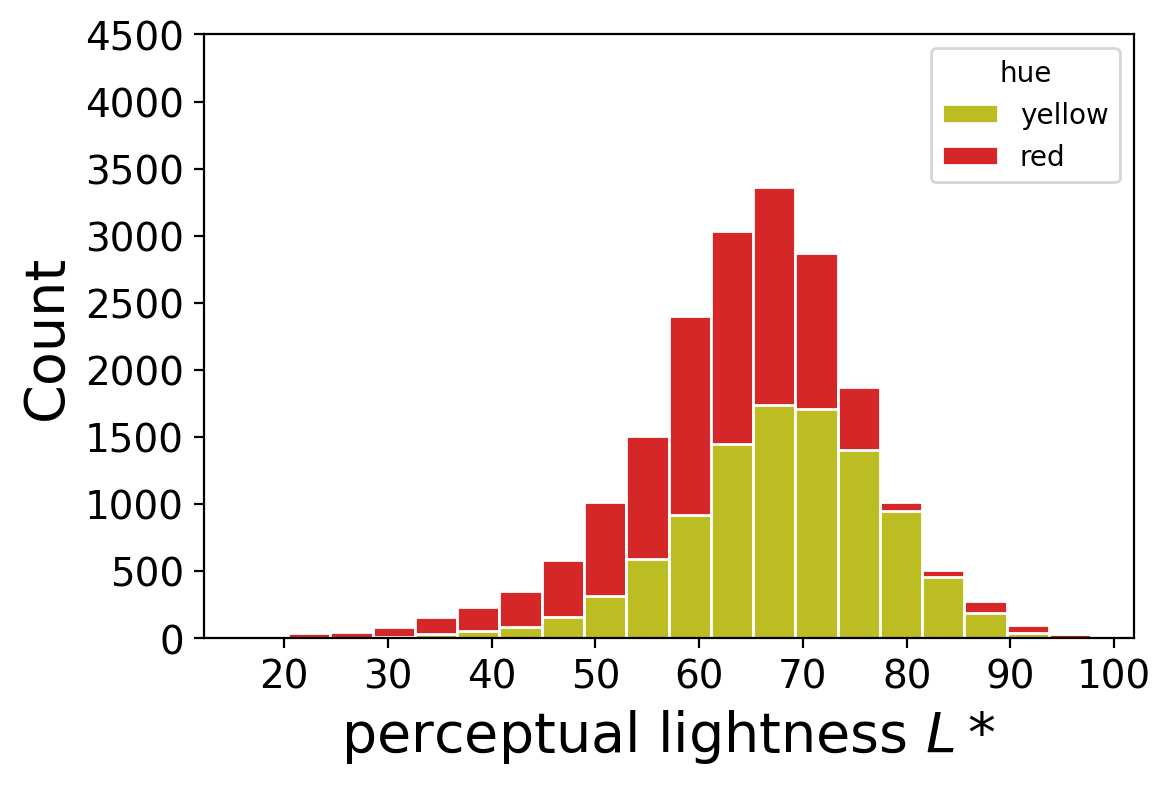}\hfill\null
  \vspace{-0.5em}
  \caption{CelebAMask-HQ\label{fig:app:data:celeba}}
  \end{subfigure}\\
  \begin{subfigure}{\linewidth}
  \centering
  \hfill
  \includegraphics[width=.5\linewidth]{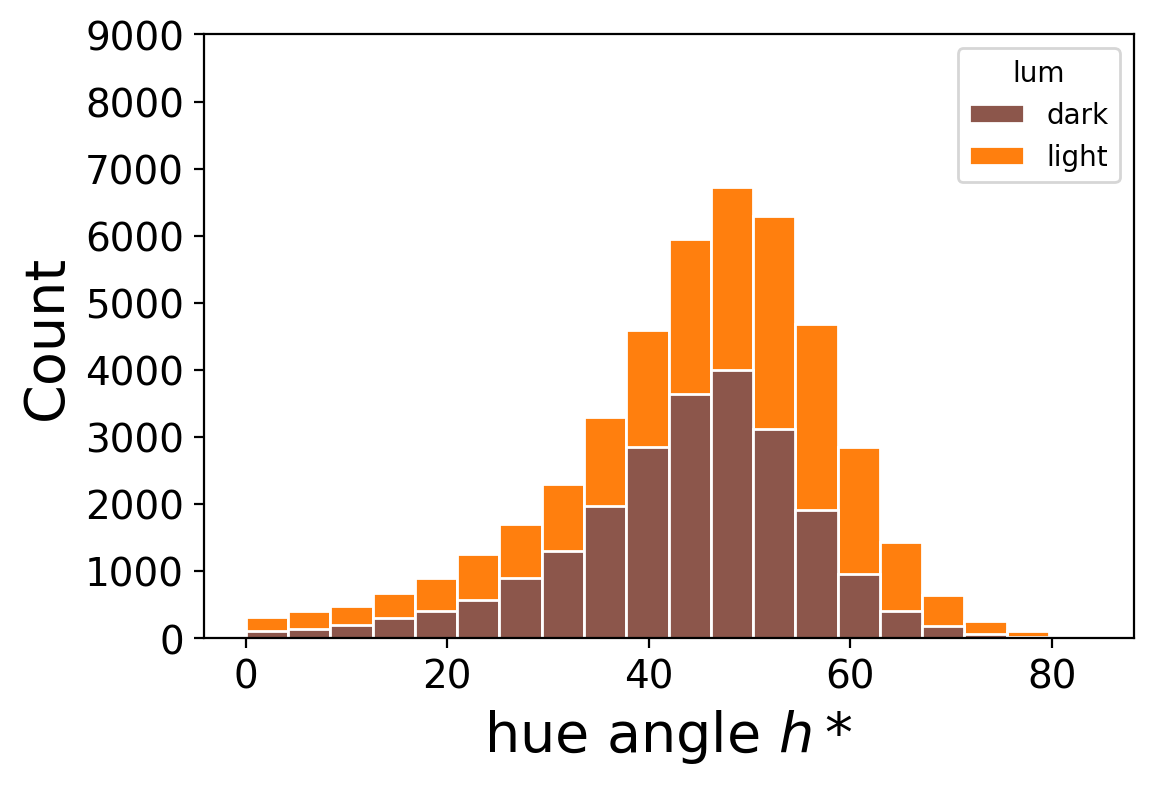}\hfill
  \includegraphics[width=.5\linewidth]{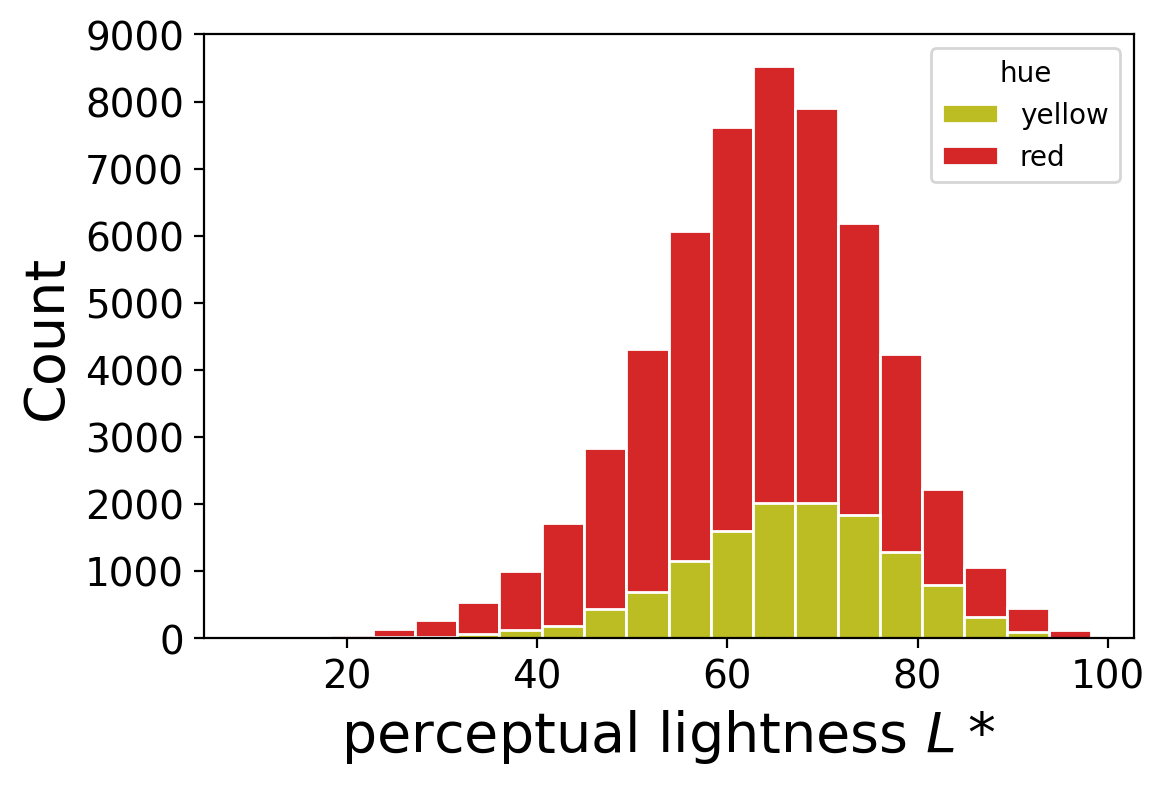}\hfill\null
  \vspace{-0.5em}
  \caption{FFHQ\label{fig:app:data:ffhq}}
  \end{subfigure}
  \caption{\textbf{Skin color distribution} on common face datasets.
  Histograms show a dominance of light skin tone and red skin hue in both CelebAMask-HQ and FFHQ.
  }
  \label{fig:app:data}
\end{figure}

\subsection{Additional results on skin color bias in common face datasets}
\label{sec:app:data}

Figure~\ref{fig:app:data} offers an alternative representation to highlight the skewed skin color distribution in common face datasets.
Instead of a binary thresholding for both perceptual lightness $L^*$ and hue angle $h^*$, we plot histograms of both scores with 20 bins.
In both CelebAMask-HQ and FFHQ, distributions are unimodal with a bell curve shape. Individuals with a light skin tone and a red skin hue are over-represented with a much larger count.
When considering the skin tone and varying the hue angle thresholding, we observe that the hue angle has a lower spread for dark skin tones than light skin tones.
Conversely, when considering the skin hue and varying the perceptual lightness thresholding, the yellow skin hue tends to have a larger skewness towards light skin tones than the red skin hue.
These representations confirm the relevance of a multidimensional measure for skin color, which could help increase the diversity when collecting a human-centric dataset.